\documentclass[10pt]{article} % For LaTeX2e
\usepackage[preprint]{tmlr}

\usepackage{amsmath,amsfonts,bm}

\def\eqref#1{equation~\ref{#1}}
\def\1{\bm{1}}

\def\va{{\bm{a}}}

\def\vd{{\bm{d}}}
\def\ve{{\bm{e}}}

\def\vp{{\bm{p}}}
\def\vq{{\bm{q}}}
\def\vr{{\bm{r}}}

\def\vu{{\bm{u}}}
\def\vv{{\bm{v}}}

\def\vx{{\bm{x}}}

\def\mA{{\bm{A}}}

\def\mE{{\bm{E}}}

\def\mI{{\bm{I}}}

\def\mL{{\bm{L}}}

\def\mS{{\bm{S}}}

\def\mU{{\bm{U}}}

\def\mX{{\bm{X}}}

\def\mZ{{\bm{Z}}}

\DeclareMathAlphabet{\mathsfit}{\encodingdefault}{\sfdefault}{m}{sl}
\SetMathAlphabet{\mathsfit}{bold}{\encodingdefault}{\sfdefault}{bx}{n}

\newcommand{\sigmoid}{\sigma}

\newcommand{\Var}{\mathrm{Var}}

\DeclareMathOperator*{\argmin}{arg\,min}

\usepackage{algorithm}
\usepackage{algcompatible}
\usepackage{comment}
\usepackage{soul}
\usepackage{wrapfig}
\usepackage{layouts} % may be causing warning: Warning: Layout scale set to 0.5 on input line 87.
\usepackage{pgfplots} % Required for plotting with TikZ
\usepgfplotslibrary{groupplots}
\pgfplotsset{compat=1.16, width=7cm, height=6cm}
\usepackage{multirow}
\usepackage{tabularx}

\usepackage{siunitx}
\usepackage{float}

\usepackage[utf8]{inputenc} % allow utf-8 input
\usepackage[T1]{fontenc}    % use 8-bit T1 fonts
\usepackage{hyperref}       % hyperlinks
\usepackage{url}            % simple URL typesetting
\usepackage{booktabs}       % professional-quality tables
\usepackage{amsfonts}       % blackboard math symbols
\usepackage{nicefrac}       % compact symbols for 1/2, etc.
\usepackage{microtype}      % microtypography
\usepackage{xcolor}         % colors
\usepackage{amsthm}         % defining theorem like environments

\usepackage{graphicx}
\usepackage{graphbox}
\usepackage{subfigure}

\newtheorem{lemma}{Lemma}

\title{Graph Structure Learning with Interpretable\\ Bayesian Neural Networks}

\author{\name Max Wasserman \email mwasser6@cs.rochester.edu \\
      \addr Department of Computer Science\\
      University of Rochester
      \AND
      \name Gonzalo Mateos \email gmateosb@ece.rochester.edu\\
      \addr Department of Electrical and Computer Engineering\\
      University of Rochester
      }

\def\month{MM}  % Insert correct month for camera-ready version
\def\year{YYYY} % Insert correct year for camera-ready version
\def\openreview{\url{https://openreview.net/forum?id=XXXX}} % Insert correct link to OpenReview for camera-ready version

\begin{document}

\maketitle

\begin{abstract}
% Most common mistake in abstract to avoid: To talk about results before the reader is ready to understand them.
%% The first sentence orients the reader by introducing the broader field in which the particular research is situated.
Graphs serve as generic tools to encode the underlying relational structure of data. Often this graph is not given, and so the task of inferring it from nodal observations becomes important.
%% Then, this context is narrowed until it lands on the open question that the research answered. A successful context section sets the stage for distinguishing the paper’s contributions from the current state of the art by communicating what is missing in the literature (i.e., the specific gap) and why that matters (i.e., the connection between the specific gap and the broader context that the paper opened with).
Traditional approaches formulate a convex inverse problem with a smoothness promoting objective and rely on iterative methods to obtain a solution. In supervised settings where graph labels are available, one can unroll and truncate these iterations into a deep network that is trained end-to-end. 
Such a network is parameter efficient and inherits inductive bias from the optimization formulation, an appealing aspect for data constrained settings in, e.g., medicine, finance, and the natural sciences. But typically such settings care equally about \textit{uncertainty} over edge predictions, not just point estimates. 
%% The content (“Here we”) first describes the novel method or approach that you used to fill the gap or question. Then you present the meat—your executive summary of the results.
Here we introduce novel iterations with \textit{independently interpretable parameters}, i.e., parameters whose values - independent of other parameters' settings - proportionally influence characteristics of the estimated graph, such as edge sparsity. %; e.g. increasing the value of a parameter $\theta$ produces increasingly sparser graph outputs. 
After unrolling these iterations, prior knowledge over such graph characteristics shape \textit{prior distributions} over these independently interpretable network parameters to yield a Bayesian neural network (BNN) capable of graph structure learning (GSL) from smooth signal observations.
%The iterations are constructed so as to decouple the sparsity promoting parameter from the edge-weight scale parameter. This induces a notion of parameter interpretability: the value of the parameters have an understandable effect on characteristics of the network output, which we leverage to construct \textit{informative} prior distributions.
Fast execution and parameter efficiency %- characteristic to unrollings - 
allow for high-fidelity posterior approximation via Markov Chain Monte Carlo (MCMC) and thus uncertainty quantification on edge predictions. Informative priors unlock modeling tools from Bayesian statistics like prior predictive checks. %previously unavailable to BNNs. 
Synthetic and real data experiments corroborate this model's ability to provide well-calibrated estimates of uncertainty% - those that can be used as a proxy to error in prediction -
, in test cases that include unveiling economic sector modular structure from S$\&$P$500$ data and recovering pairwise digit similarities from MNIST images. 
%% Finally, the conclusion interprets the results to answer the question that was posed at the end of the context section. There is often a second part to the conclusion section that highlights how this conclusion moves the broader field forward (i.e., “broader significance”). This is particularly true for more “general” journals with a broad readership.
Overall, this framework enables GSL in modest-scale applications where uncertainty on the data structure is paramount.
\end{abstract}

%%%%%%%%%%%%%%%%%%%%%%%%%%%%%%%%%%%%%%%%%%%%%%%%%%%%

\section{Introduction}\label{sec:intro}

%%%%%%%%%%%%%%%%%%%%%%%%%%%%%%%%%%%%%%%%%%%%%%%%%%%%

%%%%%
%% The introduction highlights the gap that exists in current knowledge or methods and why it is important. This is usually done by a set of progressively more specific paragraphs that culminate in a clear exposition of what is lacking in the literature, followed by a paragraph summarizing what the paper does to fill that gap.

%% Structure of intro paragraphs:
% 1. Context = 1-2 sentences to orient the reader to the topic
% 2. Content = explains the “knowns” in the relevant literature
% 3. Conclusion = critical “unknown” that makes the paper matter at the relevant scale.

%%%%%

%% Background Paragraph: Generic intro for why GSL matters
Graphs serve as a foundational paradigm in machine learning, data science, and network science for modeling complex systems, capturing intricate relationships in data that range from social networks to gene interactions; see e.g.,~\citep{kolaczyk2009book,hamilton2020grl}. In computational biology, accurate graph structures can offer insights into gene regulatory pathways, enhancing our ability to treat diseases at the genetic level~\citep{cai2013inference}. In finance, the ability to recover precise financial networks can be useful for risk assessment and market stability~\citep{marti2021review}. In geometric deep learning~\citep{bronstein2017gdl}, the lack of observed graph structure underlying data often limits the use of efficient learning models such as graph neural networks (GNNs); see e.g.,~\citep{cosmo2020latent}. Yet, despite their importance, existing methodologies for graph structure learning (GSL) from multivariate nodal data face significant limitations. 

%% Background Paragraph: Traditional GSL methods and Unrollings
This work addresses the GSL problem where nodal observations are used to predict the completely unobserved graph structure. Traditional approaches summarize the nodal observations with a pairwise vertex (dis)similarity matrix, e.g., correlation or Euclidean distance matrices, and formulate GSL as a regularized convex inverse problem~\citep{mateos2019ConnecDotsSpMag,dong2019SignalRepresentation}. The primary objective encourages data fidelity, according to a chosen model linking the nodal observations and the sought latent graph, while the regularization objectives capture prior structural knowledge, such as edge sparsity or desired connectivity patterns. These \emph{model-based} methods solve the inverse problem using optimization algorithms, which often come with convergence rate guarantees~\citep{Saman2021DPG,wang2023rate}. However, as noted in~\citep{Pu2021L2L} their expressiveness is constrained to graph characteristics that can be modeled via convex criteria and they may suffer complexity and scalability issues. Specifically, these model-based approaches typically necessitate an outer loop for grid-based optimization of regularization parameters (and often other algorithm parameters, e.g., a step-size), and an inner loop that, given this fully defined optimization problem, may demand thousands of iterations to converge on a solution. When nodal observations come with corresponding graph labels, recent supervised GSL approaches partially overcome these obstacles using `algorithm unrolling'~\citep{Shrivastava2020GLAD, Pu2021L2L, wasserman2023GDNtmlr}. Unrollings truncate these inner-loop iterations to yield a deep network architecture that is trained end-to-end to approximate the solution to the inverse problem~\citep{monga2021spmag}. The choice of a custom loss function, which need not be convex, plus optional architectural refinements of an already well-motivated initial network tend to improve performance on the task of interest. Truncation depth provides explicit control on the complexity of prediction, now a forward pass in the network. And the use of backpropogated gradients makes learning the regularization (and step-size) parameters, now network parameters, more scalable. 
% Explicit control on prediction complexity + The iterations are typically simple, fast to execute, and require few parameters --> appealing in real-time and memory constrained applications.
Additionally, the unrolled deep networks tend to inherit appealing aspects from the original model-based formulation, namely inductive bias and low dimensionality, as their outputs are approximations to solutions of the original inverse problem.
\begin{comment}
%%% Inductive Bias
% - Why inductive bias? Network output ARE THE OPTIMIZATION VARIABLE at the D-th step of the optimization procedure and thus represent APPROXIMATIONS TO AN OPTIMAL POINT in the original inverse problem.
% - Specific Sources of Inductive Bias:
%   -- It can (not always) automatically satisfy constraints before any learning takes place. Examples:
%       -- Unrolled DPG has symmetriy, non-negativity, and no disconnected nodes at each step (is lastpart true?)
%       -- GLAD (unrolled AM on l1 reg. MLE for Gaussian data): always outputs symmetric, PD graphs
%   -- Highly Constrained Function Space
%      -- Few parameters --> smaller function space --> less data needed
%   -- Highly Structured Function Space
%      -- The parameters (originally optimization weights) \textit{paramterize} steps in the network which promote their respective objective terms. These objective terms encode prior knowledge. E.g. \lambda in ISTA scales \ell-1 term and now parameterizes a soft-threshold function. With suitable chosen loss, this makes it easy for the model to learn values of these paramters to promote outputs with will reduce loss. This is soemtimes called "learning the prior" in unrolling, particularly plug-and-play literature.
%%% Low Dimensionality
% - convex inverse problems tend to have only a few regularization terms
\end{comment}

\subsection{Towards interpretability and uncertainty quantification: Desiderata and contributions}\label{ssec:desiderata_contributions}

%% Novel Gap/Contributions Paragraph: Interpretability and Unrollings
% Contributions to motivate:
% 1: Clearly defining, recognizing, and leveraging independent interpretability with Bayesian techniques
In composite inverse problems with multiple regularizers, understanding how each regularization parameter values affect desirable solution traits (e.g., image sharpness or number of graph edges) may be complex and often beyond the modeler's knowledge. This obscurity necessitates a naive—and consequently costly—multi-dimensioanl grid search for suitable parameter values; see e.g., the numerical study in~\citep[Section V-B]{dong2016learning}. %For example, consider denoising an observed image $\mY$ via $\argmin_{\mX} \|\mX-\mY\|_2^2 + \lambda_1 \|\mX\|_{\text{TV}} + \lambda_2\|\mX\|_2^2$, where $\|\mX\|_{\text{TV}}$ is the total-variation norm of $\mX$. Here, we may care about the pixel-wise sparsity, sharpness, or aliasing present in $\mY$, but such characteristics can be a complicated function of both $\lambda_1$ and $\lambda_2$, leading to a discrete search for their appropriate values. 
When the relationship between the regularization parameters and a solution characteristic is clear, we say the parameters are interpretable with respect to (w.r.t.) the output characteristic. {When an output characteristic is influenced by a single parameter, independent of all others, we call this parameter \textit{independently interpretable} w.r.t. the output characteristic}. Such instances make interpretability actionable, namely they allow prior knowledge on the solution characteristic to be incorporated onto the value of the independently interpretable parameter. A key contribution of this work is: i) in recognizing optimization problems with independently interpretable parameters as ripe for the adoption of unrollings, which inherit this interpretability because they approximate inverse problem solutions; and ii) in leveraging the Bayesian framework to seamlessly incorporate such prior knowledge into the parameters of the unrolled network.

%% Novel Gap/Contribution Paragraph: True Neural Network Unrollings
% Contributions to motivate:
% 1: True Neural Network Unrolling by removing parameter products
An issue unrollings often face (within GSL and beyond) is their tendency to produce layers with pre-activation outputs that are nonlinear functions of the parameters; e.g., parameter products and parameters in denominators~\citep{monga2021spmag}. This prevents the deep network from being a \textit{true neural network} (NN), i.e., 
{a function composed of layers, where each layer consists of affine transformations of data or intermediate activations followed by non-linear functions applied pointwise~\citep{bishop1995neural}.}
%a composition of affine transformations followed by point wise non-linearities. 
To address this issue, network designers may opt for reparameterization, but at the expense of parameter interpretability and often degraded empirical performance~\citep{monga2021spmag, Shrivastava2020GLAD}. Unrollings can be impeded by nuisance parameters - parameters not of direct interest, but which must be accounted for - like a step-size. Nuisance parameters are typically a vestige of the chosen optimization algorithm rather than being intrinsic to the problem formulation. %% nuissance paramter: a parameter that is not of direct interest but must be accounted for in the model. These parameters do not contribute to the primary objective of the analysis but can influence the estimates of the parameters of interest. 
Nuissance parameters can also undermine training stability (see Appendix~\ref{app:pds}) and since they lack a clear connection to specific solution characteristics they hinder informative prior modeling. The preceding discussion motivates the need for innovative GSL techniques to produce true NN unrollings amenable to incorporation of prior knowledge on their parameters.

%% Novel Gap/Contribution Paragraph: Uncertainty Quantification for GSL
% Contributions to motivate:
% 1: first GSL unrolling method with uncertainty estimates on it's outputs
% 2: first BNN for GSL
% 3: first BNN for GSL with informative prior modeling over network parameters.
% 4: ?? high fidelity posterior approximation ??

% why is uncertainty estimates important for GSL
Providing estimates of uncertainty on an inferred graph structure is important for downstream applications, e.g. in biology, finance, and machine learning with GNNs.
% the general bayesian approach modeling
In general Bayesian statistics, low-dimensional models with interpretable parameters are constructed using background information on the specific problem. Interpretability allows informative prior modeling and thus access to enticing Bayesian modeling tools, e.g., prior predictive checks~\citep{gabry2017visualizationBayesWorkflow}, % see How Good is the Bayes Posterior in Deep Neural Networks Really? for a prior predictive experiment in BNN (naive tho)
while the low dimensionality allows tractable high-quality posterior inference, typically via Markov Chain Monte Carlo (MCMC) sampling. 
% The traditional Bayesian approach to GSL
For example, traditional Bayesian approaches to GSL tend to address the transductive (tied to a single graph) setting by constructing a joint model over parameters, observed data, and latent graph structure, use MCMC to draw samples, and marginalize out all but the graph samples~\citep{butts2003network, crawford2015hidden, gray2020bayesianInfNetStruct}.
% the general BNN approach to modeling
Stepping back from the specific case of GSL, the inability to specify a sufficiently expressive model often motivates the use of Bayesian neural networks (BNNs), an alternative paradigm which typically bypasses domain-specific modeling opting instead to feed the output of a performant NN directly into the likelihood function; see e.g.,~\citep{jospin2022BNNTutorial}.
% how we get uncertainty estimation in both
Both traditional Bayesian and BNN approaches derive uncertainty estimates over predictions by marginalizing out the parameter posterior. 
% difficulties with BNNs: prior construction/prior knowledge incorporation +  posterior approximation
However, BNNs present notable challenges, especially in the incorporation of prior information and posterior approximation. Due to the non-interpretability of parameters, priors are often selected for computational convenience, such as a zero-mean isotropic Gaussian, which can inadvertently bias predictions \textit{against} prior beliefs, a phenomenon known as `unintentionally informative priors'~\citep{Wenzel2020HowGoodBayesPosterior}, thus negating a key benefit of Bayesian methods. Additionally, the parameters' high dimensionality and complex posterior geometry make high-quality posterior approximation a formidable task, often requiring significant approximations that undermine the interpretability of the results~\citep{Gal2016McDropout, Wenzel2020HowGoodBayesPosterior, monga2021spmag}. 
% the gap
This highlights the need for an inductive GSL method that not only provides reliable uncertainty estimates over edge predictions but also combines the expressive power of BNNs with the traditional Bayesian approach's strengths in integrating prior knowledge, employing robust modeling tools like predictive checks, and ensuring high-fidelity posterior approximation.

%% Summary of Gaps Filled/Technical Approach
%
%%%% The last paragraph of the introduction is special: it compactly summarizes the results, which fill the gap you just established. It differs from the abstract in the following ways: it does not need to present the context (which has just been given), it is somewhat more specific about the results, and it only briefly previews the conclusion of the paper, if at all.
\noindent\textbf{Summary of contributions.} In this paper, we introduce the first BNN for supervised GSL from smooth signal observations. This BNN produces a distribution over unseen test graphs allowing estimation of uncertainty over edge predictions. It leverages the independent intepretability of the paramters in the GSL formulation to allow informative prior modeling over the weights of the NN, itself % the neural network
a result of unrolling novel iterations for a model-based formulation with well-documented merits. We make the following contributions:
\begin{itemize}
    \item In Section~\ref{sec:graph-learning-smooth-signals}, we develop a novel optimization algorithm for GSL from smooth signals (Algorithm~\ref{alg:dpg-iterates}), which is step-size free and parameterized to yield independent interpretability w.r.t. edge sparsity.    %We formalize the notion of independent interpretability, and show how unrollings applied to optimization problems with this structure are amenable to the incorporation of prior information on their parameters.
    \item In Section~\ref{sec:gsl_smooth_signals_with_bnns}, we unroll Algorithm~\ref{alg:dpg-iterates} to produce the first strict unrolling for GSL from smooth signals, which results in a true NN. This is to be contrasted with existing supervised-learning approaches to GSL, where GLAD~\citep{Shrivastava2020GLAD} and Unrolled PDS~\citep{Pu2021L2L} are not true NNs, and GDN~\citep{wasserman2023GDNtmlr} is not a strict unrolling because it resorts to gradient truncation and reparameterization. The proposed unrolled NN is used to define our BNN, dubbed `DPG' since Algorithm~\ref{alg:dpg-iterates} is a dual-based proximal gradient method~\citep{beck2014fdpg}.
    \item In Section~\ref{sec:bayesian_modeling}, we introduce a methodology to integrate prior knowledge into BNN prior distributions, specifically for networks derived from unrolled optimization algorithms for inverse problems with independent interpretability. We show this approach unlocks classical Bayesian modeling tools like predictive checking, which we fruitfully apply to DPG. High-fidelity parameter posterior inference via Hamiltonian Monte Carlo (HMC) sampling enables {the first instance of a model for GSL from smooth signals, capable of producing estimates of uncertainty over edge predictions}. % TODO: IF we include Bayesian perspective on depth, note it here
    \item In Section~\ref{sec:experiments}, we validate DPG's ability to produce high-quality and well-calibrated uncertainty estimates from synthetic data, stock price time series (S$\&$P$500$), as well as graphs learnt from images of MNIST digits. The reliability of these estimates is underscored by notable Pearson correlations between predictive uncertainty and error: $0.70$ for the stock data and $0.62$ for the MNIST digits. Additional experimental evidence is provided in the appendices.
    % Other Misc Novelties
    % - Derive a computationally efficient formula for the euclidean distance matrix of smooth graph signals in the infinite signals case.
    % - First transfer learning for smooth graph learning unrolling. Largest graphs than all previous for smooth unrolling setting.
\end{itemize}

%%%%%%%%%%%%%%%%%%%%%%%%%%%%%%%%%%%%%%%%%%%%%%%%%%%%

\section{Related Work}\label{sec:related}

%%%%%%%%%%%%%%%%%%%%%%%%%%%%%%%%%%%%%%%%%%%%%%%%%%%%

%%%%%%%%%%%%%%%%%%%%%%%  GSL via unrollings %%%%%%%%%%%%%%%%%%%%%%% 
A recent body of work addresses the GSL task via algorithm unrollings under varying data assumptions. Noteworthy contributions include
GLAD~\citep{Shrivastava2020GLAD}, which unrolls alternating-minimization iterations for Gaussian graphical model selection, Unrolled PDS~\citep{Pu2021L2L} which unrolls primal-dual splitting (PDS) iterations for the GSL problem from smooth signals, and GDN~\citep{wasserman2023GDNtmlr} which unrolls linearized proximal-gradient iterations for a network deconvolution task that posits a polynomial relationship between the observed pairwise vertex distance matrix and the latent graph. Here we deal with the GSL problem from observations of smooth signals as in~\citep{Pu2021L2L}, but the optimization algorithm we unroll is different (cf. DPG in Algorithm~\ref{alg:dpg-iterates} versus PDS), more compact and devoid of step-sizes. Additionally, none of the previous GSL unrolling result in true NNs, provide estimates of uncertainty on their adjacency matrix predictions, nor leverage the interpretability of network parameters in the modeling process. 

Building a probabilistic model to connect observed network data to latent graph structure has a long history in the computer and social network analysis communities; see e.g.,~\citep{coates2002maximum, butts2003network,kolaczyk2009book}. Gibbs sampling was used mostly as a means to efficiently explore the resulting network posterior rather than quantify the uncertainty the model placed on particular edges. % see first paragraph of Section 4 of Coates 2002: "Maximum Likelihood Network Topology Identification" 2002 
Since then, this probabilistic modeling approach has been used with alternate network models, types of observed data (e.g., information cascades and protein-protein interactions), and posterior approximation approaches; see~\citep{ShaghaghianCoates2016bayesian, gray2020bayesianInfNetStruct, kolaczyk2012LatentEigenLinkUncertainty, sinead2016BayesNonParametricLinkPred, palCoates2019scalable}. Such approaches are typically transductive - tying themselves to a single training graph - and require expensive \textit{joint} inference of the model and the latent graph structure. Our approach only requires inference of the BNN model parameters and thus is naturally inductive, i.e., able to generalize to new nodes, or entirely new graphs.
% we construct priors only over BNN parameters and sample only from the parameter space, rather than the joint parameter-network space. This makes us inductive (not tied to single training graph) and makes posterior sampling more tractable (way lower dimension)
%% building on previous probabilistic approach of graph
Some recent works build on such graph distribution modeling approaches in an approximate Bayesian manner, to incorporate the uncertainty in an observed graph for downstream tasks with GNNs~\citep{ZhangCoates2019bayesian, palCoates2020NonParamGraphLearnBNNs}. 
%% Other misc bayesian approaches
Others forgo modeling the distribution of the observed graph but take approximate Bayesian approaches to modeling with GNNs. For instance,~\citep{opolka2022GraphGausianProcesses} uses a deep graph convolutional Gaussian process with variational posterior approximation for link prediction, and~\citep{sevilla2023langevinGnn} pre-trains a score-matching GNN for use in annealed Langevin diffusion to draw approximate samples from the network posterior. All such Bayesian GNN approaches require (partial) observation of graph structure, and rely on approximate inference methods due to large dimensionality. Our DPG approach uses no observed graph structure (except for graph labels during training) and allows for high-fidelity posterior approximation. To the best of our knowledge, BNNs have so far not been used for GSL with uncertainty quantification.

%%%%%%%%%%%%%%%%%%%%%%% BNNs with Unrollings  %%%%%%%%%%%%%%%%%%%%%%%
% Related Papers
\begin{comment}
Other Related Papers
2020 - Quantifying Model Uncertainty in Inverse Problems via Bayesian Deep Gradient Descent, https://arxiv.org/abs/2007.09971
2020 - Quantifying Sources of Uncertainty in Deep Learning-Based Image Reconstruction, https://arxiv.org/abs/2011.08413 (Builds on Previous paper).
2021 - Conditional Variational Autoencoder for Learned Image Reconstruction, https://arxiv.org/abs/2110.11681 (VAE method)
2022 - Uncertainty Quantification for Deep Unrolling-Based Computational Imaging, https://arxiv.org/abs/2207.00698 (MC Dropout on Unrolling Inspired CNN - Canberk paper)
\end{comment}
%
% cover the term 'learning the prior'?
More broadly, unrolling-inspired Bayesian deep networks have recently found success in uncertainty quantification for computational imaging~\citep{barbano2020quantifying, zhang2021conditional, canberk2022uncertainty}. %, commonly replacing the proximal operators with convolutional neural networks. 
The inductive bias provided by the original iterations lead to gains in data efficiency, but still have limited parameter interpretability and high dimensionality leading to naive priors and coarse posterior approximation. This exciting line of work inspired some crucial ideas in this paper, cross-pollinating benefits to GSL.
\begin{comment}
%% BNN related papers and resources
% Parameter Interpretability highly influenced by "Functional Variational Bayesian Neural Networks", https://arxiv.org/pdf/1903.05779.pdf

% Original BNN papers: ~\citep{hinton1993BNN, neal2012BNN}

% How to choose Priors: ~\citep{Blundell2015BayesByBackprop} - computational convienenve, ~\citep{ghosh2018structured, louizos2017bayesianCompression} - a specific purpose e.g. model compression or model selection

% Variational BNN methods: ~\citep{hinton1993BNN, graves2011VarBNN, Blundell2015BayesByBackprop, Gal2016McDropout, Daxberger2021Laplace}

% Learn Variational Approximation to posterior predictive directly: ~\citep{rudner2021rethinking-VarPostPred, Sun2019Func-VarPostPred}

% Approximate Posterior Sampling: ~\citep{2011Welling-SGLD}

% Questioning how good posterior approximation is, e.g. poor calibration and generalization: \citep{Wenzel2020HowGoodBayesPosterior}

% Partial Stochasticity: "Do Bayesian Neural Networks Need To Be Fully Stochastic?", https://arxiv.org/pdf/2211.06291.pdf

% Evaluating Uncertainty Estimates from a Model: (see https://www.youtube.com/watch?v=veYq6EWZyVc).\\
\end{comment}
%%%%%%%%%%%%%%%%%%%%%%%%%%%%%%%%%%%%%%%%%%%%%%%%%%%%

\section{Model-based Formulation and Optimization Preliminaries}
\label{sec:graph-learning-smooth-signals}

%%%%%%%%%%%%%%%%%%%%%%%%%%%%%%%%%%%%%%%%%%%%%%%%%%%%
% graph notation
Let $\mathcal{G}(\mathcal{V}, \mathcal{E}, \mA)$ be an undirected graph, where $\mathcal{V}=\{1,\ldots N\}$ are the vertices (or nodes), $\mathcal{E} \subseteq \mathcal{V} \times \mathcal{V}$ are the edges, and $\mA \in \mathbb{R}_{+}^{N \times N}$ is the symmetric adjacency matrix collecting the non-negative edge weights. For $(i, j) \notin \mathcal{E}$ we have $A_{ij} = 0$. We exclude the possibility of self loops, so that $\mA$ is hollow meaning $A_{ii} = 0$, for all $i \in \mathcal{V}$. For the special case of unweighted graphs that will be prominent in our models, then $\mA \in \{0,1\}^{N \times N}$.

In this paper, we consider that $\mA$ is unknown and we want to estimate the latent graph structure from nodal measurements only\footnote{This is different to the link prediction task, where one is given measurements of edge status for a training subset of node pairs (plus,  optionally, node attributes), and the transductive goal is to predict test links from the same graph\citep{kolaczyk2009book}}. To this end, we acquire graph signal observations $\vx = [x_1, \dots, x_N]^{\top} \in \mathbb{R}^N$, where $x_i$ denotes the signal value (i.e., a nodal attribute or feature) at vertex $i \in \mathcal{V}$. When $P$ such signals are available we construct matrix $\mX = [\vx_1, \dots, \vx_P]\in \mathbb{R}^{N \times P}$, where each row $\bar{\vx}^{\top}_i \in \mathbb{R}^{P}$, $i=1,\ldots,N$, of $\mX$ represents a vector of features or nodal attributes at vertex $i$. We can summarize this dataset using a pairwise vertex dissimilarity matrix, here the Euclidean distance matrix $\mE \in \mathbb{R}_{+}^{N \times N}$, where $E_{ij} = \| \bar{\vx}_i - \bar{\vx}_j \|_2^2$.
% smoothness: the manifold perspective
% Gonzalo: No need to add a statistical ingredient here, when we have not introduced any probability distribution on the data. Let's avoid any potential confusions and stick to geometric interpretations.
% TODO: Possible Follow up work: Dig into this for particular manifolds...see how optimization problem changes...any new corrections terms for a sphere?
Assuming our data lie on a smooth %statistical 
manifold, we interpret $\mathcal{G}$ as a discrete representation of this manifold. When nodes $i\neq j \in\mathcal{V}$ have large edge weight $A_{ij}$, reflecting close points on the manifold, %small geodesic distance on the manifold, 
%the probability distribution at these points are defined to be similar and 
$E_{ij}$ will be small. %, expressed as $\|\vx_i - \vx_j\|^2$. 
%See~\citep{amari2016information} for an introduction to statistical manifolds. 
% smoothness: on graphs
Accordingly, smooth (w.r.t. $\mathcal{G}$) vectors in $\mX$ have small total variation or
%-- whose topology is encoded via the combinatorial graph Laplacian $\mL = \text{diag}(\mA\mathbf{1}) - \mA$ -- 
Dirichlet energy~\citep{Belkin2001DirichletEnergy}, namely
\begin{equation}\label{eq:dirichlet}
TV_{\mathcal{G}}(\mX)=\frac{1}{2}\sum_{i,j} A_{ij} \|\bar{\vx}_i - \bar{\vx}_j\|_2^2 %= \text{tr}(\mX^{\top}\mL\mX) 
= \|\mA \circ \mE\|_{1,1},    
\end{equation}
where $\|\mZ\|_{1,1}$ is the $\ell_1$-norm of $\mZ$ and $\circ$ is the Hadamard (entrywise) product. The prevalence of smooth network data, for instance sensor measurements~\citep{chepuri2017learning}, protein function annotations~\citep{kolaczyk2009book}, and product ratings~\citep{huang2018rating}, justifies using a smoothness criterion for the GSL task.

\subsection{Graph structure learning from smooth signals}\label{ssec:gsl_problem}

Given $\mX$ assumed to be smooth on $\mathcal{G}$, a popular model-based GSL approach is to minimize the Dirichlet energy in (\ref{eq:dirichlet}) w.r.t. $\mA$; see e.g.,~\citep{hu2013graph, dong2016learning, Kalof2016Smooth, kalofolias2019LargeScaleGL}. %include \citep{lake2010SparseLaplacian}if room.
The inverse problems posed in these works can be unified under the general composite formulation
\begin{equation}\label{eq:smooth-graph-optim-problem}
\mA^*=\argmin_{\mA \in \mathcal{A}}\left\{ \|\mA \circ \mE\|_{1,1} + h(\mA)\right\},    
\end{equation}
where the feasible set is $\mathcal{A} := \{ \mA \in \mathbb{R}^{N \times N} :\text{ diag}(\mA) = \mathbf{0}, A_{ij} = A_{ji} \geq 0, \forall i, j \in \mathcal{V} \}$, i.e., hollow, symmetric,  non-negative  matrices. The regularization term $h(\mA)$ typically promotes desired structure on the estimated edge set (e.g., sparsity, no isolated nodes) and can be used to avoid the trivial solution $\mA^* = \mathbf{0}$. We henceforth use $h(\mA) = - \alpha \mathbf{1}^{\top} \text{log}(\mA \mathbf{1}) + \frac{\beta}{2}\|\mA\|_F^2$ ($\alpha,\beta\geq 0$ are regularization parameters), which excludes the possibility of isolated nodes and has achieved state-of-the-art results~\citep{Kalof2016Smooth}.

It is convenient to reformulate (\ref{eq:smooth-graph-optim-problem}) in an unconstrained, yet equivalent form. We start by compactly representing variable $\mA$ and data matrix $\mE$ with their vectorized upper triangular parts $\va, \ve \in \mathbb{R}_{+}^{N(N-1)/2}$, implicitly enforcing symmetry and hollowness, while also halving the problem dimension. To enforce non-negativity the indicator function $\mathbb{I}\{\va \geq 0\} = \{0$ if $\va \geq 0$ else $\infty\}$ is included in the objective. Finally, we substitute the nodal degrees $\vd=\mA\mathbf{1}$ with the vectorized equivalent $\vd=\mS \va$, where $\mS \in \{0, 1\}^{N \times N(N-1)/2}$ is a fixed binary matrix that maps vectorized edge weights to degrees. The resulting optimization problem is given by
\begin{equation}
\label{eq:smooth-graph-optim-problem-vectorized}
\va^*(\ve, \alpha, \beta) =  \argmin_{\va \in \mathbb{R}^{N(N-1)/2}} \hspace{3pt} \left\{2\va^{\top}\ve - \alpha \mathbf{1}^{\top} \text{log}(\mS \va) + \frac{\beta}{2} \|\va\|^2_2 + \mathbb{I}\{\va \geq 0\}\right\},
\end{equation}
which is convex and admits a unique optimal solution; see e.g.,~\citep{Saman2021DPG}. Next, we comment on the role of the regularization parameters $\alpha,\beta$ and their interpretability properties. We then offer a brief discussion on optimization algorithms to tackle problem (\ref{eq:smooth-graph-optim-problem-vectorized}). These ingredients will be essential to build a BNN model for supervised GSL in Sections \ref{sec:gsl_smooth_signals_with_bnns} and \ref{sec:bayesian_modeling}.

\textbf{Independent interpretability of regularization parameters.} The weights $\alpha$ and $\beta$ are not independently interpretable w.r.t. to relevant % or any I can think of
graph characteristics, frustrating straightforward interpretation of their effect on the solution $\va^*(\ve, \alpha, \beta)$. Specifically, for fixed $\alpha$, increasing $\beta$ leads to denser edge patterns, as we have (quadratically) increased the relative cost of large edge weights. Indeed, the sparsest graph is obtained for $\beta=0$. But in general, many interesting graph characteristics, e.g., sparsity, connectivity, diameter, and edge weight magnitude, are non-trivial functions of \textit{both} $\alpha$ and $\beta$; see also~\citep{dong2016learning} for a similar issue. 

To facilitate \textit{independent} control over the sparsity pattern and scale of the edge weights of recovered graphs,~\citep[Prop. 2]{Kalof2016Smooth} %, kalofolias2019LargeScaleGL} 
introduced an equivalent ($\theta, \delta$)-parameterization of (\ref{eq:smooth-graph-optim-problem-vectorized}), namely
\begin{equation}
    \label{eq:reparam-smooth-graph-optim-problem-vectorized}
    \va^*(\ve, \alpha, \beta) = \sqrt{\frac{\alpha}{\beta}} \va^*\left(\frac{1}{\sqrt{\alpha \beta}} \ve, 1, 1\right) = \delta \va^*(\theta \ve, 1, 1).
\end{equation}
We can map from the former parameterization to the latter by first scaling $\ve$ by $\theta = 1/\sqrt{\alpha \beta}$, solving (\ref{eq:smooth-graph-optim-problem-vectorized}) with $\theta \ve$ using $\alpha = \beta = 1$, and finally scaling the recovered edges by the constant $\delta = \sqrt{\alpha/\beta}$; we refer the reader to~\citep{Kalof2016Smooth,kalofolias2019LargeScaleGL} for a proof of the equivalence claim. 
Due to the separable structure of the right-hand-side of (\ref{eq:reparam-smooth-graph-optim-problem-vectorized}), any GSL algorithm would require a single input parameter $\theta$, and the obtained solution $\va^*(\theta \ve, 1, 1)$ can then be scaled by $\delta$. All in all, {the sparsity level of $\va^*$ is determined solely by $\theta$, making $\theta$ independently interpretable w.r.t. sparsity}. Moreover, $\delta$ is interpretable w.r.t. edge weight magnitude, but not independently so, as larger $\theta$ produces smaller weights [see Figure~\ref{fig:prior_modeling_theta_delta} (bottom-left)].
\begin{comment}
% Some things to consider including
% 1) The ($\theta, $\delta$) parameterization has independent control of sparsity/edge weights, which is in contrast to the ($\alpha, \beta$)-parameterization which instead allows direct (but still NOT independent), influence on nodal degrees ($\alpha$) and edge weight magnitude ($\beta$).
% 2) In the Model-Based setting where we are performing a grid/random/bayesian search over the parameter space, this allows us to *independently* search over \theta for good structure recovery and THEN search over \delta for good scaling. Dramatic reduction in computation because exponential growth in parameter space.
\end{comment}
%

\subsection{Optimization algorithms}\label{ssec:optimization}

Problem (\ref{eq:smooth-graph-optim-problem-vectorized}) has a favorable structure that has been well documented, and several efficient optimization algorithms were proposed to obtain a solution $\va^*(\ve, \alpha, \beta)$ with $\mathcal{O}(N^2)$ complexity per iteration. Specifically, a forward-backward-forward PDS algorithm was first proposed in \citep{Kalof2016Smooth}. PDS introduces a step-size parameter which must be tuned to yield satisfactory empirical convergence properties, thus increasing the overall computational burden.
% Gonzalo: This is an obvious tradeoff for any step-size, no need to elaborate
%, and can be difficult to work with: $\gamma$ values too large produce NaN's from divergent behavior and $\gamma$ values too small can produce impractically slow convergence. 
We find that effective step-size values tend to lie on a narrow interval beyond which PDS exhibits divergent behavior, further frustrating tuning; see Appendix~\ref{app:pds} for a supporting discussion. GSL algorithms based on the alternating-directions method of multipliers~\citep{wang2021admm} or majorization-minimization~\citep{fatima2022mm} have been developed as well.
Recently,~\citep{Saman2021DPG} introduced a fast dual proximal gradient (FDPG) algorithm to solve (\ref{eq:smooth-graph-optim-problem-vectorized}), which is devoid of step-size parameters and -- different from all three previous approaches -- it comes with global convergence rate guarantees. For this problem, the strongest convergence results to date are in~\citep{wang2023rate}.

\begin{figure}[!t]
  \centering
  \begin{minipage}[t]{0.46\textwidth}
    \vspace{0pt}
    \begin{algorithm}[H]
      \caption{Dual Proximal Gradient Descent}
      \label{alg:dpg-iterates}
    \begin{algorithmic}
    \STATE \textbf{Inputs}: Fixed parameters $\theta, \delta \in \mathbb{R}$ and data $\ve$
    \STATE \textbf{Initialize}: $\va_0$ and $\boldsymbol{\lambda}_0$ at random.
    %\FOR{k = 1, 2, \dots}
    \FOR{$k = 1, 2, \dots$}
    \STATE $\textcolor{orange}{\vd_k} = \mS \textcolor{red}{\va_{k-1}} - (N-1) \textcolor{magenta}{\boldsymbol{\lambda}_{k-1}}$  
    \STATE $\textcolor{magenta}{\boldsymbol{\lambda}_k} = \frac{-1}{2(N-1)}\Big(\textcolor{orange}{\vd_k} - \sqrt{\textcolor{orange}{\vd_k}^2 + 4(N-1) \boldsymbol{1}}$ \Big) 
    \STATE $\textcolor{red}{\va_k} = \max \Big( \boldsymbol{0} , \frac{1}{2} \mS^\top \textcolor{magenta}{\boldsymbol{\lambda}_{k}} - \theta\ve \Big)$
    %\STATE $\textcolor{red}{\va_k} = \text{ReLU} \Big( \mS^\top \textcolor{magenta}{\boldsymbol{\lambda}_{k}} - \theta\ve \Big)$
    \ENDFOR
    \STATE \textbf{Return:} $\delta \va_{k}$
    \end{algorithmic}
    \end{algorithm}
    \end{minipage}
  \hfill
  \begin{minipage}[t]{0.53\textwidth}
    \vspace{0pt}
    \includegraphics[width=\textwidth, vshift=-5cm]{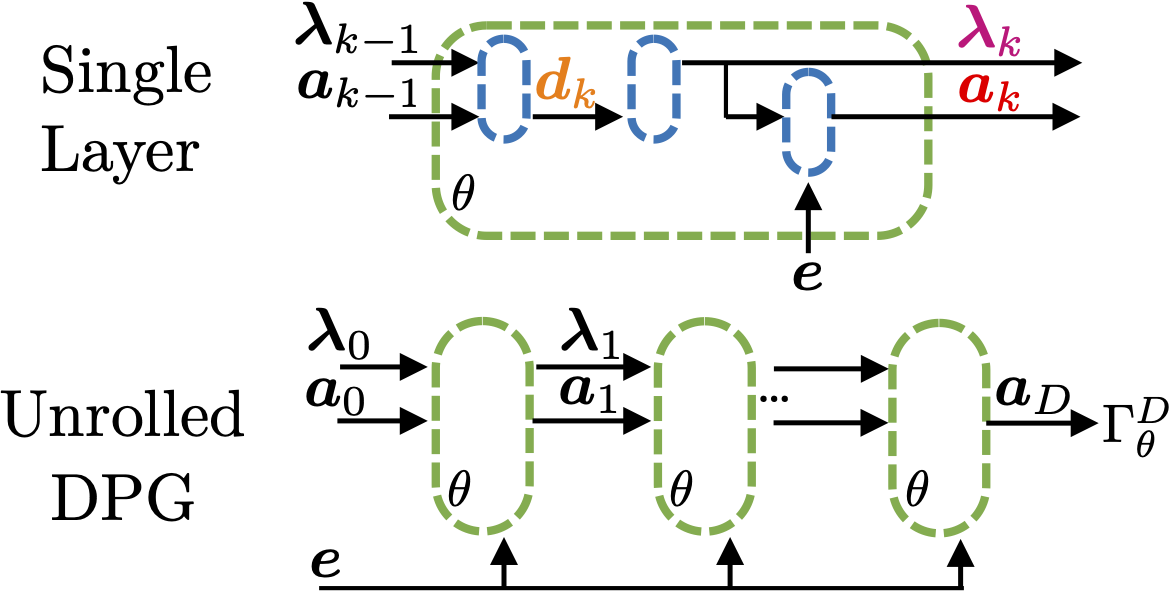}
    \label{fig:layer_and_unrolling}
  \end{minipage}
  \caption{ \emph{Left:} The dual proximal gradient (DPG) algorithm to solve the GSL problem (\ref{eq:smooth-graph-optim-problem-vectorized}). \emph{Right:}
    Unrolling and truncating Algorithm~\ref{alg:dpg-iterates} after $D$ iterations produces Unrolled DPG.}
\end{figure}

Our starting point in this work is the FDPG optimization framework, but different from~\citep{Saman2021DPG} we: (i) develop a solver for the ($\theta, \delta$)-parameterization of (\ref{eq:smooth-graph-optim-problem-vectorized}); and (ii) turn-off the Nesterov-type acceleration from the proximal-gradient iterations used to solve the dual problem of (\ref{eq:smooth-graph-optim-problem-vectorized}). This yields a dual proximal gradient (DPG) method, tabulated under Algorithm~\ref{alg:dpg-iterates}.
% Add this appendix later
%; see Appendix \ref{app:dpg} for details on the algorithmic construction process. 
In a nutshell, during iterations $k=1,2,\ldots$ Algorithm~\ref{alg:dpg-iterates} updates the vectorized adjacency matrix estimate $\va_k\in \mathbb{R}_{+}^{N(N-1)/2}$, an auxiliary vector of nodal degrees $\vd_k\in \mathbb{R}_{+}^{N}$, as well as dual variables $\boldsymbol{\lambda}_k\in \mathbb{R}^{N}$ used to enforce the variable splitting constraint $\vd=\mS\va$. A naive DPG implementation incurs $\mathcal{O}(N^2)$ computational and memory complexities, and we note all nonlinear operations involved (i.e., $\textrm{ReLU}(\cdot)=\max(0,\cdot)$, $(\cdot)^2$, and $\sqrt{(\cdot)}$) are pointwise on their vector arguments. As a result of the design choices (i)-(ii), the DPG algorithm requires the fewest operations per iteration and the fewest number of parameters among existing solvers of (\ref{eq:smooth-graph-optim-problem-vectorized}), and is devoid of any uninterpretable nuisance parameters, e.g., step-sizes. FDPG was only considered on the original $(\alpha, \beta)$-parameterization of (\ref{eq:smooth-graph-optim-problem-vectorized}); by instead opting for DPG iterations to solve the $(\theta, \delta)$-parameterization of (\ref{eq:smooth-graph-optim-problem-vectorized}), we reveal independent interpretability of $\theta$ w.r.t. sparsity of the optimal graphs. 

Next, we will unroll Algorithm~\ref{alg:dpg-iterates} to produce a GSL NN which inherits its advantages - namely simple, efficient, minimally parameterized layers, with independent interpretability - forming the backbone of our BNN. 
%

%%%%%%%%%%%%%%%%%%%%%%%%%%%%%%%%%%%%%%%%%%%%%%%%%%%%

\section{Graph Structure Learning from Smooth Signals with Bayesian Neural Networks}
\label{sec:gsl_smooth_signals_with_bnns}

%%%%%%%%%%%%%%%%%%%%%%%%%%%%%%%%%%%%%%%%%%%%%%%%%%%%

So far we have described a model-based approach to (point) estimation of graphs from smooth signals. In this work, we assume a labeled training dataset is available. We aim to construct a BNN model to produce uncertainty estimates on graph predictions for unseen test data.

\noindent\textbf{Our BNN approach for GSL in a nutshell.} Here, we restrict ourselves to binary graphs $\va \in \{0, 1\}^{N(N-1)/2}$; weighted graphs only require a change to the ensuing likelihood function.  We denote all training data as $\mathcal{T} = \{\mathcal{T}_e, \mathcal{T}_a\} =  \{\ve^{(t)}, \va^{(t)} \}_{t=1}^T$, an unseen test sample as $(\tilde{\ve}, \tilde{\va})$, and the collection of all parameters of our yet to be defined BNN model as $\boldsymbol{\Theta}$.
% brief intro to constructing a bayesian posterior predictive
Following the principles of BNNs, our goal is to construct a posterior predictive distribution $p(\tilde{\va} \mid \tilde{\ve}, \mathcal{T} )$ by marginalizing out model parameters $\boldsymbol{\Theta}$, i.e.,
\begin{equation}\label{eq:predictive_posterior}
p(\tilde{\va} \mid \tilde{\ve}, \mathcal{T}) = \int p(\tilde{\va} \mid \tilde{\ve}, \boldsymbol{\Theta}) \cdot p(\boldsymbol{\Theta} \mid \mathcal{T})
\, \textrm{d}\boldsymbol{\Theta} \approx \frac{1}{M} \sum_{m = 1}^M p(\tilde{\va} \mid \tilde{\ve}, \boldsymbol{\Theta}^{(m)}),
\end{equation}
where we use $M$ Monte Carlo draws from the posterior $\boldsymbol{\Theta}^{(m)} \sim p(\boldsymbol{\Theta}\mid  \mathcal{T})$ to approximate the intractable integral. We can then designate our edge-wise point and uncertainty estimates as the first two moments of the posterior predictive marginals $p(\tilde{a}_i \mid \tilde{\ve}, \mathcal{T})$, respectively, for each candidate edge indexed by $i \in \mathcal{V}\times \mathcal{V}$. 

\noindent\textbf{Roadmap.} In Section \ref{ssec:alg_unrolling}, we first discuss the development of a GSL NN which takes in the vectorized Euclidean distance matrix $\ve$ of nodal observations $\mX$ and assigns a probability to all possible edges. We do so by unrolling Algorithm~\ref{alg:dpg-iterates}, producing an GSL NN with an independently interpretable parameter w.r.t. sparsity of graph outputs. Treating the model parameters of this unrolled NN as stochastic, choosing a likelihood function (Section \ref{ssec:stoch_model}), and setting appropriate priors (Section \ref{sec:bayesian_modeling}), we produce a BNN. For inference we use HMC (Section \ref{ssec:inference}), an unusual choice in BNNs made viable by the low dimensionality and fast execution of our NN, stemming from its origins as an unrolling. Pushing the test inputs through each such GSL NN and averaging the resulting predictive distributions provides %the approximation  (\ref{eq:predictive_posterior}) to the posterior predictive, from which we derive edge-wise point and uncertainty estimates as elaborated in Section \ref{ssec:prediction}.
provides the approximate (\ref{eq:predictive_posterior}) posterior predictive, from which we derive edge-wise point and uncertainty estimates as elaborated in Section \ref{ssec:prediction}.

\subsection{Algorithm unrolling: Iterative optimization as a neural network blueprint}\label{ssec:alg_unrolling}
% new iterations =  Kalof (\theta, \delta) PDS formulation +  Saman FDPG w/o acceleration
% advantages of new iterates: true NN when unrolled, no uninterpretable lr parameter (problematic: divergence when too large, far from convergence when small), minimal parameter count, fewest operations & just plain simpler than other iterations
% quick 1-2 sentence on unrollings
{Unrollings (also known as deep unfoldings) use iterative algorithms as templates of neural architectures that are trained to approximate solutions of optimization problems. Iterations are truncated and mapped to NN layers, while optimization (e.g., regularization and step-size) parameters are turned into learnable weights. Originating from convergent iterative procedures, unrolled NNs inherit many desirable properties; namely, a low number of parameters, fast layer-wise execution, and favorable generalization in low-data environments~\citep{monga2021spmag}. 
%One can also inject prior information on a solution in the intitialization of optimization variables (here, $\a_0, \boldsymbol{\lambda}_0$). 
Typical architectural design choices include  whether to replace intermediate operators %(typically proximal operator) 
with more expressive (possibly pre-trained) learnable modules - making them no longer `strict' unrollings - and 
whether a common set of parameters should be used in all layers, or, to decouple these parameters across layers. Both decisions are context dependent and often driven by the amount of data available as well as complexity considerations. Deviating from strict unrollings with shared parameters can naturally lead to gains in expressive power, but often at the cost of inductive bias and training stability.}
%

% what has been done before: PDS, problems with PDS, how to solve these problems
\noindent\textbf{Unrolling DPG iterations}. Pioneered by \citep{gregor2010fastApproxSparseCodes} to efficiently learn sparse codes from data, algorithm unrolling ideas are recently gaining traction for GSL as well. Starting from the formulation in Section \ref{ssec:gsl_problem}, \citep{Pu2021L2L} unrolls a PDS solver of the $(\alpha, \beta)$-parameterization (\ref{eq:smooth-graph-optim-problem-vectorized}). Unrolled PDS has no independently interpretable parameters, is not a true NN (pre-activation outputs are nonlinear functions of the parameters), and includes a nuisance step-size parameter  -- arguably a shortcoming as gradient-based learning can easily produce large-enough weight values for divergent behavior, and thus NaN's. 
% OPTIONAL: Parameter products also limit identifiability. The product of parameters determines the output, not each individually
Attempting to unroll PDS iterates for the $(\theta, \delta)$-parameterization of (\ref{eq:smooth-graph-optim-problem-vectorized}) would not fix these practical problems. 
% Optional: include statement on why not unroll FDPG on (\theta, \deltat) param directly: empirically does not work as well, more expensive, another hyperparameter to tune.

% solution: by instead unrolling DPG to (\theta, \delta)-param of (1), we produce iterations with only 2 parameters
We instead advocate unrolling the DPG iterations (Algorithm~\ref{alg:dpg-iterates}) developed to solve the $(\theta, \delta)$-parameterization of (\ref{eq:smooth-graph-optim-problem-vectorized}). This way, we obtain a \textit{true NN without nuisance step-size parameters} - avoiding the aforementioned issues and reducing the parameter count by a third - while inheriting independently interpretable parameter $\theta$ (w.r.t. sparsity of graph outputs). Incidentally, layers in Unrolled DPG [depicted in Figure \ref{alg:dpg-iterates} (right)] are markedly simpler and require fewer operations than Unrolled PDS. We denote the output of a $D$-layer unrolling of Algorithm~\ref{alg:dpg-iterates} as $\delta \Gamma_{\theta}^D$. As we would like probabilities over candidate edges, we subtract a learnable mean shift $b$ and drive the output through a sigmoid $\sigma(\cdot)$, producing our desired GSL NN output
\begin{equation}\label{eq:unrolled_DPG_output}
\hat{\vp}=\sigmoid(\delta \Gamma_{\theta}^D(\ve) - b \mathbf{1}) \in (0, 1)^{^{N(N-1)/2}},
\end{equation}
with parameters $\boldsymbol{\Theta} = \{\theta, \delta, b\}$.  To see that Unrolled DPG is a true NN, note that $\theta$ is only involved in a linear function $\theta \ve$ of the input data. Likewise, $\delta$ and $b$ are only involved in an affine mapping of the activations $\Gamma_{\theta}^D(\ve)$. All non-linear operations (specifically squaring, square root, and $\max$) are pointwise functions of intermediate activations. Going back to the design considerations mentioned at the beginning of this section, here we keep the unrolling strict and share parameters across layers to retain independent interpretability of $\theta$, minimize parameter count, and simplify upcoming Bayesian inference. Tradeoffs arising with model expansion using multiple input and output channels per layer are discussed in  Section \ref{ssec:model_metrics_details}. 
% notation, and massaging positive outputs into probabilities.

All in all, unrolled DPG is the first true NN for GSL from smooth signals, and the first strict unrolling for GSL which produces a true NN. For the various reasons laid out in the preceding discussion, Unrolled DPG is of independent interest as a new model for point estimation of graph structure in a supervised setting.  As we show next, it will be an integral component of the stochastic model used to construct a BNN to facilitate uncertainty quantification for adjacency matrix predictions.

\subsection{Stochastic model}\label{ssec:stoch_model}
Here we specify a stochastic model for the random variables of interest, namely the binary adjacency matrix $\va$, nodal data entering via the Euclidean distance matrix $\ve$, and the BNN weights $\boldsymbol{\Theta}$. The Bayesian posterior from which we wish to sample satisfies $p(\boldsymbol{\Theta} \mid \mathcal{T}) \propto p(\mathcal{T}_a \mid \mathcal{T}_e, \boldsymbol{\Theta}) p(\boldsymbol{\Theta})$, and a first step in BNN design is to specify the likelihood $p(\va \mid \ve, \boldsymbol{\Theta})$ and the prior $p(\boldsymbol{\Theta})$. 
% conditional independence assumptions and resulting factorizations
An i.i.d. assumption on the training data $\mathcal{T}$ allows the likelihood to factorize over samples $ p(\mathcal{T}_a \mid \mathcal{T}_e, \boldsymbol{\Theta}) = \prod_{t=1}^T p(\va^{(t)} \mid \ve^{(t)}, \boldsymbol{\Theta})$. Moreover, assuming edges within a graph sample $\va^{(t)}$ are mutually conditionally independent given parameters $\boldsymbol{\Theta}$ leads to further likelihood factorization as $ p(\mathcal{T}_a \mid \mathcal{T}_e, \boldsymbol{\Theta}) = \prod_{t=1}^T \prod_{i=1}^{N(N-1)/2} p(a^{(t)}_i \mid \ve^{(t)}, \boldsymbol{\Theta})$. % pairwise independence i.e. $y_i \perp y_j | \Theta; \ve$ for $i \neq j$, <-- DOES NOT allow joint to factorize
We model $a_i\mid \ve, \boldsymbol{\Theta}\sim \textrm{Bernoulli}(\hat{p}_i)$, where the success (or edge $i\in\mathcal{V}\times \mathcal{V}$ presence) probability is given by the output of the Unrolled DPG network, i.e., $\hat{p}_i = \sigma\Big(\delta [\Gamma_{\theta}^D(\ve)]_i - b\Big)$ as in (\ref{eq:unrolled_DPG_output}). Putting all the pieces together, the final expression for the likelihood is 
%$p(\mathcal{T}_y \mid \Theta; \mathcal{T}_e) = \prod_{t=1}^T \prod_{i=1}^{|\mathcal{E}|} {\hat{p}^{(t)}_i}^{y^{(t)}_i}(1-\hat{p}^{(t)}_i)^{(1-y^{(t)}_i)}$.
%
\begin{equation}\label{eq:likelihood}
p(\mathcal{T}_a \mid  \mathcal{T}_e, \boldsymbol{\Theta}) = \prod_{t=1}^T \prod_{i=1}^{N(N-1)/2} (\hat{p}^{(t)}_i)^{a^{(t)}_i}(1-\hat{p}^{(t)}_i)^{1-a^{(t)}_i}.
\end{equation}
We reiterate that, crucially, the Unrolled DPG outputs enter the stochastic model via the likelihood, as the means of the edge distributions. The specification of the prior $p(\boldsymbol{\Theta})$ will be addressed in Section~\ref{sec:bayesian_modeling}. 
 % removes error
%Together, which allows further factorization over edges $p(\vy \mid \Theta; \ve) = \prod_{i=1}^{|\mathcal{E}|} \hat{p}_i^{y_i}(1-\hat{p}_i)^{(1-y_i)}$. %, where $\hat{p} = \sigmoid(\delta \Gamma_{\theta}^D(\ve) - b)$

% I do not think this equation adds much
%All together, the log joint can be written as 
%
%\begin{align}
 %   \label{eq:log-joint-factorized}
  %  \text{log } p(\mathcal{T}_y , \mathcal{T}_e, \Theta) &= \text{ log } p(\Theta) +  \sum_{t=1}^{T} \sum_{i=1}^{|\mathcal{E}|} y^{(t)}_i \text{log } \hat{p}^{(t)}_i + (1-y^{(t)}_i) \text{log }(1-\hat{p}^{(t)}_i)
%\end{align}
%%.

%
\subsection{Inference}\label{ssec:inference}

We aim to generate $M$ Monte Carlo draws from the posterior $\boldsymbol{\Theta}^{(m)} \sim p(\boldsymbol{\Theta} \mid  \mathcal{T}) \propto p(\mathcal{T}_a \mid \mathcal{T}_e, \boldsymbol{\Theta}) p(\boldsymbol{\Theta})$. In modern BNNs based on overparameterized deep networks, high dimensionality and parameter unidentifiability are prevalent. This translates to highly multi-modal posteriors which make sampling challenging~\citep{izmailov2021bayesian}. Even when posterior geometries are more benign, the computational and memory requirements for evaluating the network and its gradients render the generation of Monte Carlo posterior samples impractical. As a typical workaround one can resort to posterior approximation using inexpensive mini-batch methods such as mean-field variational inference or stochastic-gradient MCMC~\citep{izmailov2021bayesian}. 
In contrast, we now argue that generation of high-fidelity Monte Carlo posterior samples using HMC is uniquely compatible with our proposed BNN; for implementation details see Section \ref{ssec:model_metrics_details}.
%% What makes our posterior well behaved
% identifiability (weak? local?)
% informative prior modeling helps for smoother geometry in low data settings
%\textit{Our posterior geometries tend to better behaved}. Unrolled DPG approximates solutions to a strongly convex problem and so with gentle assumptions on convergence we can show (strong? weak? local? if important, I can prove...) identifiability, limiting multi-modality of the posterior. 

% does this need citation for this? This is clearly true by inspection. I suppose we can cite Bishop's ML book, Murphy's Prob. ML book, Gelman's Bayes. Data Analysis.
%% What makes it fast
% simple network: easily implement in auto-grad framework + easy to compile network op's
% low dimensionality: allow for forward-mode auto-diff; memory complexity indep of depth
% dont need deep networks: limit runtime complexity
For starters, Unrolled DPG's repetitive layers simplify model construction and allow straightforward compilation in automatic differentiation software, significantly boosting runtime efficiency. The low parameter count of Unrolled DPG facilitates the use of forward-mode auto-differentiation; this eliminates the need to store intermediate activations and perform a backward pass, thus it ensures memory usage does not increase with model depth $D$. Moreover, as Unrolled DPG can effectively approximate inverse problem solutions with relatively shallow depths (see Section~\ref{sec:experiments}), selecting a smaller $D$ significantly reduces runtime complexity. {For small- to moderately-sized graphs and datasets, our computational and memory requirements for network and gradient evaluation are low}; see Section~\ref{sec:experiments} for GSL problem instances where inference is attainable in $<2$ minutes on a M$2$ MacBook laptop.
%% What makes it sample efficient: low dimensional <--> avoid curse of dimensionality
{The limited parameter count also aids in sampling efficiency} as we avoid the curse of dimensionality that complicates sampling in higher-dimensional models. Finally, the ability to place informative priors over independently interpretable parameters (as we do in Section~\ref{sec:bayesian_modeling}) is known to smoothen the posterior geometry, especially in data scarce regimes where the likelihood and prior are of comparable magnitude~\citep{gelman1995BDA}.
\subsection{Prediction}\label{ssec:prediction}
By conditioning on data $\mathcal{T}$ and integrating out model parameters $\boldsymbol{\Theta}$, we obtain a predictive distribution on unseen graph adjacency matrices $\tilde{\va}$ given nodal signals $\tilde{\ve}$. As the integral is intractable, we use the $M$ posterior samples obtained via HMC in the inference stage to approximate the predictive distribution $p(\tilde{\va} \mid \tilde{\ve}, \mathcal{T}) $. This approximation process is summarized in
(\ref{eq:predictive_posterior}). . 
% For how to approximate things in the posterior predictive: https://mc-stan.org/docs/stan-users-guide/sampling-from-the-posterior-predictive-distribution.html

Importantly, we can generate samples from the posterior predictive %$\tilde{y}^{(m)} \sim p(\tilde{\vy} \mid \mathcal{T}; \tilde{\ve})$ 
by randomly drawing from the sampling distribution with each parameter sample plugged in, i.e., $\tilde{\va}^{(m)} \sim p(\tilde{\va} \mid \tilde{\ve}, \boldsymbol{\Theta}^{(m)})$. Given the form of our BNN's stochastic model as introduced in Section \ref{ssec:stoch_model}, such a draw reduces to sampling a Bernoulli distribution for each possible edge. %with probability of success $\hat{p}^{(m)}_i$. %from our mutual conditional independence assumption on edges
Randomly drawing from the sampling distribution is critical as it accounts for both forms of uncertainty in posterior predictive quantities, namely, sampling uncertainty and estimation uncertainty
~\citep{gelman1995BDA}. Using these posterior predictive samples, we can approximate the mean (`pred. mean') and standard deviation (`pred. stdv.') of the edge-wise marginals of the posterior predictive as 
\begin{equation}
\mathbb{E}[\tilde{a}_i \mid \tilde{\ve}, \mathcal{T}] \approx \frac{1}{M}\sum_{m=1}^M \tilde{a}^{(m)}_i \quad 
\text{and} 
\quad 
\Var[\tilde{a}_i \mid \tilde{\ve}, \mathcal{T}]^{\frac{1}{2}} \approx \left[\frac{1}{M-1}\sum_{m=1}^M (\tilde{a}_i^{(m)} - \mathbb{E}[\tilde{a}_i \mid \tilde{\ve}, \mathcal{T}])^2\right]^{\frac{1}{2}}.
\label{eq:edge_point_and_uncert_estimate}
\end{equation}
Naturally, $\mathbb{E}[\tilde{a}_i \mid \tilde{\ve}, \mathcal{T}]$ is a Bayesian point estimate for $\tilde{a}_i$, while $\Var[\tilde{a}_i \mid \tilde{\ve}, \mathcal{T}]^{\frac{1}{2}}$ offers a measure of uncertainty in such prediction of edge $i$.

%%%%%%%%%%%%%%%%%%%%%%%%%%%%%%%%%%%%%%%%%%%%%%%%%%%%

\section{Bayesian Modeling of Unrolling-Based BNNs with Independent Interpretability}
\label{sec:bayesian_modeling}

%%%%%%%%%%%%%%%%%%%%%%%%%%%%%%%%%%%%%%%%%%%%%%%%%%%%
%
\begin{figure}
    \centering
    \includegraphics[width=\textwidth, vshift=0cm]{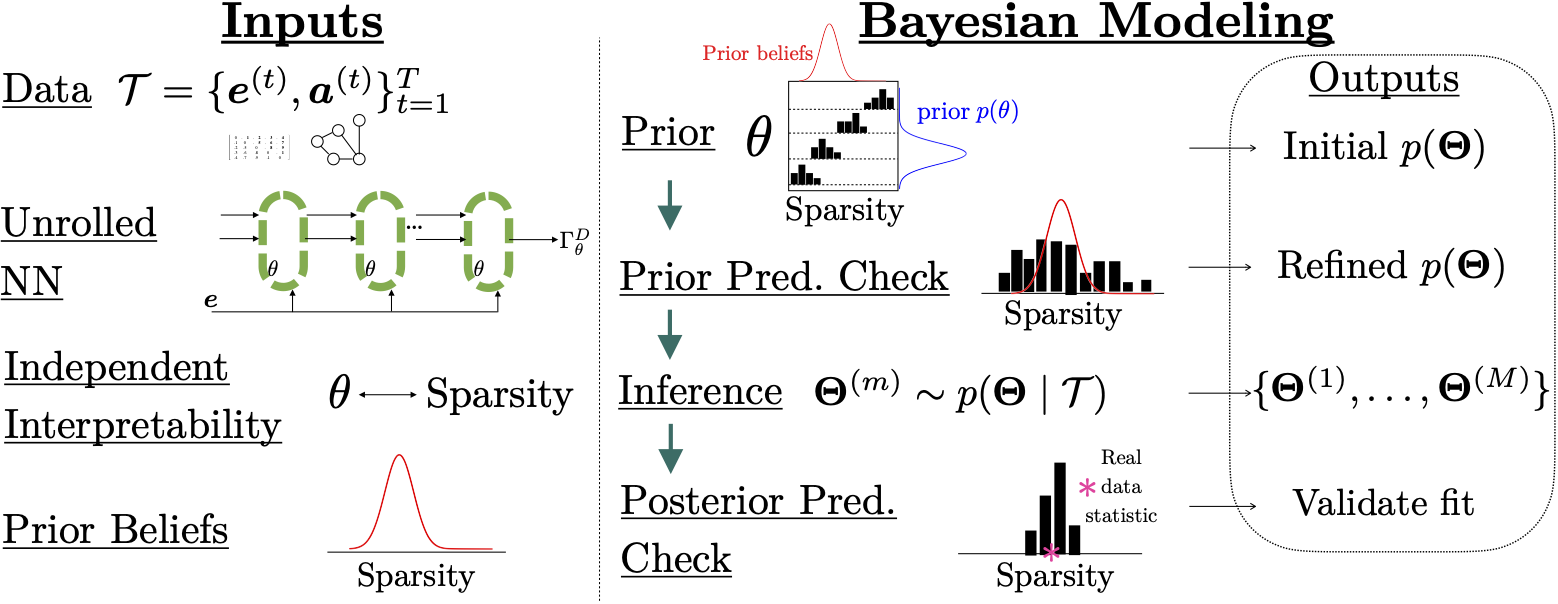}
    \caption{Bayesian workflow with independent interpretability. Inputs: A labeled data set $\mathcal{T}$, an inverse problem with independently interpretable parameter $\theta$ w.r.t. some characteristic of the solution, prior beliefs over this solution characteristic, and an unrolled NN which approximate solutions to the inverse problem. Bayesian Modeling: We use independent interpretability of $\theta$ to convert prior beliefs on solution characteristics to a prior distribution on the independently interpretable parameter. % which independently influences it. 
    We use prior predictive checks to ensure priors generate data sets which encompass all plausible values of the solution characteristic, while still preferentially generating data sets which we believe are more likely apriori. If not, we can leverage independent interpretability to refine the prior. We then sample from the posterior, and use posterior predictive checks to provide a subjective validation of model fit.
    }
    \label{fig:bayes_workflow}
\end{figure}
%
% Commented out since it is somewhat repetitive
%Solutions to inverse problems are often objects about which there is significant prior knowledge. For instance, in image deconvolution, the recovered images typically exhibit spatial correlations and natural scene statistics~\citep{krishnan2011blindDeconv}, while in seismology, when inferring wave-speeds from seismogram data, the wave-speeds are generally smooth and vary gradually over space~\citep{malinverno2004expanded}. And GSL is no exception. A contribution of this work is recognizing a subset of such inverse problems where some interesting characteristic(s) of the solution are uniquely determined by a single optimization parameter. As stated in Section \ref{ssec:gsl_problem}, we call such optimization parameter independently interpretable w.r.t. this solution characteristic. When solution iterates to such a problem are appropriately unrolled, this independently interpretable regularization parameter, now a NN parameter, retains its interpretability as the network approximates solutions to the original optimization problem. This allows us to transform prior information about the solution into a prior distribution over the independently interpretable NN parameter. Doing so unlocks modeling tools from Bayesian statistics like predictive checks, typically unavailable to BNNs. 

In this section, we present a method for Bayesian modeling for BNNs which use a network produced by unrolling an optimization algorithm to solve an inverse problem with independent interpretability.
% We leave exploration of other optimization problems with independently interpretable parameters, e.g., ISTA with the $\ell$-$1$ weight controlling ..., for future work
We instantiate these ideas on the GSL problem (\ref{eq:smooth-graph-optim-problem-vectorized}) - where solutions are undirected graphs - and use prior knowledge over the sparsity of such graphs to shape prior distributions over the corresponding independently interpretable parameter $\theta$ in Unrolled DPG introduced in Section \ref{ssec:alg_unrolling}.

\subsection{Prior modeling}\label{ssec:prior_modeling}

For Bayesian models, priors over unobserved quantities play two main roles: encoding information germane to the problem being analyzed and aiding the ensuing Bayesian inferences. Despite the name, a prior can in general only be interpreted in the context of the likelihood with which it will be paired. There are many types of priors, and we here we list the common ones in order of degree in which it is intended to affect the information in the likelihood: non-informative, reference, structural, regularizing, weakly informative, and strongly informative~\citep{gelman2017prior}. The more domain specific information present, the further to the end of this list we may be able to move, which often results in better behaved posterior geometries, and thus more stable Bayesian inference. Exactly how this domain specific information is encoded in the prior depends on both the problem and the specific chosen model.

\textbf{Informative priors with independent interpretability}. In the context of the GSL problem dealt with here, independent interpretability makes setting a strongly informative prior over DPG's independently interpretable parameter $\theta$ feasible. To this end, one first discretizes $\theta$ over a few orders of magnitude. For each $\theta$ value in the grid, we run Algorithm~\ref{alg:dpg-iterates} on a subset of observed inputs $\mathcal{T}_{\ve}$ and record the relevant data characteristic (edge sparsity) of each recovered solution. We then plot the histogram of such data characteristic and observe their agreement with prior beliefs. Finally, one chooses an appropriate parametric distribution $p(\theta)$ which approximates this relative level of agreement across the range of $\theta$ values. We can use the recovered solutions at a priori probabale $\theta$ values to set weakly informative priors over the remaining model parameters. In our setting, that involves the scale of the edge weights of the recovered solutions. In Section~\ref{ssec:pred_checking} we discuss evaluation of this initial $p(\boldsymbol{\Theta})$ in terms of the predictions it makes. For a visualization of this prior modeling workflow, see Figure~\ref{fig:bayes_workflow}. A numerical example is now presented to ground this methodology.

%\textit{Aside: the data}.    Posterior predictive checking require only the summary statistic on the graph labels $\mathcal{T}$, in this case their mean edge density.

\textbf{Example (Informative prior modeling of DPG parameters)}.
%
% what we are doing at a high level: using prior beliefs over the sparsity of graph solutions to shape prior distributions over its corresponding independently interpretable parameter $\theta$.
In this example that continues in Section ~\ref{ssec:pred_checking}, we use a data set $\mathcal{T}$ of $T=50$ random geometric graphs ($N=20$ nodes, connectivity of $\frac{1}{3}$), denoted RG$_{\frac{1}{3}}$, and their corresponding analytic Euclidean distance matrices as inputs defined in Section~\ref{sec:experiments}. See Appendix~\ref{app:further_experiments} for similar analysis on other random graph distributions. Prior modeling and upcoming prior predictive checks require only the distance matrices $\ve$, and we use only $5$ such distance matrices from $\mathcal{T}$. 

\begin{figure}[t]
\centering
  \begin{minipage}[b]{0.4\textwidth}
    \includegraphics[width=0.9\textwidth, vshift=0cm]{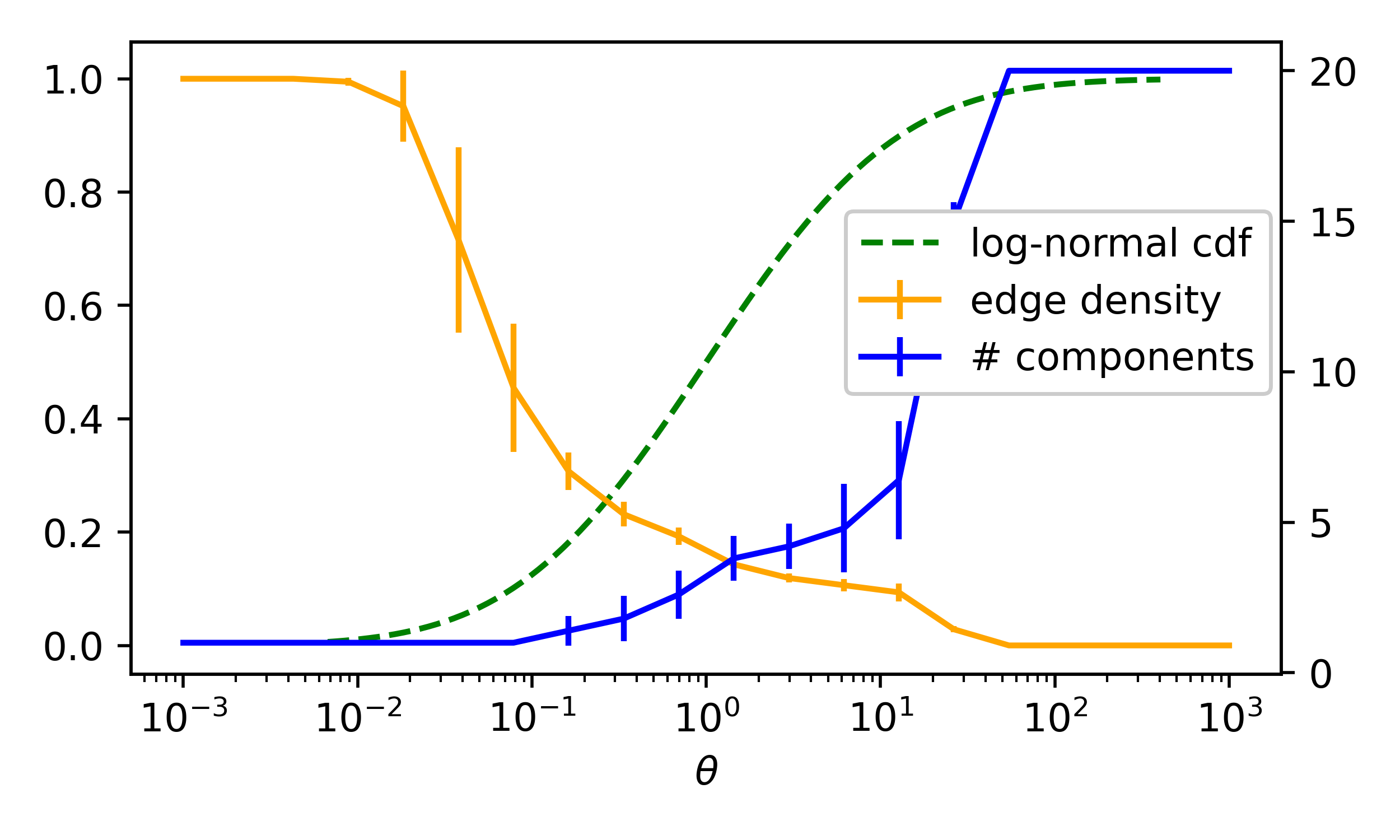}
    \includegraphics[width=0.85\textwidth, vshift=0cm]{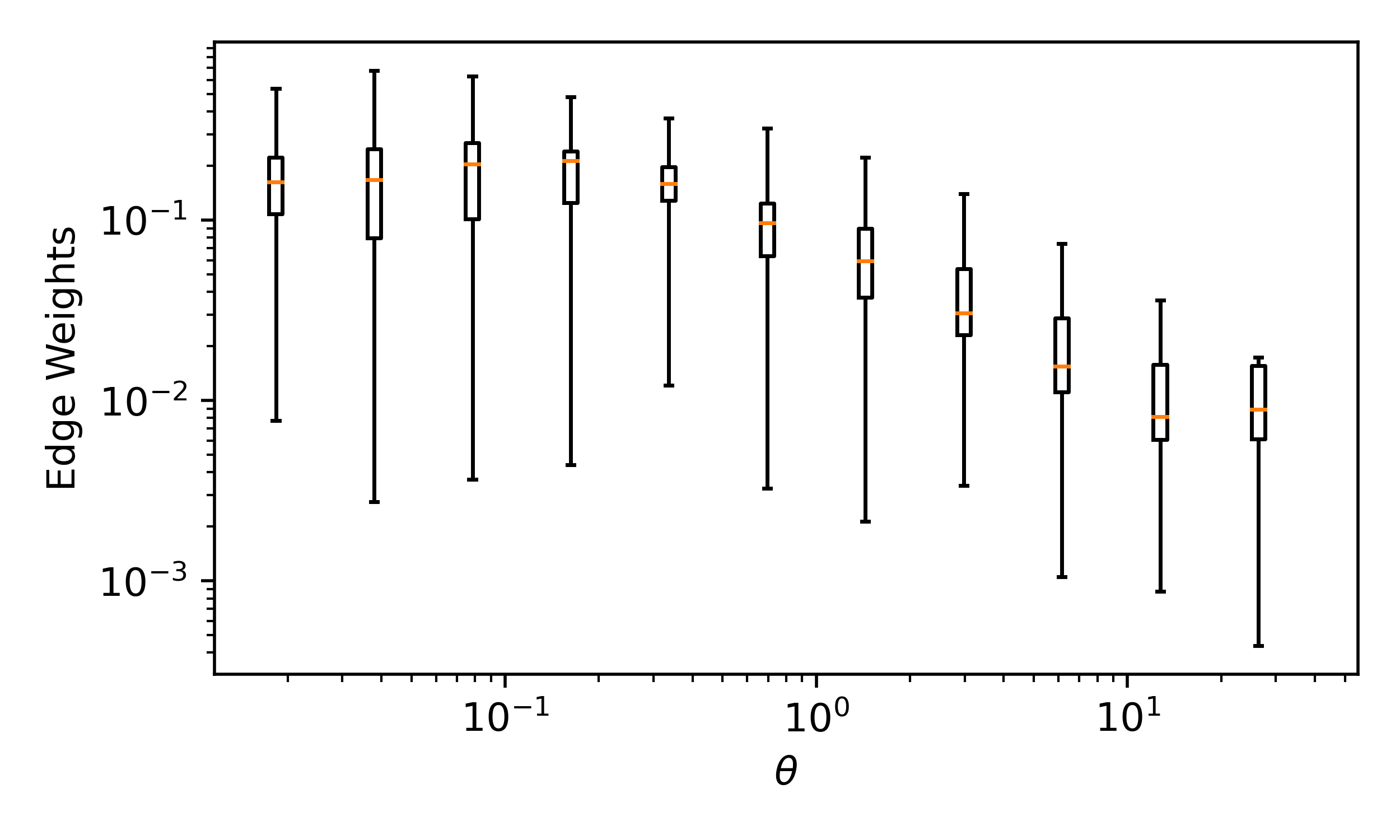}
  \end{minipage}
  \hfill
  \begin{minipage}[b]{0.58\textwidth}
   \centering
   \includegraphics[width=0.8\textwidth, vshift=0.5cm]{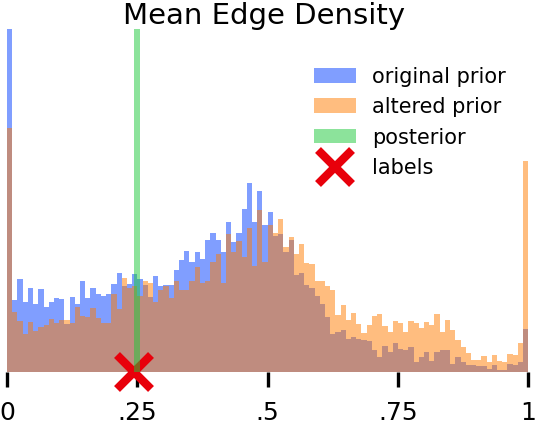} 
  \end{minipage}
  \caption{\emph{Left:} Prior modeling with independently interpretable parameter $\theta$. We run Algorithm~\ref{alg:dpg-iterates} to convergence over discretized $\theta$. \textit{Top}: Larger $\theta$ produces sparser graphs. % cdf which assigns large mass to consistent $\theta$ values. 
  \textit{Bottom}: Larger $\theta$ produces smaller edge weight magnitudes. \emph{Right:} Predictive Checks. %The prior predictive check ensures the average edge densities of replicated data sets encompass plausible data sets. 
  %encompass average edge densities of all data sets we find plausible.
    The original prior generates very few data sets with densities of $\approx .9$, a value we feel is plausible. We can use the independent interpretability of $\theta$ w.r.t. edge density to alter its prior accordingly. A prior predictive check with this altered prior now encompasses these plausible data sets. The posterior predictive check ensures the replicated data sets - now sampled after conditioning on the training data - have similar edge densities to the observed training labels. Indeed, these edge densities, denoted as `posterior', are tightly distributed around the average edge density of the labels.}
  \label{fig:prior_modeling_theta_delta}
\end{figure}

Suppose we have prior beliefs on the sparsity related characteristics of recovered graphs, namely sparsity itself ($=1$ $-$ edge density) or the number of connected components. For example, suppose we believe recovered graphs should have edge densities around $[.05, .5]$ and $\leq 5$ connected components. Since Unrolled DPG approximates solutions to (\ref{eq:smooth-graph-optim-problem-vectorized}) - and $\theta$ determines such sparsity related characteristics independently of other optimization parameters - we can simply run Algorithm~\ref{alg:dpg-iterates} to convergence using the $5$ inputs on a set of discretized $\theta$ values across several orders of magnitude, as demonstrated in Figure~\ref{fig:prior_modeling_theta_delta} (top-left). We observe $\theta \in [10^{-1}, 10^{1}]$ produces solutions with sparsity characteristics consistent with our prior beliefs. Choosing $\theta \sim $ Lognormal($0, 4$) concentrates approximately $75\%$ probability mass on interval $[10^{-1}, 10^{1}]$
while still covering a broader range of values to accommodate uncertainty stemming from sampling variance (small data set) % % from a particular finite iid dataset sampled
and approximation error (truncated iterations).
% FUTURE WORK: Incorporate Theorem 1 from Kalof 2019 which provides an analytical bound on the maximum edge weight recovered. Observing the edge weight distribution
%Weakly informative
Priors are placed over the non-independently interpretable parameters $\delta$ and $b$ with the purpose of guarding against implausible values %, e.g., negative $\delta$, 
and stabilizing posterior sampling; such priors are sometimes called `weakly informative'~\citep{gelman1995BDA}. Because $\delta \Gamma^D_{\theta}(\ve) - b$ will be driven through a sigmoid, we aim for both terms to be of comparable magnitude. Since our graphs are binary, $\delta \Gamma^D_{\theta}(\ve)$ can reasonably be assumed to be $\approx 1$. % reasonable = moderate assumption on convergence of the finite number of iterations, and moderate recovery performance
Utilizing Algorithm~\ref{alg:dpg-iterates} solutions obtained over the $\theta$ grid, we examine the edge weight distribution (prior to scaling by $\delta$) in Figure~\ref{fig:prior_modeling_theta_delta} (bottom-left) to inform $p(\delta)$. Setting $\delta \sim $ Lognormal($2$, $2$) brings scaled edge weights close to $1$ for high (prior) probability values of $\theta$. Similarly, setting $b \sim $ Lognormal($1$, $2$) ensures $b$ is also approximately $1$. Both distributions assign probability mass across several orders of magnitude, both above and below $1$, reflecting the breadth of our uncertainty concerning their specific value ranges. Next, we employ prior predictive checks to validate that - with the established priors - our model generates graphs that align with our prior beliefs.

\subsection{Predictive checking}\label{ssec:pred_checking}
% resources: 
% Viz Bayesian Workflow: https://arxiv.org/pdf/1709.01449.pdf
% STAN prior pred checking: https://mc-stan.org/docs/stan-users-guide/prior-predictive-checks.html#ref-GabryEtAl:2019
% Bayesian Data Analysis: Page 146 middle for a description of how to sample from posterior predictive:

% intro to predictive checking: what is it/its purpose. Why can we do it (generative nature of model)? Why cant typical BNN's do it?
In Bayesian statistics, predictive checks evaluate a model's efficacy in capturing data characteristics like means, standard deviations, and quantiles~\citep{gelman1995BDA}. Predictive checks leverage the existence of a model of the joint distribution $p(\mathcal{T}, \boldsymbol{\Theta})$ between data and parameters. This model allows us to draw samples from the data marginal - called \textit{replicated} data - by drawing samples from the joint and simply dropping the parameter samples. Prior predictive checks draw replicated data from the joint before conditioning on data $\mathcal{T}$, while posterior predictive checks do so after. In both cases, we apply the chosen statistic capturing our data characteristic of interest to the replicated data, producing a histogram. The goal in prior predictive checking is to shape priors that encompass all \textit{plausible} data sets, while still guiding the model towards data sets we deem more \textit{likely} apriori. %w.r.t. prior beliefs. 
Thus prior predictive checking often consists of visual inspection of the histogram to ensure this holds, and adjusting the location and/or scale of prior distributions until it does. The goal in posterior predictive checking is to subjectively validate model fit; here we compare the statistic on the replicated data (a histogram) to the same statistic applied to the actual data $\mathcal{T}$ (a single scalar). A well-fit model should have a histogram tightly concentrated around the real data statistic. 
% Optional: why predictive checks cant usually be used in BNNs

In typical BNNs, the lack of parameter interpretability prevents effective use of predictive checking. \citep{Wenzel2020HowGoodBayesPosterior} highlights this issue by demonstrating that an isotropic Gaussian prior on NN weights (ResNet-$20$) fails to generate data consistent with prior expectations in image classification tasks. But the lack of parameter interpretability prevents the use of these findings to modify the prior for improved alignment. Below, we instantiate predictive checking for DPG, with average edge density as our test statistic. %data characteristic and test statistic.

\textbf{Example (Prior predictive check over DPG parameters)}. % parameter $\theta$)}. %\noindent\textbf{DPG: Prior Predictive Check} 
% More verbose version if needed:
%   Mechanistically, we draw samples from the prior $\boldsymbol{\Theta}^{sim} \sim p(\boldsymbol{\Theta})$ and use them to produce replications $\va^{sim} \sim p(\va|\ve, \boldsymbol{\Theta}^{sim})$, resulting in simulations from the joint $\va^{sim}, \boldsymbol{\Theta}^{sim}) \sim p(\va, \boldsymbol{\Theta}^{sim} | \ve)$. Dropping $\boldsymbol{\Theta}^{sim}$ produces samples from the prior predictive. % this is simply ancestral sampling in fully unobserved tree structured graphical model. See https://mc-stan.org/docs/stan-users-guide/prior-predictive-checks.html for basic argument.
To perform a prior predictive check we use the $5$ inputs from $\mathcal{T}_{\ve}$ and draw $10^4$ replicated data sets $\va^{\text{rep}}$. Figure~\ref{fig:prior_modeling_theta_delta} (right) shows the histogram of average edge densities of replicated data sets under the prior settings laid out in Section~\ref{ssec:prior_modeling}, which we call the `original prior'. We observe that such a prior succeeds in producing data sets that coincide with those we deem more likely apriori, but falls short of encompassing all plausible data sets, namely those with edge densities of $\approx 0.9$. To address this, we lean on independent interpretability of $\theta$ w.r.t. the sparsity of graph solutions: decreasing the location of $p(\theta)$ will increase edge densities (lower sparsity) of graph solutions, and so we lower $p(\theta)$'s  location parameter from $0$ to $-\frac{1}{2}$, with all other parameters of the prior distribution remaining the same. We call this new prior the `altered prior'. Repeating the prior predictive check with this `altered prior' we now observe $\approx 12\%$ of replicated data sets with edge densities in $\left[.75, 1\right]$, as opposed to $4\%$ before, and we confirm through visual inspection that data sets with all possible edge densities are indeed produced. The `altered prior' thus encompasses all plausible data sets while still preferentially generating data sets we feel are more likely apriori. % if a reviewer asks: did this actually help? No, not in this instance, but it shows that we CAN do this with our model, and it proven to be an important modeling technique across many problems.

\textbf{Example (Posterior predictive check over DPG parameters)}. % parameter $\theta$)}. 
Now, we repeat this procedure, but draw replicated data sets from the joint after conditioning on $\mathcal{T}$. We perform inference drawing $M=10^4$ posterior samples; for each posterior sample we draw a single replicated data set. In Figure~\ref{fig:prior_modeling_theta_delta} (right) we plot the histogram of the average edge densities of these replicated data sets, denoted `posterior', and compare against the average edge density of the graph labels in $\mathcal{T}$, denoted `labels'. Because all outcomes are tightly distributed around the mean edge density of the real data, we can have confidence the model parameters have fit appropriately.

%
%%%%%%%%%%%%%%%%%%%%%%%%%%%%%%%%%%%%%%%%%%%%%%%%%%%%

\section{Experiments}
\label{sec:experiments}

%%%%%%%%%%%%%%%%%%%%%%%%%%%%%%%%%%%%%%%%%%%%%%%%%%%%

We have introduced DPG, the first BNN for GSL from smooth signals, which is capable of providing estimates of uncertainty of its edge predictions. We showed DPG can effectively incorporate prior information into the prior distribution over its parameters. In this section, we evaluate DPG across synthetic and real datasets, and introduce other baseline models for comparison, including a more expressive variant of DPG.

\subsection{Models, metrics, and experimental details}\label{ssec:model_metrics_details}
\noindent\textbf{Strict unrollings: DPG and PDS.} In the following experiments, the prior used for the DPG's parameters $\{\theta, \delta, b\}$ is the `altered prior' as developed in Section~\ref{sec:bayesian_modeling}: $\theta \sim $ Lognormal($-1/2, 4$), $\delta \sim $ Lognormal($2$, $2$), and $b \sim $ Lognormal($1$, $2$). We similarly refer to the BNN with Unrolled PDS as its NN model - and no change to the stochastic model - as PDS. As none of PDS's parameters $\{\alpha, \beta, \gamma, b\}$ have independent interpretability, we cannot use the prior modeling techniques outlined in Section~\ref{sec:bayesian_modeling}. 
% gamma poses a problem: divergent transitions in HMC
Attempting HMC on PDS produces significant number of divergent simulated Hamiltonian trajectories, % https://mc-stan.org/docs/reference-manual/divergent-transitions.html#ref-Betancourt:2016b
a phenomenon known to be caused by high posterior curvature~\citep{betancourt2016diagnosing}. This makes HMC run slowly, produce poorly mixed chains, and ultimately non-performant PDS models. Indeed, we find that values of step-size $\gamma$ which produce divergent behavior (very low likelihood) are close those which are most performant (very high likelihood), indicating high-curvature in the likelihood function, and thus the posterior; see Figure~\ref{fig:app-pds-vs-dpg-stability-and-speed} (left). Because $\gamma$ is not interpretable, it is unclear how to shape $p(\gamma)$ to reduce posterior curvature without first running a discrete search. %over all parameters. 
Indeed, to find a performant PDS model amenable to efficient HMC, we first run such a discrete search, fix $\gamma$ to a found performant value of $0.1$, and then set priors $\alpha$, $\beta$ $\sim$ LogNormal($0, 10)$ and $b \sim \mathcal{N}(0, 10^3)$.

\noindent\textbf{Model expansion via MIMO and partial stochasticity: DPG-MIMO and DPG-MIMO-E.}
Here, we expand the Unrolled DPG NN for more expressive power by making each layer multi-input multi-output (MIMO). Each MIMO layer maps inputs $\{\boldsymbol{\lambda}_{k-1} \in \mathbb{R}^{N \times C}$, $\mathbf{a}_{k-1} \in \mathbb{R}^{N(N-1)/2 \times C}\}$ to outputs $\{\boldsymbol{\lambda}_{k} \in \mathbb{R}^{N \times C}$, $\mathbf{a}_{k} \in \mathbb{R}^{N(N-1)/2 \times C}\}$, where $C$ is the fixed number of input and output channels across layers. As before, layers share parameters, now $\boldsymbol{\theta} \in \mathbb{R}^{C \times C}$. In the $k$-th layer, parameter $\theta_{ji} \in \mathbb{R}$ is used to process $\boldsymbol{\lambda}_{k-1}[:,i]$ and $\mathbf{a}_{k-1}[:,i]$ as in a regular DPG iteration. Doing so for $i \in \{1, \dots, C\}$ and averaging the refined $\boldsymbol{\lambda}$'s and $\mathbf{a}$'s % purposefully don't provide notation for this as it will get crazy - need extra dimension which we reduce over.
produces $\boldsymbol{\lambda}_{k}[:,j]$ and $\mathbf{a}_{k}[:,j]$, respectively. %Repeating this for $D$ layers produces $\va_D \in \mathbb{R}^{N(N-1)/2 \times C}$.
We use parameters $\boldsymbol{\delta} \in \mathbb{R}^C$ and $b \in \mathbb{R}$ to produce edge probabilities $\sigma(\frac{1}{C}\va_D \boldsymbol{\delta} - b \mathbf{1})$. The priors are unchanged from the `altered prior' developed in the non-MIMO setting from Section~\ref{sec:bayesian_modeling}. %: $\theta_{ij} \sim $ Lognormal($-1/2, 4$) $\delta_{i} \sim $ Lognormal($2$, $2$), $b \sim $ Lognormal($1$, $2$). 
The increased parameterization and complexity of operations make the posterior geometry significantly more complex, such that we could not find a setup which produced well-mixed posterior sampling via HMC. Instead, we resort to Maximum a Posteriori (MAP) estimation, equivalent to maximizing the posterior with fixed observed data, using full gradient descent.
%Instead, we find a maximum a posteriori (MAP) point estimate by maximizing (\ref{eq:log-joint-factorized}) with data now observed, via full gradient descent. 
We call this non-stochastic setup DPG-MIMO. 

A common approach to overcome the intractable posterior inference in BNNs is partial stochasticity, where we learn point estimates of a subset of the parameters and distributions over the rest. Recent work has shown partial stochasticity of a BNN can produce similarly useful posterior predictive distributions, even outperforming fully stochastic networks in prediction in some setting~\citep{sharma2022bayesianPartialStoch}. 
%Such a partially stochastic BNN can be constructed by first pretraining a point estimator then adding stochastic parameters, or replacing a subset of the parameters with stochastic ones; when the last $N$ layers are made to be stochastic it is denoted a N-Last Layers method~\citep{tutorial}.
Here, we `decapitate` the MAP trained DPG-MIMO by discarding $\boldsymbol{\delta}_{\text{MAP}}$ and $b_{\text{MAP}}$; 
instead we feed each of the $C$ output channels $\va_D[:, j]$ of the base MAP DPG-MIMO (only parameterized by $\boldsymbol{\theta}_{\text{MAP}}$) into a new depth $20$ stochastic single channel DPG $\delta_j\Gamma_{\theta_j}^{20}(\va_D[:, j])$. %, each with their own $\theta$ and $\delta$. 
We average their output, shift by new stochastic $b$, and drive through a sigmoid: $\sigmoid \big(\frac{1}{C}\sum_{j=1}^C \delta_j\Gamma_{\theta_j}^{20}(\va_D[:, j])) - b\mathbf{1}\big)$. %of these $C$ single channel DPG networks 
%and similarly shift by a new stochastic $b$ and drive through a sigmoid. 
We can now run inference on these stochastic head parameters $\{\theta_1, \delta_1, \dots, \theta_C, \delta_C, b\}$ keeping $\boldsymbol{\theta}_{\text{MAP}}$ fixed. We denote this model as DPG-MIMO-E. All MIMO models use $C=4$: $C=8$ offered negligible performance gains and posed (partially stochastic) inference challenges, while $C=2$ was less performant than $C=4$.

\noindent\textbf{Metrics.} To provide summaries of our models' predictive accuracy and quality of uncertainty we use two {proper scoring rules}: %scoring metrics: 
the Negative Log-Likelihood (NLL) $p(\tilde{\va} \mid \tilde{\ve}, \mathcal{T})$ and the Brier Score $\frac{1}{|\mathcal{E}|}\mathbb{E}_{\boldsymbol{\Theta} \mid \mathcal{T}}[\| \hat{\vp}(\tilde{\va} \mid \tilde{\ve}, \boldsymbol{\Theta}) - \tilde{\va}]\|^2)$. % provide a summary measure of a probabilistic model's predictive accuracy.
%Specifically, we consider two proper scoring rules: the Negative Log-Likelihood (NLL) and the Brier Score. 
%The NLL, i.e., the negative log predictive density $-\log p(\va \mid \ve, \boldsymbol{\Theta})$, 
% we intentionally do not use the log posterior density which incorporates prior densities - the priors role is to aid in the estimation of model parameters, but here we are only interested in assessing a model's accuracy in prediction. 
% see "Can You Trust Your Model's Uncertainty" https://proceedings.neurips.cc/paper_files/paper/2019/file/8558cb408c1d76621371888657d2eb1d-Paper.pdf
Beyond {proper} scoring rules, we use Expected Calibration Error (ECE)~\citep{guo2017calibration}, %a popular and intuitive measure of calibration error in neural networks 
which measures the correspondence between predicted probabilities and empirical accuracy, and Error, defined as the percentage disagreement between a thresholded pred. mean {$\mathbb{E}_{\boldsymbol{\Theta} \mid \mathcal{T}}[\tilde{\va} \mid \tilde{\ve}, \mathcal{T}] > 0.5$ and the actual label $\tilde{\va}$.} %the actual label, \(\tilde{\va}\), and the thresholded average output from our NN across posterior samples, \((\frac{1}{M} \sum_{m=1}^M \hat{p}(\tilde{\ve},\boldsymbol{\Theta}^{(m)}) ) > 0.5\). % this removes sampling uncertainty.
See Appendix~\ref{app:experimental_details} for full details.

\noindent\textbf{Hyperparameters and inference details.}
% inference
Utilizing NumPyro's NUTS implementation of HMC~\citep{phan2019numPyro}, our experiments run $4$ chains in parallel%on $T=50$ training samples
, each chain taking $500$ warm-up steps before generating $1000$ samples, accumulating $M=4000$ total samples. We use depth $D=200$ for all models unless otherwise specified, as we did not observe significant improvements in predictive performance beyond this depth. %With $D=30$ this $3$-parameter model's entire inference procedure completes in $1$ minute on a standard m2 Apple laptop. 
% initializing lam_0, w_0 -- pull from appendix
We choose $\va_{0} = \frac{1}{2} \cdot \mathbf{1}$ (reflecting prior uncertainty of existent edges) and $\boldsymbol{\lambda}_{0} = 17 \cdot \mathbf{1}$ (approximate average value of limiting $\boldsymbol{\lambda}$ when running Algorithm~\ref{alg:dpg-iterates} to convergence on RG$_{\frac{1}{3}}$ analytical distance matrices for $\theta = 1$) % \theta = 1 arbitrary
for all experiments. We did not find performance to be sensitive to such choices.
Further details can be found in Appendix~\ref{app:experimental_details}.
% depth

%%%%%%%%%%%%%%%%%%%%%%%%%%%%%%%%%%%%%%%%%%%%%%%%%%%%
\subsection{Synthetic data evaluation}
\begin{figure}
    \centering
    \includegraphics[width=\textwidth, vshift=0cm]{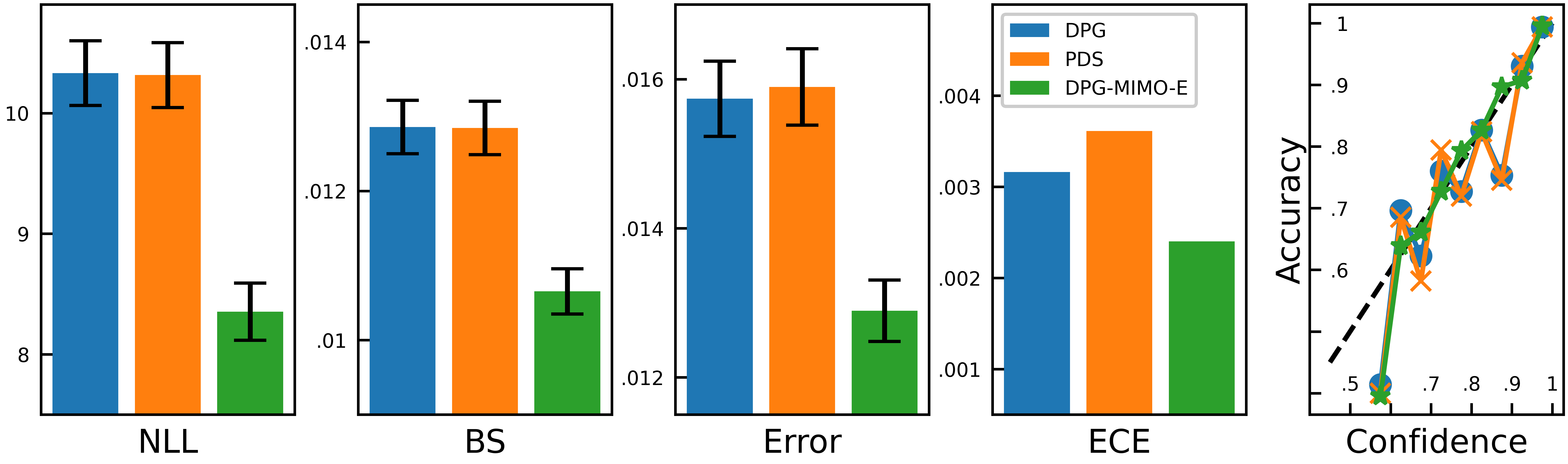} %synthetic_experiments.png}
    \caption{Effective i.i.d. generalization. Both DPG and PDS are performant and well calibrated BNNs, although PDS requires $\approx 3 \times$ more time for inference. Further performance gains are found with the expanded, partially stochastic DPG-MIMO-E model. The rightmost plot (reliability diagram) indicates high calibration across confidence levels; the left-most bin is least calibrated but contains $<0.4\%$ of edges across all models. This plot shows experiments on $N=20$ RG$_{\frac{1}{3}}$ graphs. Error bars are scaled by $.05$ for compact visual effect. 
    %Note in Reliability diagram, i.e. Accuracy vs Confidence, leaves out the distribution of confidence outputs $\hat{p}$: $>99\%$ of edge confidences are in the final two bins encompassing $\left[.9,1\right]$. The least calibrated bin contains $< 0.4\%$ of edges for all models. Error bars are scaled by $.05$ for compact visual effect.
    }
    \label{fig:iid_generalization_evaluation}
\end{figure}

\noindent\textbf{Generative smooth signal model.}
% For detailed argument, see "Synthetic Data" section of https://wandb.ai/max_wasserman/dpg_geom_weighted_hp_search/reports/Graph-Learning-Uncertainty-Estimation--VmlldzozMjcyMjUx
We build on~\citep{dong2016learning}'s work on probabilistic smooth graph signal generation to validate our methods using synthetic data. Let $\mL = \mU \boldsymbol{\Lambda}\mU^{\top}$ be the eigendecomposition of the graph Laplacian $\mL=\textrm{diag}(\mA\mathbf{1})-\mA$, with diagonal eigenvalue matrix $\boldsymbol{\Lambda}$ having associated eigenvalues $\lambda_i$ sorted in increasing order, $\vu_i$ ($i$-th column of $\mU$) is the eigenvector of $\mL$ with eigenvalue $\lambda_i$. \citep{dong2016learning} proposed that smooth signals can be generated from a colored Gaussian distribution as $\vx = \boldsymbol{\mu} + \sum_i \hat{x}_i \vu_i$, where $\hat{x}_i \sim \mathcal{N}(0, \lambda^{\dagger}_i)$ and $\boldsymbol{\mu} \in \mathbb{R}^N$ denotes an arbitrary mean vector. %, where $\mL = \mU \boldsymbol{\Lambda}\mU^{\top}$ is the eigendecomposition of $\mL$ with diagonal eigenvalue matrix $\boldsymbol{\Lambda}$ having associated eigenvalues sorted in increasing order, 
%$, and \dagger$ denotes the pseudoinverse,
Therefore $\vx \sim \mathcal{N}(\boldsymbol{\mu}, \mL^{\dagger})$. To sample such a distribution, it suffices to draw an initial non-smooth signal $\vx_0 \sim \mathcal{N}(\mathbf{0}, \mI)$ and then compute $\vx \leftarrow \boldsymbol{\mu} + \sqrt{\mL^{\dagger}}\vx_0$. % with $h(\mL) = \sqrt{\mL^{\dagger}}$.
%, or equivalently filter it by $h(\lambda) = \sqrt{\lambda^{-1}}$ if $\lambda > 0$ else $0$ and add mean $\bar{\vx}$.
%
% Below results are *novel*.
When $\boldsymbol{\mu} = \mathbf{0}$, this procedure produces signals such that the power on the $i$-th frequency component is $\hat{x}_i^2  \sim \Gamma{(k=\frac{1}{2}, \theta = 2\lambda_i^{\dagger})}$. Thus the expected power $\mathbb{E}[\hat{x}_i^2] = \lambda_i^{\dagger}$ is inversely proportional to the associated eigenvalue, concentrating energy on the low frequencies of the Laplacian spectrum. 
% The above argument requires spread in the eigenvaulue distribution. If most/many eigenvalues are super close together then the filtering won't substantially change the relative frequency composition. E.g. Suppose most/many eigenvalues are close to 0.5. Then a filter of h(\lambda = 1/\lambda is not sharp enough around 0.5 to substantially downweight the frequencies on the larger end of 0.5 vs the smaller end of 0.5.
We construct the data matrix $\mX = [\vx_1, \dots, \vx_P]\in \mathbb{R}^{N \times P}$, where $\vx_p$ are drawn i.i.d. from $\mathcal{N}(\mathbf{0}, \mL^{\dagger})$ as described above. %We will find it useful to work instead with the pairwise distance matrix  $\mE \in \mathbb{R}_+^{n \times n}$, where $E_{ij} := \frac{1}{P}\|\vx^{\top}_i - \vx^{\top}_j\|_2^2$. 
Recalling the definition of the Euclidean distance matrix $\mE$ in Section \ref{sec:graph-learning-smooth-signals}
and using the fact that 
$\bar{\vx}_i^{\top}\bar{\vx}_i \sim \Gamma(k=\frac{P}{2}, \theta=2L^{\dagger}_{ii})$ and $\mathbb{E}[\bar{\vx}_i^{\top}\bar{\vx}_j] = \sum_{p=1}^{P} \mathbb{E}[x_{ip}x_{jp}] = \sum_{p=1}^{P} \text{Cov}(x_{ip}, x_{jp}) =  L^{\dagger}_{ij}, \forall (i,j)$, we can show
%$E_{ij} := \frac{1}{P}\|\mX_i - \mX_j\|_2^2 = \frac{1}{P}(\mX_i^{\top}\mX_i + \mX_j^{\top}\mX_j - 2\mX_i^{\top}\mX_j)$. Using the fact that $\mX_i^{\top}\mX_i \sim \Gamma(k=\frac{P}{2}, \theta=2L^{\dagger}_{ii})$ and $\mathbb{E}[\mX_i^{\top}\mX_j] = \sum_{p=1}^{P} \mathbb{E}[\mX_{ip}\mX_{jp}] = \text{Cov}(\vx^{(i)}_p, \vx^{(j)}_p) =  L^{\dagger}_{ij}, \forall (i,j)$, we can show
%
$\mathbb{E}[\mE] = \mathbf{1} \text{diag}(\mL^{\dagger})^{\top} + \text{diag}(\mL^{\dagger})\mathbf{1}^{\top} - 2\mL^{\dagger}$
which we call the `analytic distance matrix'. We henceforth use $\mathbb{E}[\mE]$ and its finite sample approximations from $P$ signals for evaluation. % approximation = finite signals. Test our model can translate high epistemic uncertainty into higher predictive uncertainty
\begin{figure}[t]
    \centering
    \includegraphics[width=\textwidth, vshift=0cm]{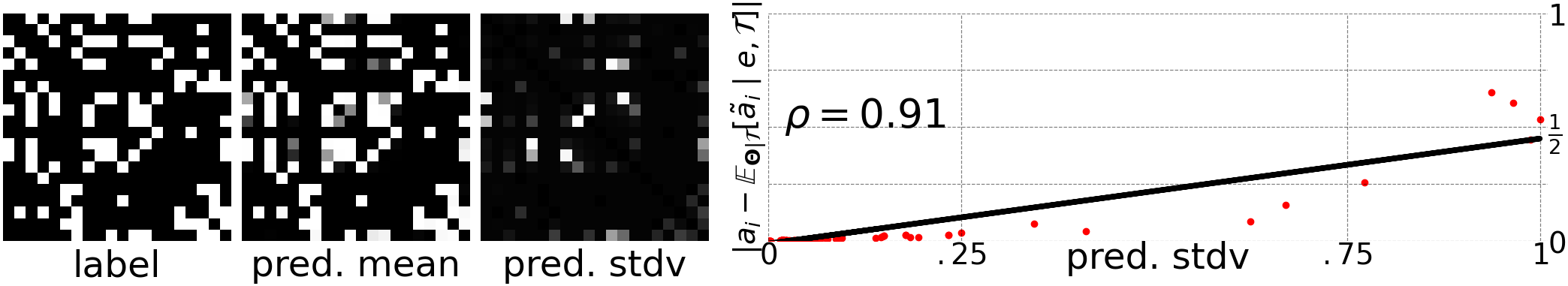}
    \caption{Qualitative i.i.d. generalization. \emph{Left:} For a random test sample we show the label $\tilde{\va}$, and estimated mean (pred. mean) and standard deviation (pred. stdv) of the edge-wise marginal posterior predictive $p(\tilde{a_i} \mid \tilde{\ve},\mathcal{T})$.  %$\mathbb{E}_{\Theta|\mathcal{T}} \left[p(\va|\ve)\right]$
    Comparing pred. mean %of the posterior predictive $\mathbb{E}_{\Theta|\mathcal{T}}[\va^{\text{rep}}|\ve]$ 
    to the label $\tilde{\va}$ adds qualitative evidence that the model is well fit to the data. \emph{Right:} The edge-wise uncertainty estimate (pred. stdv.) and error $|\tilde{a}_i - \mathbb{E}_{\boldsymbol{\Theta}|\mathcal{T}}[\tilde{a}_i|\tilde{\ve}, \mathcal{T}]|$ have a strong positive Pearson correlation $\rho=0.91$. % note: we normalize by max pred. stdv. in a sample for prettier plotting
    %tends to be larger for edges with larger discrepancy between label $a_i$ and mean $\mathbb{E}_{\boldsymbol{\Theta}|\mathcal{T}}[a_i^{\text{rep}}|\ve]$
    %\caption{Comparing the mean of the posterior predictive $\mathbb{E}_{\boldsymbol{\Theta}|\mathcal{T}}[\va^{\text{rep}}]$ to the label $\va$ for a single test sample adds evidence that the model is well fit to the data. The standard deviation, interpreted as the uncertainty, tends to be larger for edges with larger discrepancy between label $a_i$ and mean $\mathbb{E}_{\boldsymbol{\Theta}|\mathcal{T}}[a_i^{\text{rep}}]$.
    %Posterior Predictive: Compute estimates of the mean and standard deviation of the probability of success $\vs$ over all edges in the posterior predictive distribution on test sample $(\ve, \va)$. The standard deviations are normalized to have maximum $= 1$ for visualization. In the rightmost plot observe the model tends to assign higher uncertainty to edges with larger differences between mean and label.
    }
    \label{fig:synthetic_qualitative_eval}
    %\label{fig:predictive_check_and_qualitative_iid}
\end{figure}

\noindent\textbf{Evaluation of i.i.d. generalization.}
Here, we train and test DPG, PDS, and DPG-MIMO-E on $N=20$ RG$_{\frac{1}{3}}$ graphs and their corresponding analytic distance matrices. We use $T=50$ graphs for training and $100$ for testing. We present the results in Figure~\ref{fig:iid_generalization_evaluation}. DPG and PDS perform comparably; although DPG has marginally better accuracy and calibration. %See Appendix~\ref{app:pds} on the difficulties of using Unrolled PDS as a functional model.
Significant improvement is found in the more expressive DPG-MIMO-E across all metrics, showing our simple 3-parameter model can be effectively expanded. These models are similarly performant across graph distributions (see the experiments reported in Table~\ref{tab:distrib_shift_label_mismatch} and also in Appendix~\ref{app:experimental_details}); we show a single graph ensemble here for brevity. 

Figure~\ref{fig:synthetic_qualitative_eval} depicts estimates of the first two moments of the posterior predictive distributions produced by DPG on a random test sample.
% our simple shared $3$-parameter model, 
%$\mu := \mathbb{E}_{\boldsymbol{\Theta}|\mathcal{T}}[y|\ve] = \frac{1}{T} \sum_{t=1}^{T} \hat{\vp}(\ve; \boldsymbol{\Theta}^{(t)})$ and $\sigma^2 := \mathbb{E}_{\boldsymbol{\Theta}|\mathcal{T}}[(y-\mu)^2|\ve] = \frac{1}{T} \sum_{t=1}^{T} (\hat{\vp}(\ve; \boldsymbol{\Theta}^{(t)}) - \mu)^2$
We observe well-calibrated confidence and uncertainty, plus a strong correlation between error magnitude and uncertainty; indeed a useful uncertainty estimate should be a proxy for error.
% correlation coefficients between error and uncertainty across 5 graphs mean: 0.77, std: 0.17. Raw values: [0.8316975831985474, 0.5065945386886597, 0.991051197052002, 0.8892071843147278, 0.6469759941101074]
%

\begin{figure}
    \centering
    \includegraphics[width=\textwidth, vshift=0cm]{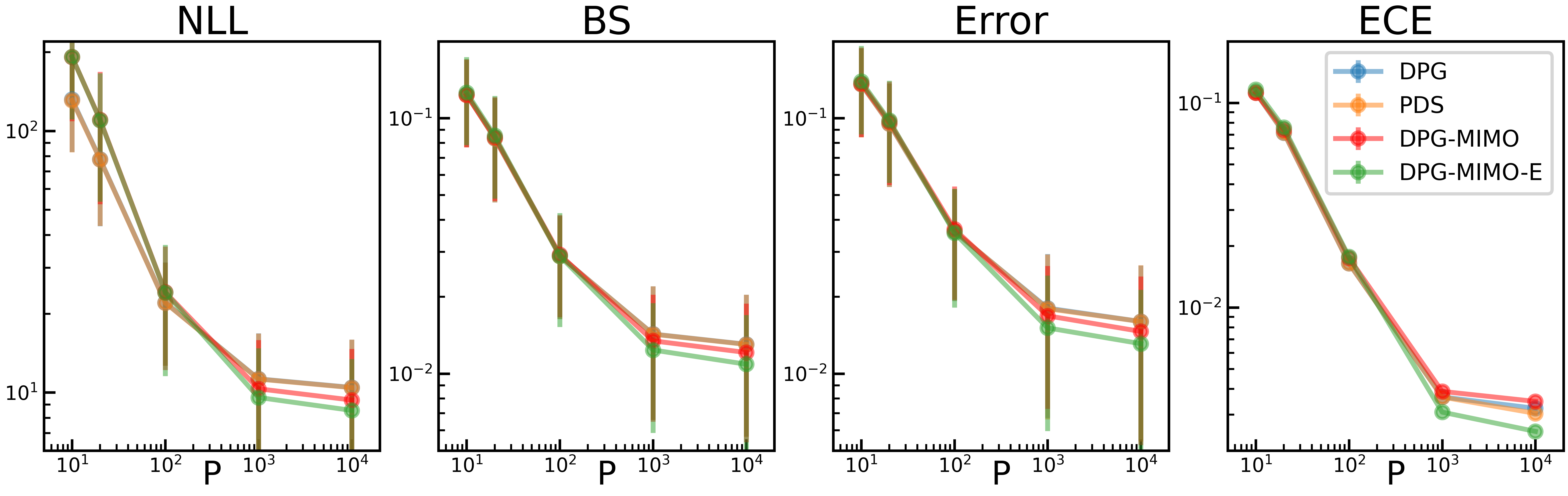}
    \caption{Detection of covariate shift across models. Using RG$_{\frac{1}{3}}$ graphs with $N=20$ nodes, we %train BNNs with varying functional models 
    fit models on analytic Euclidean distance matrices and evaluate on corrupted distance matrices in a test set. Fewer signals $P$ used tends to increase corruption magnitude. Log-scaling unintentionally makes error bars appear to grow with $P$.
    }
    \label{fig:finite_signal_covariate_shift_evaluation}
\end{figure}

\noindent\textbf{Predictive uncertainty under distribution shift.}
% more detailed background on distribution shift and OOD
%   - OOD~\citep{laksminarayanan et al 2017}, covariate shift~\citep{ovadia2019can}
%   - this definition of OOD allows us to use the same metrics, whereas typically because no output which corresponds to this class is present, typically use (i) histograms of confidence and predictive entropy on known vs OOD inputs and (ii) accuracy vs confidence plots. See https://proceedings.neurips.cc/paper_files/paper/2019/file/8558cb408c1d76621371888657d2eb1d-Paper.pdf right above Section 4.
%Distribution shift, or \textit{covariate shift}~\citep{quinonero2008datasetShift}, is when the test label $\va$ stays the same, but the test input $\ve$ is shifted, often by corruptions or perturbations~\citep{hendrycks2019benchmarking}. Out-of-Distribution (OOD) inputs - typically used in the context of $k$-class classification - refers to the case where the ground truth label is not one of the $k$-classes. Instead, we will use this to refer to the case of the label $\va$ being generated from a different synthetic graph ensemble than that used for training. \citep{ovadia2019can} argues evaluation of predictive uncertainty is most meaningful using distribution shift due to the failure of post-hoc calibration under mild input shift and the prevalence of such shifts - and the need to detect them - in real world applications. 
We evaluate two forms of distribution shift: corruptions to the noiseless analytic distance matrix $\mathbb{E}[\mE]$  %via finite sampled signals $P$ (thus far we have used the 'analytic distance matrix') 
and label mismatch.
To investigate corruptions, we train using $T=50$ RG$_{\frac{1}{3}}$ ($N=20$) labels  and their %corresponding input being the \textit
{analytic distance matrix}, % from Section~\ref{sec:graph-learning-smooth-signals}, 
and test on $100$ RG$_{\frac{1}{3}}$ samples which instead use $P$ signals to compute an Euclidean distance matrix $\mE$. Fewer samples $P$ tend to produce larger corruption magnitudes, i.e., deviations from  $\mathbb{E}[\mE]$. %\textit{observed distance matrix} Corruption magnitude is increased by decreasing the number of signals $P$ used in estimation of the distance matrix.  
Figure~\ref{fig:finite_signal_covariate_shift_evaluation} shows that indeed, all models display lower predictive accuracy and higher uncertainty on the increasingly shifted input data, with the simpler models outperforming in the presence of larger corruption. 
% Label Mismatch
%For label mismatch, we evaluate on a different random graph ensemble than was trained on.
To investigate label mismatch, models are fit to the same training data as above - $T=50$  RG$_{\frac{1}{3}}$ ($N=20$) labels with analytic distance matrix input - but now tested on $100$ test samples from the following \textit{different} $N=20$ random graph distributions: (i) random geometric graphs with connectivity radius \(\frac{1}{2}\) (RG$_{\frac{1}{2}}$), Erd\H{o}s-R\'enyi graphs with edge probability \(p=0.5\) (ER$_{\frac{1}{2}}$), and Barab\'asi-Albert graphs with \(m=1\) link per node (BA$_1$). Results are displayed in Table~\ref{tab:distrib_shift_label_mismatch}. All models show lower accuracy ($\uparrow$ Brier Score) and commensurately higher uncertainty ($\uparrow$ NLL) on these mismatched data. DPG methods %, even the MAP-based DPG-MIMO,
tend to maintain better calibration than PDS. % - i.e. correctly knowing when it is wrong - on mismatched data.
\begin{table}[h]
% run file src/visualization/comparison/ood_detection.py
    \centering
    \caption{Detection of label mismatch. Models are fit to $\text{RG}_{\frac{1}{3}}$ and tested on $100$ samples from the same RG$_{\frac{1}{3}}$ as well as three \textit{different} graph distributions. For each metric, the mean over the test set is reported. %: $\text{RG}^*_{\frac{1}{2}}$ which differs only in connectivity radius r, $\text{ER}_{\frac{1}{2}}$, and $\text{BA}_{1}$. 
    %All models effectively detect graphs from outside its training distribution.
    }
    \label{tab:distrib_shift_label_mismatch}
    \setlength{\tabcolsep}{5.45pt} % Reduce space between columns
    \begin{tabular}{@{}cccccccccccccc@{}} % Removed extra space between columns
         \toprule
         & \multicolumn{4}{c}{NLL} 
         & \multicolumn{4}{c}{BS (\si{\times 10^{-2}})} 
         %& \multicolumn{4}{c}{Error ($\%$)} 
         & \multicolumn{4}{c}{ECE (\si{\times 10^{-3}})} \\
         \cmidrule(lr){2-5}\cmidrule(lr){6-9}\cmidrule(lr){10-13}
         %\cline{2-13}
         & RG$_{\frac{1}{3}}$ & $\text{RG}_{\frac{1}{2}}$ & $\text{ER}_{\frac{1}{2}}$ & $\text{BA}_{1}$
         & RG$_{\frac{1}{3}}$ & $\text{RG}_{\frac{1}{2}}$ & $\text{ER}_{\frac{1}{2}}$ & $\text{BA}_{1}$
         %& RG$_{\frac{1}{3}}$ & $\text{RG}_{\frac{1}{2}}$ & $\text{ER}$ & $\text{BA}$
         & RG$_{\frac{1}{3}}$ & $\text{RG}_{\frac{1}{2}}$ & $\text{ER}_{\frac{1}{2}}$ & $\text{BA}_{1}$\\ 
         \hline
        DPG 
        & 10.33 & 17.53 &  33.01 & 20.91
        & 1.29  &  2.29 &  4.66  & 2.10
        %& 1.57  &  1.90 &  0.51  & 2.22
        & 3.42  & 52.06 & 171.38 & 16.60 \\
        PDS 
        & 10.32 & 17.70 &  33.43 & 20.80
        & 1.29  &  2.33 &   4.80 &  2.10
        %& 1.60  &  1.93 &   0.56 &  2.22
        & 3.75  & 52.87 & 173.53 & 16.60\\
        DPG-MIMO 
        & 9.19 &  17.13 &   40.49 &  13.37
        & 1.19 &  1.74  &   4.62  &  2.06
        %& 1.42 &  1.33  &   1.88  &  2.71
        & 3.50 & 54.61  & 161.71  & 11.24\\
        DPG-MIMO-E 
        & 8.35 & 12.66 & 37.77 & 8.35
        & 1.07 &  1.52 &  7.82 & 1.32
        %& 1.29 &  1.57 &  1.20 & 1.61
        & 2.40 & 29.17 & 29.07 & 2.84 \\ 
        \hline
    \end{tabular}
\end{table}

\noindent\textbf{Scaling to larger graphs.} 
%% Direct inference on large graphs hard
Direct inference in large graph settings is challenged by substantial memory demands. Specifically, providing gradients for HMC via naive backpropagation necessitates storing all intermediate outputs, while full Bayesian inference mandates holding the full data set in memory, cumulatively leading to a memory footprint of \(\mathcal{O}(DN^2T)\). Adopting forward-mode auto-differentiation partially mitigates this issue by removing depth dependency, effectively reducing memory complexity to \(\mathcal{O}(N^2T)\). The time complexity for MCMC also presents scalability challenges: each forward pass has time complexity of \(\mathcal{O}(DN^2T)\), and each posterior sample requires multiple forward passes.
%% Can we instead perform transfer learning? Directly, no.
Attempting to instead perform transfer learning -- where inference occurs with smaller graphs to then test on larger graphs -- still faces an obstacle. The objective function (\ref{eq:smooth-graph-optim-problem-vectorized}) contains the terms $2\va^{\top}\ve = 2 \sum_{i=1}^{N(N-1)/2} a_i e_i$, $\frac{\beta}{2} \|\va\|^2_2 = \frac{\beta}{2} \sum_{i=1}^{N(N-1)/2} a_i^2$, and  $- \alpha \mathbf{1}^{\top} \text{log}(\mS \va) = - \alpha \sum_{i=1}^N \text{log}(\mS \va)_i$. The former two terms involve $N(N-1)/2$ summands, while the latter is a sum over $N$ elements. 
%$2\va^{\top}\ve$ and $\frac{\beta}{2} \|\va\|^2_2$, each term composed of $N(N-1)/2$ individual elements, and term $- \alpha \mathbf{1}^{\top} \text{log}(\mS \va)$ comprises only $N$ elements. 
This results in disparate growth rates for these terms as $N$ increases, which poses a challenge for parameter optimization across different graph sizes, i.e., we find parameters optimized for a graph with $N_i$ nodes are not effective for $N >> N_i$.
%$2\va^{\top}\ve$ and $\frac{\beta}{2} \|\va\|^2_2$ have $N(N-1)/2$ components, while $- \alpha \mathbf{1}^{\top} \text{log}(\mS \va)$ has $N$ components. This discrepancy means parameters optimized for a graph with $N_i$ nodes are not effective for $N >> N_i$. 
Moreover, the differing growth rates cause standard cardinality-based normalization schemes (dividing each term by its number summands) to fail, and determining analytically how the parameters should change as a function of size is non-trivial; see Appendix~\ref{app:scaling} for a detailed discussion. % Mention that we believe this is why previous works have failed?
%% Instead, altered transfer learning experiment

In light of this, we perform an adjusted transfer learning experiment by fitting the empirical growth trends of MAP DPG parameter estimates on $N=\{20, 50, 100, 200\}$ ER$_{\frac{1}{4}}$ graphs and their corresponding analytic distance matrices. By using a moderate depth $D=200$ and $T=10$ training samples, MAP inference on larger graphs with $N=200$ nodes takes $\sim1$ hour on a M$2$ MacBook laptop.  Using this empirical parameter fit we can extrapolate parameter values and perform transfer learning on $100$ ER$_{\frac{1}{4}}$ test graphs of size up to $N=1000$; again done locally without any GPU. % we can also emphasize that once fit, prediction is very fast; we inherit this characteristic from NN's 
The transfer learning experiment reveals an expected but graceful decay in performance in NLL as $N$ increases. % why? partially due to the imperfect empirical fit to the parameters growth trends. 
Further information and plots on these scaling experiments are available in Appendix~\ref{app:scaling}.
%Some numerical results which we can include if needed. This is all in appendix as well:
%   - Our results indicate an improvement in performance with larger graph sizes, with mean NLL (normalizing for number of edges) decreasing monotonically from $3.5e-2 (\pm 2.8e-2)$ for $N=20$ to $2.1e-6 (\pm 1.9e-6)$ for $N=200$. Improved performance with a larger dimension is explained by the analytic distance matrix becoming smoother with increasing size.
%
\subsection{Ablation studies}
\label{subsec:ablation}
\noindent\textbf{Prior modeling.} %DPG with and without prior modeling
% See chatgpt - "Ablation: DPG Prior Modeling", created March 28, 2023 for formatted data.
To inspect the influence of informative prior modeling on DPG, we define model `DPG-U' which replaces DPG's informative priors with the following uninformative priors: log $\theta \sim U\left[10^{-6}, 10^{6}\right]$, $\delta, b \sim \mathcal{N}(\mathbf{0}, 10^{3}\mathbf{I})$. What we find is that in data-rich regimes an informative prior mostly acts to improve efficiency of posterior inference, e.g., when fitting to $T=50$ RG$_{\frac{1}{3}}$ graphs with $N=20$ nodes (using analytic distance matrices) and testing on $100$ i.i.d. samples, it reduces the time needed to generate $M=4000$ samples by a factor of $7.8\times$ %\frac{343.97}{44.15}$
and increases the effective sample size - a measure of the number of independent samples with the same estimation power as the observed correlated samples~\citep{gelman1995BDA} - for $\theta$ (ESS$_{\theta}$) by a factor of $5\times$. %;\frac{19.7}{3.9}$
%an expected result given an informative prior's helpful influence on posterior geometry~\citep{}. 
In data poor regimes it also tends to help slow down performance degradation, e.g., when instead using $T=2$ with the same graph data%$N=20$ RG$_{\frac{1}{3}}$ samples
, NLL and ECE are 
$37\%$  %(17.06 - 10.66)/17.06
and $13.7\%$ % (8.75 - 7.55)/8.75
better with the prior than without. See Table~\ref{tab:ablation-prior} in Appendix \ref{app:further_experiments} for the full numerical results.

\noindent\textbf{Partial stochasticity.} % this is simply the last 2 rows of the table with RG_{1/3} graphs; as the percentage reduction matches
%%%% Raw data
%% partially stochastic model: lower NLL ($8.354 \pm 4.738$), a smaller BS ($0.01065 \pm 0.00604$), a lower error rate ($0.01289 \pm 0.00823$), and a lower ECE ($0.00240$) 
%% MAP model: NLL $9.191 \pm 5.155$, BS $0.01188 \pm 0.00665$, Error $0.01416 \pm 0.00941$, ECE $0.00350$ 
We conducted an ablation study investigating the benefits of adding partial stochasticity to the DPG-MIMO model ($C=4, D=200$) using MAP point estimates; the partially stochastic setup is described in Section~\ref{ssec:model_metrics_details}. The training and test sets are identical to the i.i.d. generalization experiment above. 
%The partially stochastic model involves removing the $\delta$ and $b$ terms and running stochastic single channel dpg unrollings ($D=20$) on the raw output channels of the MAP mimo base. 
Our experiments showed that introducing partial stochasticity in DPG-MIMO-E markedly improved NLL, BS, error, and ECE by $9.1\%$, $10.3\%$, $9.0\%$, and $31.4\%$, respectively.

\subsection{Real data evaluation}
% Ideas
% - Financial Data. See data from Viniciuis from NeurIPS 
% - Connectivity Across Classes of MNIST Idea: node = MNIST image. edge = euclidean distance between vectorized mnist images. True graph: connect edges between nodes representing the same number. Show good recovery and higher uncertainty over edges between similar numbers.
% - Citation Networks. Subsample ego graphs (or some other way as long as they are connected) from Cora/Citeseer (may have to be careful to get all classes in a single sample) to create a dataset.
% - Gene Networks. Use data setup from GLAD paper. Synthetic data generator so can control number of samples. Downside: still kind of synthetic data.
% - Point Cloud Data. Given point cloud (possibly subsample) compute adjacency A. Each node also has a 3D position. Compute the euclidean distance between all points forming E. Map from E back to A. Downside: Very high dimensional, will be hard to do any sort of posterior sampling. Could be more useful for a scalable follow up work.
% - Manifold Recovery. See Kalofolias "LARGE SCALE GRAPH LEARNING FROM SMOOTH SIGNALS", ICLR 2019.
%
% - \newline\noindent\textbf{Brain Graphs}:show overall fit and distribution shift abilities. Pose it as anomoly detection, e.g. poor imaging.
%   idea: split data, estimate model parameter with clean data (subjects containing all 4 scans), evaluate how NLL changes on test set using 1, 2, 3, and 4 scans. Hope to see NLL be higher for patients with less data.
%  0.495
\begin{figure}[t]
\centering
  \begin{minipage}[b]{0.51\textwidth}
    \includegraphics[width=\textwidth, vshift=0.005cm]{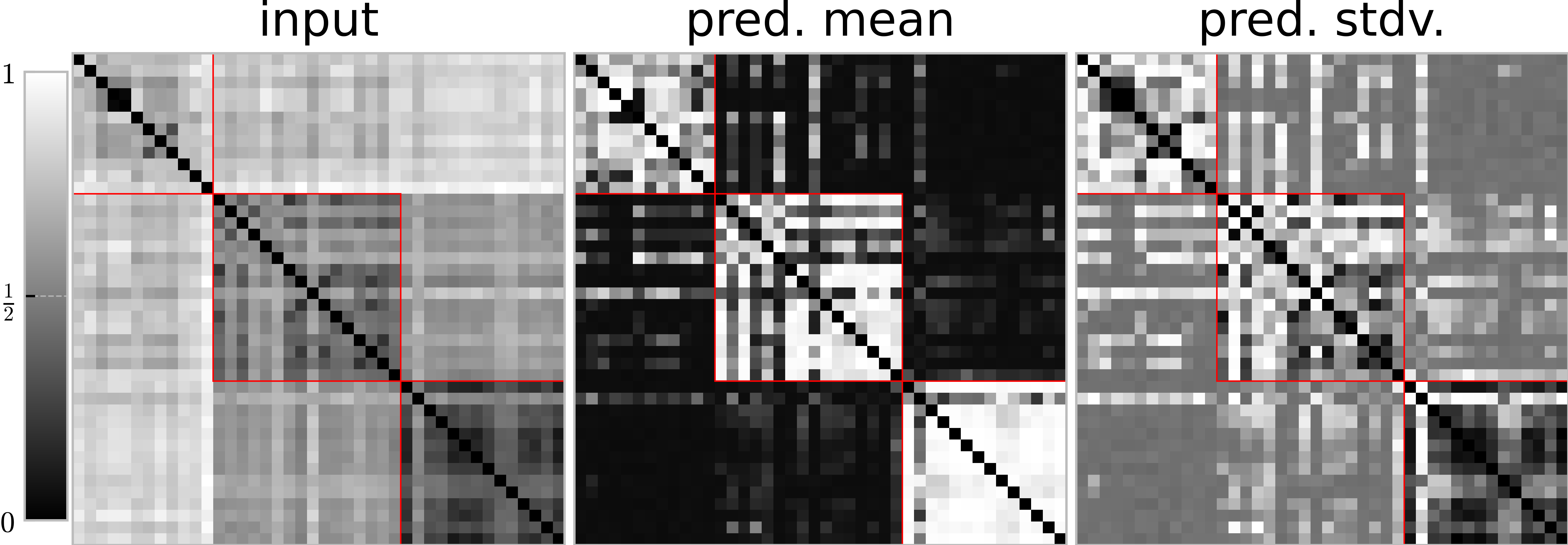}
  \end{minipage}
  \hfill
  \begin{minipage}[b]{0.4825\textwidth}
    \includegraphics[width=\textwidth, vshift=0cm]{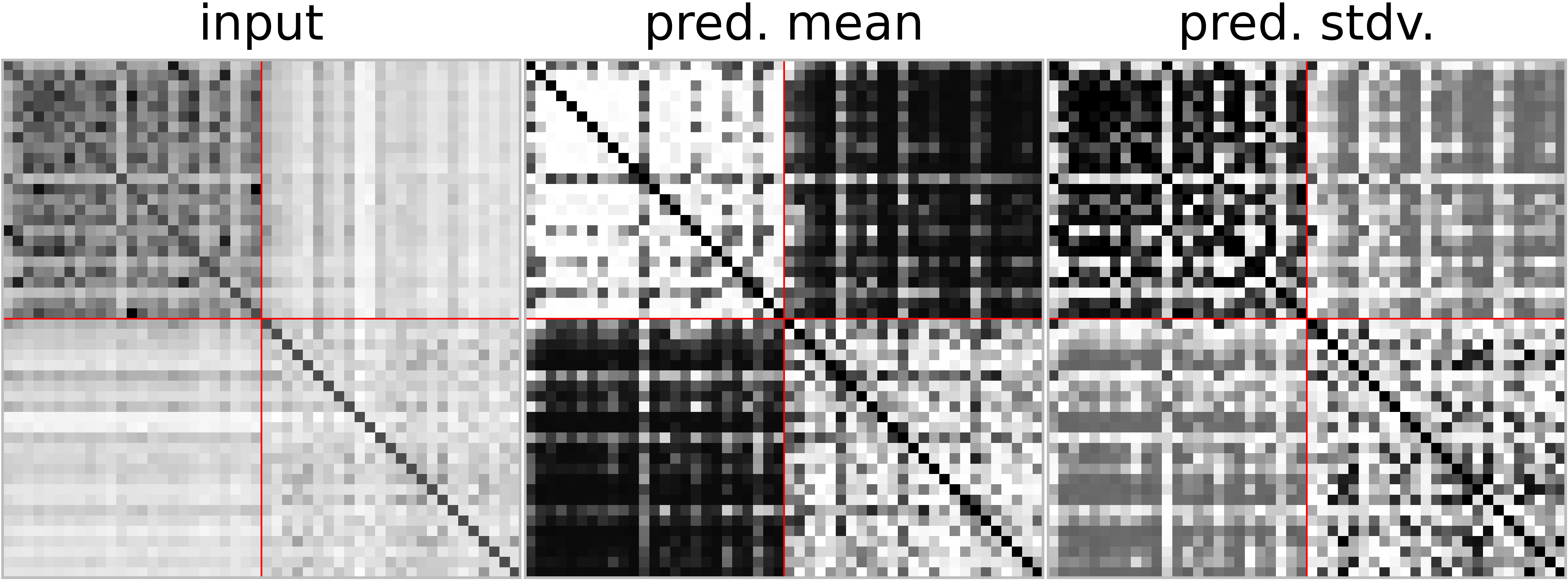}
  \end{minipage}
  \caption{Quantifying Uncertainty with Financial Data and Images of Digits. 
  % SP500
  \textit{Left}: S\&P$500$. We randomly split stocks from $3$ sectors of the S\&P$500$ and compute their input $\mathbf{1}\mathbf{1}^\top-|$\boldsymbol{$\Sigma$}$|$, where \boldsymbol{$\Sigma$} is the matrix of Pearson correlation coefficients, using log daily returns over an extended period. The graph label connects stocks of the same sector indicated by red block diagonal outline. There is only a single train and test sample. Here, we display the test sample input and estimates of the mean (pred. mean) and standard deviation (pred. stdv.) of the posterior predictive. In pred. mean, white (black) inside (outside) the block diagonal indicates correct prediction for both experiments.  Pred. mean which don't match their label tend to have larger uncertainty. Error and pred. stdv. have a Pearson correlation of $0.70$ over the test set. 
  % MNIST
  \textit{Right}: MNIST digits. We construct graphs of MNIST digits "$1$" and "$2$", connecting nodes of the same digit, similarly outlined in red ("$1$"s are the first block). The input is the log pairwise Euclidean distance of their vectorized image. Notice "$1$"s tend to be much closer to each other than "$2$"s. DPG again shows an ability to place higher uncertainty on edges with higher error. Error and pred. stdv. have a Pearson correlation of $0.62$ over the test set. Pred. stdv. in both plots are maximum normalized for visual clarity.
  }
  \label{fig:stocks_mnist}
\end{figure}
\noindent\textbf{Quantifying uncertainty in networks learnt from S\&P$500$ stock prices.}
% Data and experimental setup taken directly from "Graphical Models in Heavy-Tailed Markets", José Vinícius de M. Cardoso, Jiaxi Ying, Daniel P. Palomar
%%vIdeas for further experimentation
% 1. Inspect uncertainty using different amounts of test data. Do we see more uncertainty with smaller window of data?
%% Ideas for further discussion
% 1. We use correlation matrix as is done in "Graphical Models in Heavy-Tailed Markets". Works better than euclidean distance and more faithful to what is typically done in financial works. We can discuss that the method works with things other than euclidean distance.
To verify that DPG provides useful measures of uncertainty on edge predictions based on real data, we first use the financial dataset presented in~\citep{de2021graphical}. The time series consist of S\&P$500$ daily stock prices from three sectors (Communication Services, Utilities, and Real Estate), comprising a total of $N = 82$ stocks. The data spans the period from Jan. $3$rd $2014$ to Dec. $29$th $2017$, yielding $P = 1006$ daily observations. We divided these stocks equally into training and testing sets. For both sets, we created a log-returns data matrix, denoted as $\mX \in \mathbb{R}^{N/2\times P}$. 
% already explained what N and P are
%, with $P$ representing the number of log-return observations and $N$ the number of stocks. 
Specifically, $X_{i,j} = \text{log }SP_{i,j} - \text{log }SP_{i,j-1}$, where $SP_{i,j}$ signifies the closing price of the $i$-th stock on the $j$-th day. From each matrix (train and test), we derived a matrix of Pearson correlation coefficients \boldsymbol{$\Sigma$}. We take the input to be $\mE \leftarrow \mathbf{1}\mathbf{1}^\top-|$\boldsymbol{$\Sigma$}$|$, thus yielding measure of dissimilarity. We also constructed a binary label graph, assigning an edge weight value of $1$ for stocks within the same sector and $0$ for pairs in different sectors. Thus, our training and testing datasets each contain a single sample. Nodes from the same sector are numbered contiguously creating a block diagonal adjacency matrix structure, displayed with the red outline in Figure~\ref{fig:stocks_mnist} (left). 

We use DPG with depth $D=200$ and identify a high-performing parameter value range for $\theta$ via predictive checking as in Section~\ref{ssec:pred_checking}. Such values aligned closely with those found in the synthetic experiments, and so we keep the priors unchanged. We run Bayesian parameter inference on a M$2$ MacBook laptop which takes about $2$ minutes; we visualize the input, empirical mean, and standard deviation of the posterior predictive on the test sample in Figure~\ref{fig:stocks_mnist} (left).
% Note that sector categorizations aren't absolute; some observed 'errors' could be due to related companies being categorized differently.
Notably, while our mean recovery is robust, there are areas where it deviates from the label. These areas tend to exhibit increased variation in the posterior predictive. The Pearson correlation coefficient between the error and predictive standard deviation on all edges, true positive edges, and true negative edges is $0.70$, $0.70$, and $0.79$, respectively, suggests that our model's uncertainty can be a useful proxy for error in the absence of labels in this financial test case.

%
%\newline\noindent\textbf{Quantifying Uncertainty in Foreign Exchange Markets}
%

\noindent\textbf{Learning the graph of MNIST digits 1 and 2.}
% the central claim we are supporting: we can perform GSL with uncertainty quantification
To further demonstrate DPG's capability to quantify uncertainty on graphs estimated from images of digits, we emulate the experiment from~\citep{Kalof2016Smooth} which learns a graph to classify handwritten digits ``$1$'' and ``$2$'' from the MNIST database~\citep{lecun1998MNIST}. Each digit is an image of $28 \times 28$ pixels, each pixel taking integer value in $0, \dots, 255$. As~\citep{Kalof2016Smooth} reports, this problem is particular because digits "$1$" are much close to each other than "$2$" are (average square distance of $45$ and $102$, respectively). 
%% Experimental Set-up
% label = 1 if two images are of same digit, 0 otherwise. input = log(euclidean distance).
% 25 1's, 25 2's per sample. Order then such that all one nodes are first (produces block diagonal label).
% 5 training samples, 50 testing samples.
% training takes 10 minutes on local m2 laptop
% main takeaway: we successfully capture uncertainty. Visually we can see areas where pred. mean. disagrees with label, the pred. stdv. tends to be high. We can also measure across entire test set, and find error and pred. stdv. are significantly correlation (0.622071419156854)
A single sample in our dataset is constructed as follows. We randomly sample $25$ images of digits ``$1$'' and ``$2$''. Each image represents a node in this graph sample, hence $N=50$. We connect nodes of the same digit with a binary edge. The image corresponding to each node is vectorized into  $\bar{\vx}_i\in\mathbb{R}^{28^2}$, $i=1,\ldots N$, becoming the nodal feature vector. For ease of label visualization, we order the nodes such that all ``$1$'' digits come before ``$2$'' digits, which creates a block diagonal adjacency matrix, indicated by the red outline in Figure~\ref{fig:stocks_mnist} (right). The nodal dissimilarity matrix $\mE$ is computed as the log pairwise Euclidean distance of the node features (vectorized digit images). We construct a training and test set of size $T=5$ and $50$, respectively. 

We use the same modeling and inference setup as in the previous real data experiment; inference takes $\approx 10$ minutes. Figure~\ref{fig:stocks_mnist} (right) displays the input $\mE$, pred. mean, and pred. stdv.
%estimates of the mean and standard deviation of the edge-wise marginals of the posterioour point and uncertainty estimate of t he posterior predictive  our point estimate ofempirical mean, and standard deviation of the predictive posterior 
on a random test sample. Visual inspection reveals performant mean prediction, and that edges with high errors tend to have high variation. %pred. stdv. 
Indeed, across the entire test set the two have a strong positive Pearson correlation coefficient over all edges, true positive edges, and true negative edges of $0.62$, $0.72$, and $0.51$, respectively, suggesting DPG effectively quantifies predictive uncertainty of edges in this image analysis setting.
%
%%%%%%%%%%%%%%%%%%%%%%%%%%%%%%%%%%%%%%%%%%%%%%%%%%%%

%
\section{Concluding summary, limitations, and the road ahead}

%%%%%%%%%%%%%%%%%%%%%%%%%%%%%%%%%%%%%%%%%%%%%%%%%%%%

% explains how the results have filled the gap that was identified in the introduction, provides caveats to the interpretation, and describes how the paper advances the field by providing new opportunities. 
% typically done by recapitulating the results, discussing the limitations, and then revealing how the central contribution may catalyze future progress. 

In this work we identify independent interpretability of (regularization) parameters in inverse  problems as a key property, which allows one to incorporate prior information on solution characteristics into prior distributions over the NN parameters of an unrolled optimization algorithm. %into the values of optimization weights, and subsequently the neural network parameters of an unrolling of it's solution iterations. 
We investigate an inverse problem with this independent interpretability property in the context of GSL from smooth signals, and introduce an optimization algorithm that is simpler than existing methods, without sacrificing estimation performance. We unroll this algorithm producing the first true NN for GSL from smooth signals. We leverage independent interpretability to incorporate prior information about sparsity characteristics of sought graphs into prior distributions over unrolled network parameters, producing the first BNN for GSL from smooth signal observations. We lean into the advantages of unrollings - low parameter dimensionality, fast compiler and auto-grad friendly layers, as well as the empirical need for shallow networks - to perform posterior approximation using MCMC sampling. Doing so yields high-quality and well-calibrated uncertainty estimates over edge predictions, as demonstrated via comprehensive experiments with synthetic and real datasets.%, and showed its ability to effectively quantify uncertainty in edge predictions.
%We introduce a method to incorporate prior information over the inverse graph solution into prior distributions over the ...We introduce novel GSL iterations for an inverse problem with an independently interpretability and outline show to leverage this property to incorporate prior information about solutions into prior distributions over network parameters, producing the first BNN for GSL from smooth signal observations. We take advantage of the control on complexity unrollings provide to perform MCMC sampling on it's parameters. Doing so allows estimates of uncertainty on edge predictions. We validated this approach on synthetic and real data, and showed it's ability to effectively capture uncertainty in edge predictions.

The primary limitation of our approach lies in its scalability to moderate- and large-sized graphs and datasets. The computational and memory requirements to achieve asymptotically exact posterior inference through MCMC sampling are substantial, owing to the necessity for numerous forward network passes per posterior sample and the requirement to store the entire data set in memory. Promising directions of future work include exploration of non-asymptotically exact posterior inference methods, namely variational inference, which typically requires much less computation and allows for mini-batch training and thus use of larger datasets~\citep{jospin2022BNNTutorial}. As we produce distributions over output graphs, future work can explore Bayesian decision analysis with the introduction of a utility function, or even using DPG as a module within a larger system, where propagation of uncertainty over the inferred graph is important.

\bibliography{tmlr/main}
\bibliographystyle{tmlr}

\appendix

%%%%%%%%%%%%%%%%%%%%%%%%%%%%%%%%%%%%%%%%%%%%%%%%%%%%

\section{Appendix}

%%%%%%%%%%%%%%%%%%%%%%%%%%%%%%%%%%%%%%%%%%%%%%%%%%%%

\subsection{Optimization algorithms for model-based GSL methods}

%\subsubsection{Derivation of the dual-based proximal gradient (DPG) algorithm}\label{app:dpg}

\subsubsection{Primal-dual splitting (PDS) algorithm}\label{app:pds}
Algorithm~\ref{alg:pds-iterates} was introduced in~\citep{Kalof2016Smooth} to solve the ($\alpha, \beta$)-parameterization of (\ref{eq:smooth-graph-optim-problem-vectorized}), and was subsequently unrolled by~\citep{Pu2021L2L} on the same objective. Note the increased number of operations, intermediate variables, and parameters as compared to Algorithm~\ref{alg:dpg-iterates}.

\begin{algorithm}
\caption{Proximal Dual Splitting (PDS)}
\label{alg:pds-iterates}
\begin{algorithmic}
\STATE \textbf{Inputs}: Fixed parameters $\alpha, \beta, \gamma \in \mathbb{R}$, and data $\ve$.
\STATE \textbf{Initialize}: $\va_0$ and $\vv_0$ at random.
\FOR{$k = 1, 2, \dots$}
\STATE $\vr_{1,k} = \va_k - \gamma (2 \beta \va_k + 2\ve + \mS^{\top} \vv_k)$.
\STATE $\vr_{2,k} = \vv_k + \gamma \mS \va_k$.
\STATE $\vp_{1, k} = \text{prox}_{\gamma, \Omega_1}(\vr_{1,k})$, where $\text{prox}_{\gamma, \Omega_1}(\vr_{1,k}) = \max\{0, \vr_{1,k}\}$.
\STATE $\vp_{2, k} = \text{prox}_{\gamma, \Omega_2}(\vr_{2,k})$, where $\Bigl(\text{prox}_{\gamma, \Omega_2}(\vr_{2,k})\Bigr)_i = (r_{2_i}-\sqrt{r_{2_i}^2 + 4\alpha\gamma})/2$.
\STATE $\vq_{1,k} = \vp_{1,k} - \gamma(2\beta\vp_{1,k} + 2\ve + \mS^{\top}\vp_{2,k})$.
\STATE $\vq_{2,k} = \vp_{2,k} - \gamma\mS\vp_{1,k}$.
\STATE $\va_{k+1} = \va_k - \vr_{1,k} + \vq_{1,k}$.
\STATE $\vv_{k+1} = \vv_k - \vr_{2,k} + \vq_{2,k}$.
\ENDFOR
\STATE \textbf{Return:} $\va_{k+1}$
\end{algorithmic}
\end{algorithm}

\noindent\textbf{Parameter tuning in the convergent setting.} The optimization parameters $\alpha, \beta$ and $\gamma$ of Algorithm~\ref{alg:pds-iterates} are not interpretable, and so performing parameter tuning incurs a $\mathcal{O}(K^3)$, where $K$ is the number of discretization values used for each parameter. This is in contrast to Algorithm~\ref{alg:dpg-iterates} which has only two parameters $\theta$ and $\delta$. Since $\theta$ is independently interpretable w.r.t sparsity, we can first tune $\theta$ to produce outputs with desired sparsity level, then tune $\delta$ for appropriate edge weight scale, reducing reducing tuning costs to $\mathcal{O}(K)$.

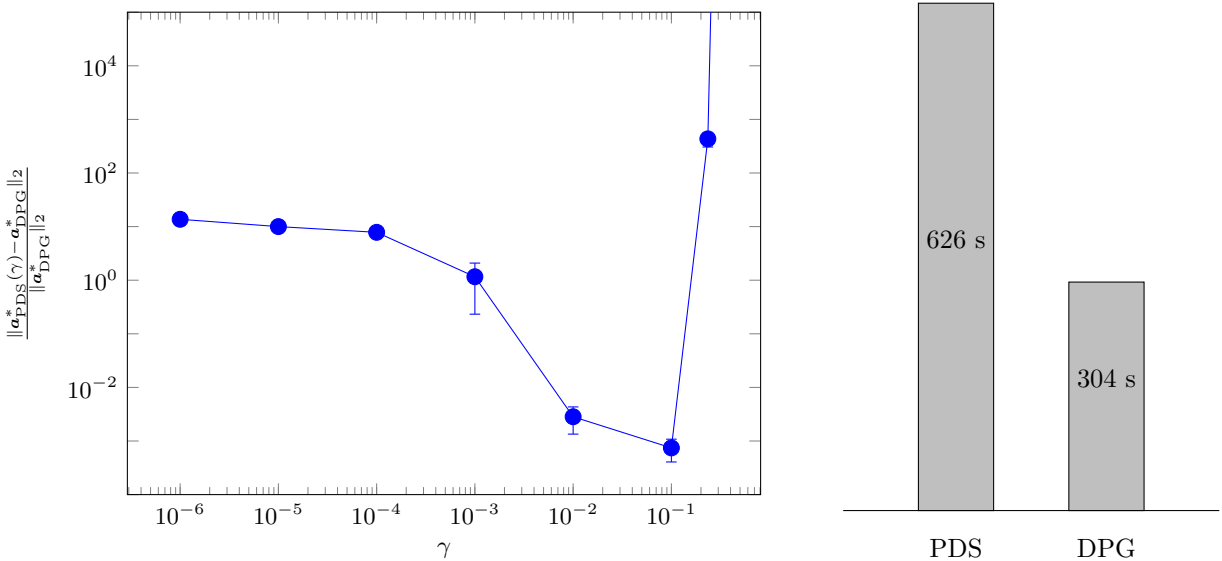
\begin{figure}
    \centering
    \begin{minipage}[b]{0.63\linewidth}
        \centering
        \begin{tikzpicture}
            \begin{axis}[
            xlabel={$\gamma$},
            ylabel={$\frac{\|\va^*_{\text{PDS}}(\gamma) - \va^*_{\text{DPG}}\|_2}{\|\va^*_{\text{DPG}}\|_2}$},
            xmode=log,
            log basis x=10,
            ymode=log,
            log basis y=10,
            ymin=0.00001,
            ymax=10000,
            ytick={0.00001, 0.0001, 0.001, 0.01, 0.1, 1, 10, 100, 1000, 10000},
            yticklabels={, , $10^{-2}$, , $10^{0}$, , $10^{2}$, , $10^{4}$},
            width=10cm,
            height=8cm,
            font=\small
            ]
            \addplot+[mark=*,mark options={scale=1.5,fill=blue},error bars/.cd, y dir=both, y explicit] coordinates {
            (1e-06, 1.36916435) +- (0, 0.32110736)
            (1e-05, 1.00000000) +- (0, 0.00000000)
            (0.0001, 0.78222471) +- (0, 0.10140465)
            (0.001, 0.11593281) +- (0, 0.09269340)
            (0.01, 0.00028351) +- (0, 0.00014908)
            (0.1, 0.00007438) +- (0, 0.00003374)
            %(0.2, 18.74204636) +- (0, 3.78449845)
            (0.235, 43.33791733) +- (0, 12.77402306)
            (0.3, 1e10) +- (0, 1e10)
            };
            \end{axis}
        \end{tikzpicture}
        %\caption{PDS Step Size $\gamma$. Running both DPG and PDS iterations with $\alpha = \beta = 1$ and $\theta = 1/\sqrt{\alpha\beta} = 1$, $\delta = \sqrt{\alpha/\beta}= 1$ for $5\times 10^4$ iterations, we plot the mean and standard deviation of the normalized $\ell_2$ distance between the output of PDS vs DPG. The value of $\gamma$ resulting in the most faithful PDS solution is close to values which yield divergent iterations, problematic for its use as our NN function.}
        %\label{fig:pds-step-size-sensitivity}
    \end{minipage}
    \hfill
    \begin{minipage}[b]{0.32\linewidth}
        \centering
        \begin{tikzpicture}
        \draw (-.5,0) -- (4.5,0);
        \draw[fill=gray!50] (0.5,0) rectangle (1.5,6.75) node[midway, above]{626 s};
        \draw[fill=gray!50] (2.5,0) rectangle (3.5,3.04) node[midway, above]{304 s};
        \node at (1,-0.5) {PDS};
        \node at (3,-0.5) {DPG};
        \end{tikzpicture}
    \end{minipage}
    \caption{\textit{Left}: PDS Step Size $\gamma$. Running both DPG and PDS iterations with $\alpha = \beta = 1$ and $\theta = 1/\sqrt{\alpha\beta} = 1$, $\delta = \sqrt{\alpha/\beta}= 1$ for $5\times 10^4$ iterations, we plot the mean and standard deviation of the normalized $\ell_2$ distance between the output of PDS vs DPG. The value of $\gamma$ resulting in the most faithful PDS solution is close to values which yield divergent iterations, problematic for its use as our NN function. \textit{Right}: PDS vs DPG inference time using $D=200$ on $T=50$ RG$_{\frac{1}{3}}$ graphs running NUTS with $4$ chains (in parallel), each taking $M=1500$ samples.}
    %\label{fig:pds-vs-dpg-inference-time}
    \label{fig:app-pds-vs-dpg-stability-and-speed}
\end{figure}

\noindent\textbf{Step size $\gamma$ frustrates Bayesian PDS.} Algorithm~\ref{alg:pds-iterates} has a nuisance step size parameter $\gamma$ which must be properly tuned for convergent and performant iterations; problematically such $\gamma$ values are a functions of $\alpha$ and $\beta$ values; see~\citep{Kalof2016Smooth}. The step size $\gamma$ introduces several issues for use as a learned parameter in a BNN. 
First, we empirically find that the $\gamma$ values which produce convergent iterations of Algorithm~\ref{alg:pds-iterates} (within $5\times10^4$ iterations) are very close to $\gamma$ values which produce divergent iterations, as shown in Figure~\ref{fig:app-pds-vs-dpg-stability-and-speed} (left). This presents the practical problem of producing NaNs during Bayesian inference, as the sampler will be drawn toward such unstable values of $\gamma$, causing the unrolling to diverge. 
Second, the values of $\gamma$ which produce divergent unrollings are themselves a function of $\alpha$, $\beta$, and the unrolling depth, frustrating simple solutions to prevent divergence, e.g., setting $p(\gamma)$ to be some fixed closed interval. %, which presents itself as divergent paths in HMC. 
%Second, the range of step sizes which produce performant (and non-divergent) iterations has a complicated relationship with the two other parameters $\alpha$ and $\beta$, as well as the depth of the unrolling; this makes it difficult to choose a viable support for $p(\gamma)$. 
Third, the products $\gamma \beta$ and $\gamma \alpha$ ensure the Unrolled PDS is \textit{not a neural network}; this causes problems in practice as products of parameters - each of which can vary over many orders of magnitude - can cause problematic gradients. These observations help explain why we could not find any prior configuration over the three parameters $\{\alpha, \beta, \gamma\}$ in PDS which produced convergent posterior sampling over multiple depths. 
%We were able to find a setting which worked for $D=30$ (...), but when used with $D=200$ it produced NaN's.
Fixing $\gamma$ to a scalar value ameliorates these three issues, indeed this was the only configuration which produced a performant PDS model. We found $\gamma = 0.1$ worked best. Even so, care has to be taken in setting the priors for $\alpha$ and $\beta$: log $\alpha$ and log $\beta$ $\sim U\left[10^{-6}, 10^{6}\right]$ produced all divergent paths - recall this naive setting for log $\theta$ in DPG worked with no issue in the Ablation Studies of Section~\ref{subsec:ablation}. A successful avenue we followed to produce a performant PDS model was to use $\alpha$, $\beta$ $\sim$ LogNormal($0, 10)$ and $b \sim \mathcal{N}(0, 10^3)$. %Doing so for depth $D=30$ produced NLL, BS, Error, and ECE of $(11.413 \pm 5.307, 0.01438 \pm 0.00768, 0.01816 \pm 0.01168, 0.00469)$, and for $D=200$ produces $(10.316 \pm 5.386, 0.01284 \pm 0.00717, 0.01589 \pm 0.01026, 0.00361)$.

\noindent\textbf{Comparing Bayesian inference time of PDS vs DPG.} Above we show that finding a range of reasonable values of the optimisation parameters can be cubic in PDS while linear in DPG, owing to independent interpretability. Further, PDS takes more than twice as long in performing inference as the DPG (and PDS with stochastic $\gamma$ took an order of magnitude longer). This may be due to the increased complexity of PDS iterates and the fact that both $\alpha$ and $\beta$ occur inside PDS iterations, compared to only $\theta$ for DPG iterates, increasing the complexity of the computational graph required to compute parameter gradients. See Figure~\ref{fig:app-pds-vs-dpg-stability-and-speed} (right) showing the difference in inference runtime for PDS and DPG, both having depth $D=200$ using $T=50$ RG$_{\frac{1}{3}}$ graphs.
\subsection{Experimental Details}
\label{app:experimental_details}
\subsubsection{Scoring rules and evaluation metrics} % Error
%\label{app:experimental_details}
Fitting the model to training data $\mathcal{T}$ produces samples $\{\boldsymbol{\Theta}^{(m)}\}_{m=1}^{M}$ from the posterior $p(\boldsymbol{\Theta} \mid \mathcal{T})$. Given test sample $\{\tilde{\ve}, \tilde{\va}\}$ we can evaluate the quality of fit using proper scoring rules, e.g., Log Predictive Density and Brier Score, and discrete metrics, e.g., the Error. % percentage of incorrectly predicted edges where the prediction is taken to be $1$ where the mean predictive posterior is $>0.5$, otherwise $0$. 

\noindent\textbf{Log predictive density, i.e., Log Likelihood.} The Log Likelihood measures the quality of the probabilistic predictions of the model. It is defined as the predicted probability of the true outcome under the model.
\begin{align}
    %\label{app:log_pred_density}
    \text{log }p(\tilde{\va} \mid \tilde{\ve}, \mathcal{T}) 
    &= \text{log }\int p(\tilde{\va} \mid \tilde{\ve}, \boldsymbol{\Theta}) p(\boldsymbol{\Theta} \mid \mathcal{T}) d\boldsymbol{\Theta} \nonumber\\
    &\approx \text{log } \big( \frac{1}{M} \sum_{m=1}^M p(\tilde{\va} \mid \tilde{\ve}, \boldsymbol{\Theta}^{(m)}) \big) \nonumber \\
    &= \text{log } \sum_{m=1}^M p(\tilde{\va} \mid \tilde{\ve}, \boldsymbol{\Theta}^{(m)}) - \text{log }S \nonumber\\
    &= \text{log-sum-exp} 
    \Bigl\{ 
    \text{log }p(\tilde{\va} \mid \tilde{\ve}, \boldsymbol{\Theta}^{(1)}) , 
    \dots, 
    \text{log }p(\tilde{\va} \mid \tilde{\ve}, \boldsymbol{\Theta}^{(M)}) 
    \Bigr\} - \text{log }S \label{app:log_pred_density}
\end{align}
%Where $\text{log }p(\va \mid \vx, \boldsymbol{\Theta}^{(s)}) = \sum_{j=1}^{|\mathcal{E}|} y_j \text{log } \hat{p}_j(\ve; \boldsymbol{\Theta}^{(s)}) + (1-y_j) \text{log } (1- \hat{p}_j(\ve; \boldsymbol{\Theta}^{(s)}))$ comes directly from (\ref{eq:log-joint-factorized}). 
The NLL is simply $-1 \cdot \text{log }p(\tilde{\va} \mid \tilde{\ve}, \mathcal{T})$. A lower NLL indicates better predictive performance, as it suggests that the model assigns higher probabilities to the observed outcomes.

\noindent\textbf{Brier Score.} The Brier Score is used to assess the accuracy of probabilistic predictions. It measures the mean squared difference between the predicted probability and the actual outcome. A lower Brier Score indicates better calibration and accuracy of the probabilistic predictions. Recall the edge-wise probabilities output by the model for parameters $\boldsymbol{\Theta}^{(m)}$ are denoted $\hat{\vp}^{(m)} = \sigma(\delta^{(m)} \Gamma_{\theta^{(m)}}^D(\ve) - b^{(m)})$. Then the Brier Score is %$\hat{p}(\va \mid \ve, \boldsymbol{\Theta}^{(m)}) = %\sigma(\Phi_{\boldsymbol{\Theta}^{(s)}}(\ve) - b^{(s)})$
%\sigma(\delta^{(m)} \Gamma_{\theta^{(m)}}^D(\ve) - b^{(m)})$.
\begin{comment}
\begin{align}
    \text{BS}(\tilde{\va} \mid \tilde{\ve}, \mathcal{T})
    &= \frac{1}{|\mathcal{E}|} \int \|\hat{p}(\tilde{\va} \mid \tilde{\ve}, \boldsymbol{\Theta}) - \tilde{\va}\|_2^2 \cdot p(\boldsymbol{\Theta}|\mathcal{T}) d\boldsymbol{\Theta} \nonumber \\
    &\approx \frac{1}{M|\mathcal{E}|} \sum_{m=1}^M \|\hat{p}(\tilde{\va} \mid \tilde{\ve}, \boldsymbol{\Theta}^{(m)}) - \tilde{\va}\|_2^2 
    \label{app:brier_score}
\end{align}
\end{comment}
%
\begin{align}
    \text{BS}(\tilde{\va} \mid \tilde{\ve}, \mathcal{T})
    &= \frac{1}{|\mathcal{E}|} \int \|\hat{\vp} - \tilde{\va}\|_2^2 \cdot p(\boldsymbol{\Theta}|\mathcal{T}) d\boldsymbol{\Theta} \nonumber \\
    &\approx \frac{1}{M|\mathcal{E}|} \sum_{m=1}^M \|\hat{\vp}^{(m)} - \tilde{\va}\|_2^2 
    \label{app:brier_score}
\end{align}

\noindent\textit{Complementary metrics.} The NLL is particularly useful for evaluating the overall fit of a probabilistic model, emphasizing the accuracy of the assigned probabilities, especially for less likely events. In contrast, the Brier Score provides a more intuitive measure of predictive accuracy and calibration, making it easier to interpret the model's probabilistic predictions in practical scenarios.
\newline\noindent\textit{Numerical Concerns.} Evaluating $\text{log }p(\va | \ve, \boldsymbol{\Theta}^{(m)})$ can cause numerical issues which stem from underflow in computing 
%log $\hat{p}(\tilde{\va} \mid \tilde{\ve}, \boldsymbol{\Theta}^{(m)})$ or 
log $\hat{a}^{(m)}$ or log $(1 - \hat{a}^{(m)})$.
%log $(1 - \hat{p}(\tilde{\va} \mid \tilde{\ve}, \boldsymbol{\Theta}^{(m)}))$
for some edges. % $j\in \mathcal{E}$. 
This becomes an issue in the covariate shift experiments in Section~\ref{sec:experiments}, where the data becomes very noisy and the model can confidently predict the wrong label. When this underflow occurs, the log evaluates to $-\infty$. To solve this we use the softplus parametrization of the log likelihood: $- \text{softplus}(-\bar{y} \cdot (\delta^{(m)} \Gamma_{\theta^{(m)}}^D(\ve) - b^{(m)}))$, where $\bar{y} = 2y - 1$, see Equation (10.13) in~\citep{Murphy2023AdvPML}.
%
%The above procedures are defined for a single sample $\va$. Given a test set with multiple such $\va$'s, we repeat this procedure for each test sample and report the mean and standard deviation in the paper body.

\noindent\textbf{Error.} We take the error to be the percentage of incorrectly predicted edges between our thresholded pred. mean $\mathbb{E}_{\boldsymbol{\Theta} \mid \mathcal{T}}[\tilde{\va} \mid \tilde{\ve}, \mathcal{T}] > 0.5$ and the true label $\tilde{\va}$.

\noindent\textbf{Calibration for GSL.}
We follow the calibration procedure laid out in~\citep{guo2017calibration}. To fit into this framework, we take our prediction to be the pred. mean thresholded at $0.5$, i.e.,  
$\mathbb{E}_{\boldsymbol{\Theta} \mid \mathcal{T}}[\tilde{\va} \mid \tilde{\ve}, \mathcal{T}] > 0.5$. % and the actual label $\tilde{\va}$.
Thus the confidence, i.e., the probabilities associated with the predicted label, will thus always be in $\in \left[0.5, 1\right]$. We will thus construct $M$ uniform bins $I_m = (.5+\frac{.5*(m-1)}{M}, .5+\frac{.5*m}{M}]$, $m=\{1, \dots, M\}$. If an edge confidence is in $I_m$, we assign it to bin $B_m$. We then evaluate the accuracy acc$(B_m) := \frac{1}{|B_m|}\sum_{i \in B_m} \mathbf{1}(\hat{a}_i = a_i)$ and average confidence conf$(B_m) :=  \frac{1}{|B_m|}\sum_{i \in B_m} \hat{a}_i$ for each bin. The Expected Calibration Error (ECE) is defined as the weighted average the bins’ accuracy/confidence difference: 
$$\text{ECE} = \sum_{m=1}^M \frac{|B_m|}{B}|\text{acc}(B_m) - \text{conf}(B_m)|,$$ 
where $B := \sum_{m=1}^M|B_m|$ are the total number of edges predicted. Thus the ECE provides a measure of calibration over all edges in the evaluation set.
\subsubsection{Inference Details}
\noindent\textbf{Details on HMC.}
\begin{figure}
    \centering
    \includegraphics[width=.5\textwidth, vshift=0cm]{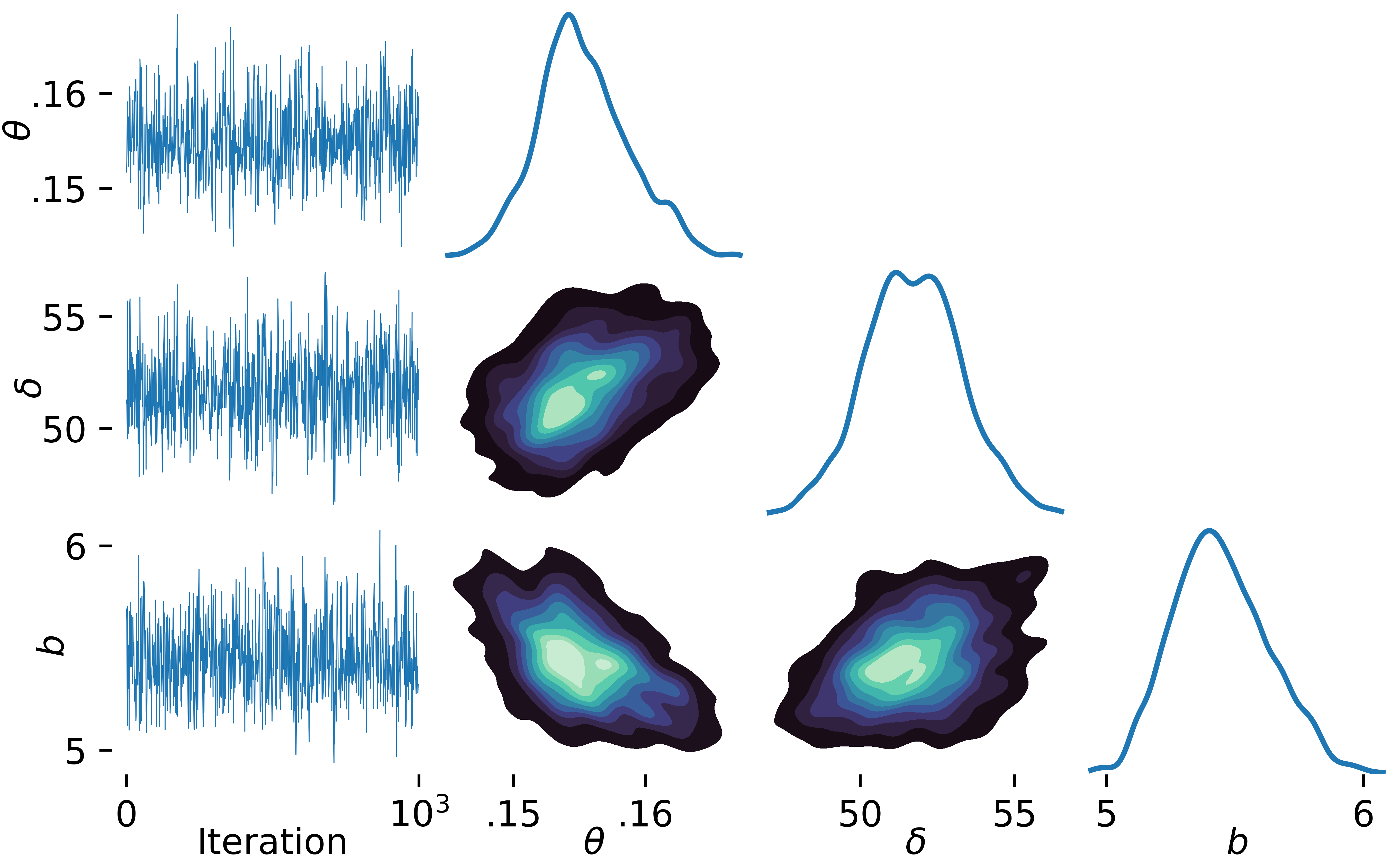}
    \caption{Trace plots and bivariate kernel density estimates of DPGs parameters on RG$_{\frac{1}{3}}$ graphs.} 
    \label{fig:RG_inference_trace_bikde}
\end{figure}
To ensure convergence has been achieved in our posterior sampling we simulate multiple chains with different starting points in the parameter space, discard the first $500$ iterations of each simulation, and monitor relevant diagnostic criteria - namely the Potential Scale Reduction Factor (PSRF) $\hat{r}$ and Effective Sample Size (ESS). We do not perform thinning on the resulting samples. Unless otherwise stated all inference was achieved without issue, namely with that $\hat{r} \approx 1$ % \leq 1.05$ 
and $\hat{n}_{eff} \geq 1000$ for each parameter. For a visualization of successful HMC inference with DPG, see  Figure~\ref{fig:RG_inference_trace_bikde} which displays the trace plots and bivariate kernel density estimates of DPG with $D=30$ on RG$_{\frac{1}{3}}$ graphs using analytic distance matrices.

\noindent\textbf{MAP vs MLE.}
Due to nonconvexity in the loss, initialization is important in finding MAP and MLE of parameters. We did not alter the default initialization provided in NumPyro with MAP, as it worked without issue. We found MLE to be more dependant on initialization; we initialized the parameters of MLE models to the median value of the priors set in Section~\ref{sec:gsl_smooth_signals_with_bnns}, i.e. $\theta, \delta, b \approx 0.32, 100, 1$. %, with some small uniform noise added to theta in the MIMO setting - break symmetry

\noindent\textbf{Log-normal parameterization.}
Let $Z$ be a standard normal variable and let $\mu$ and $\sigma > 0$ be two real numbers, then the distribution of the random variable $X = e^{\mu + \sigma Z}$ is called a log-normal distribution with parameters $\mu$ and $\sigma$, i.e. $X \sim \text{Lognormal}(\mu, \sigma^2)$. Parameters $\mu$ and $\sigma$ represent the mean and standard deviation of $\log X$, not $X$ itself. While the normality of log $X$ is true regardless of base, we will be performing modeling using powers of $10$. Thus in the main body when we specify a distribution $\text{Lognormal}(\mu, \sigma^2)$, it implies the location parameter used is $\text{ln } 10^{\mu}$, while the scale parameter is unchanged. For example, $\text{Lognormal}(2, 4)$ would imply location $\mu = \text{ln } 10^{2}$ and scale $\sigma = 2$, and thus would be invoked as a call to the probabilistic programming language distribution, here NumPyro, as numpyro.distributions.LogNormal(loc $= \text{ln } 10^{2}$, scale $ = 2$).

\subsection{Scaling to Large Graphs}
\label{app:scaling}
\begin{figure}[t]
    \centering
    \includegraphics[width=.5\textwidth, vshift=0cm]{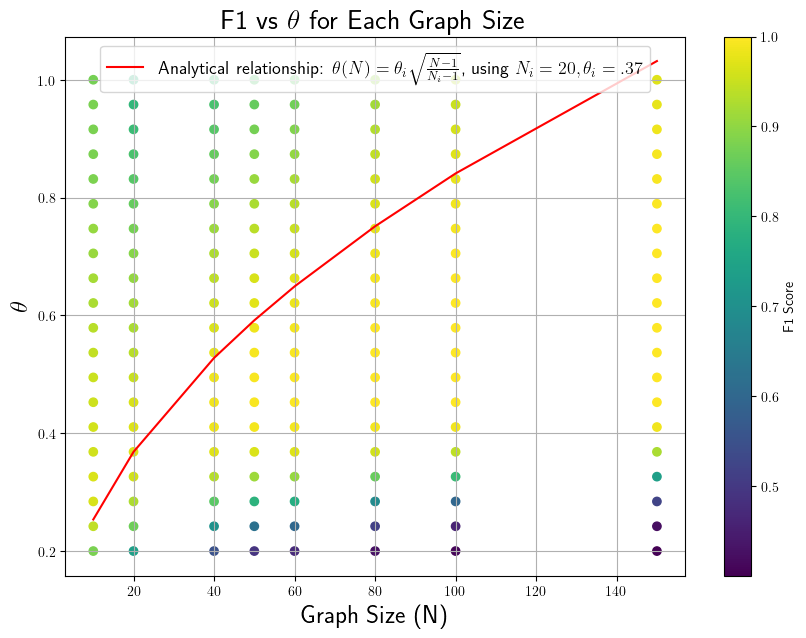}
    \caption{Transferring to Larger Graphs. Using $T=20$ ER$_{\frac{1}{4}}$ graphs of varying sizes, we run a grid search to find optimal $\theta$ for the structure recovery task using Algorithm~\ref{alg:dpg-iterates} for each graph size, and show the F$1$ score for each $\theta$. We also plot the line corresponding to Equation~\ref{eq:scaling_analytically_predicted_theta} which extrapolates performant $\theta$ values from the $N_i=20$ setting. This line coincides with empirically observed highly performant $\theta$ values in problem setting for both smaller and significantly larger graphs.}
    \label{fig:size_gen_empirical_and_analytical_theta}
\end{figure}
\noindent\textbf{Scaling via transfer learning in the convergent setting.}
In (\ref{eq:smooth-graph-optim-problem-vectorized}), terms $2\va^{\top}\ve, \frac{\beta}{2} \|\va\|^2_2$ have $N(N-1)/2$ components while term $- \alpha \mathbf{1}^{\top} \text{log}(\mS \va)$ has $N$ components, thus the former terms will tend to dominate the latter as $N$ grows (using fixed $\alpha$, $\beta$). We thus create a new problem where each term is normalized by it's number of components; we use the notation $\bar{\alpha}, \bar{\beta}$ to indicate the \textit{scale invariant} parameter values
\begin{align}
    \argmin_{\va} & \hspace{3pt} \left\{|\mathcal{E}_N|^{-1}2\va^{\top}\ve - N^{-1}\bar{\alpha} \mathbf{1}^{\top} \text{log}(\mS \va) + |\mathcal{E}_N|^{-1}\frac{\bar{\beta}}{2} \|\va\|^2_2 + \mathbb{I}\{\va \geq 0\}\right\} \label{eq:size_normalized_smooth-graph-optim-problem-vectorized}\\
    &=\va^*(|\mathcal{E}_N|^{-1}\ve, N^{-1}\bar{\alpha}, |\mathcal{E}_N|^{-1}\bar{\beta})  \hspace{5pt} \text{, $1$-$1$ correspondence between objective terms and params/inputs} \nonumber\\
    &=\va^*(\ve, |\mathcal{E}_N|N^{-1}\bar{\alpha},\bar{\beta}) \hspace{20pt}\text{, argmin invariant to positive scaling, multiply by $|\mathcal{E}_N|$} \nonumber\\
    &=\va^*(\ve, .5(N-1)\bar{\alpha},\bar{\beta}). \label{eq:size_normalized_scale_alpha}
    %
    %&=\bar{\delta} \big(.5(N-1)\big)^{\frac{1}{2}} \va^*(\bar{\theta}\big(.5(N-1)\big)^{-\frac{1}{2}} \ve, 1, 1)\\
\end{align}
Intuitively, (\ref{eq:size_normalized_scale_alpha}) tells us to scale $\alpha$ by $\mathcal{O}(N)$. This makes sense, as $\alpha$ itself scales the objective term which grows at a factor of $\mathcal{O}(N)$ slower than the other (finite) terms. This should help correct the differing component sizes as the problem dimension $N$ grows. Eq. (\ref{eq:size_normalized_scale_alpha}) formulates the solution to the size normalized problem with respect to the solutions of original problem (\ref{eq:smooth-graph-optim-problem-vectorized}). This shows that we can use the original solution procedures, now with scaled $\alpha$, to produce solutions. To do so, suppose we find optimal parameter values $\alpha_i$, $\beta_i$ to the original problem (\ref{eq:smooth-graph-optim-problem-vectorized}) with problem size $N_i$. To find the proper parameter values for a different problem size $N$, we can solve for the scale invariant parameter values, then re-scale to the appropriate size as
\begin{align*}
\alpha(N) &= \overbrace{\alpha_i \big(.5(N_i-1)\big)^{-1}}^{\bar{\alpha}} .5(N-1)\\
\beta(N) &= \underbrace{\beta_i}_{\bar{\beta}}.
\end{align*}
Note as a sanity check, when $N = N_i$ we have $\alpha(N) = \alpha_i$ and $\beta(N) = \beta_i$.
% now cover $\theta, \delta$ parameterization
It is trivial to recover similar functions for $\theta(N)$ and $\delta(N)$ by substituting into the definitions of $\theta = \frac{1}{\sqrt{\alpha\beta}}$ and $\delta = \sqrt{\frac{\alpha}{\beta}}$:
\begin{alignat}{4}
\theta(N) & = \frac{1}{\sqrt{\alpha(N)\beta(N)}} & & = \overbrace{\theta_i\big( .5 (N_i - 1 ) \big)^{\frac{1}{2}}}^{\bar{\theta}} & & \big( .5 (N - 1 ) \big)^{-\frac{1}{2}} & &= \theta_i \sqrt{ \frac{ N_i - 1 }{ N - 1 } } \label{eq:scaling_analytically_predicted_theta}\\
\delta(N) & = \sqrt{\alpha(N)\beta(N)^{-1}} & & = \underbrace{\delta_i \big( .5 (N_i - 1 ) \big)^{-\frac{1}{2}}}_{\bar{\delta}} & & \big( .5 (N - 1 ) \big)^{\frac{1}{2}} & &= \delta_i \sqrt{ \frac{ N- 1 }{ N_i - 1 } }.
\end{alignat}
See Figure~\ref{fig:size_gen_empirical_and_analytical_theta} which empirically validates this derivation on ER$_{\frac{1}{4}}$ graphs across an order of magnitude difference in problem size $N$.

\begin{figure}
    \centering
    \begin{minipage}{.48\textwidth}
        \centering
        \includegraphics[width=1\linewidth]{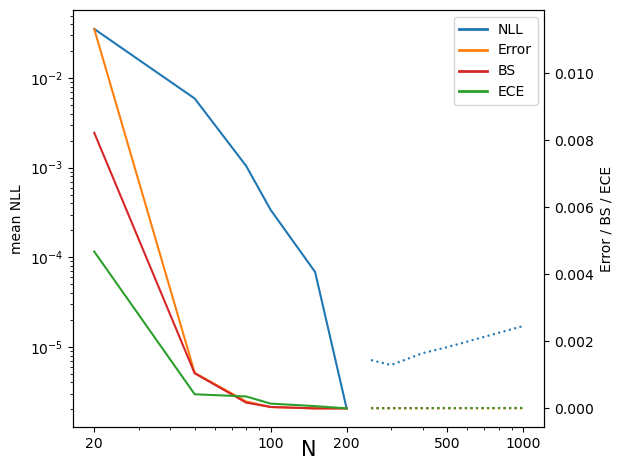}
        %\caption{Computing MAP estimates at increasing graph sizes shows improving recovery performance (solid lines). Extrapolating from the trends we can perform transfer learning larger graph sizes (dotted lines) with gradual decay in performance.}
        %\label{fig:size_gen_map_performance}
    \end{minipage}
    \hfill
    \begin{minipage}{.48\textwidth}
        \centering
        \includegraphics[width=1\linewidth]{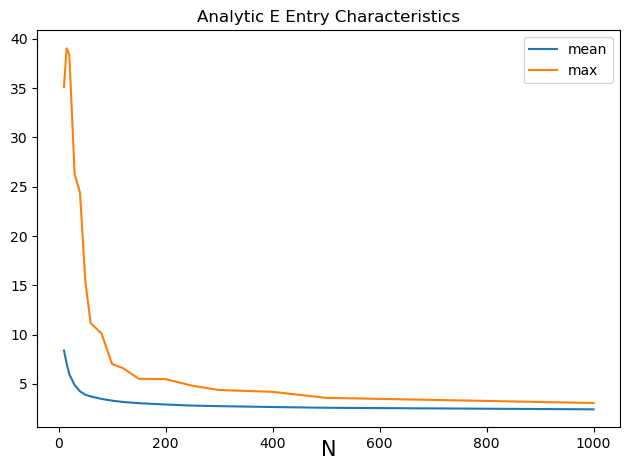}
        %\caption{The analytic distance matrix E get smoother for increasingly sized ER$_{\frac{1}{4}}$ graphs.\newline\newline\newline}
        %\label{fig:size_gen_analytic_e_entries}
    \end{minipage}
    \caption{
    \textit{Left}: Computing MAP estimates of DPG at increasing graph sizes shows improving recovery performance (solid lines). Extrapolating from the trends using the approach outlined in Appendix~\ref{app:scaling}, we can perform transfer learning larger graph sizes (dotted lines) with gradual decay in performance. 
    \textit{Right}: The analytic distance matrix E get smoother for increasingly sized ER$_{\frac{1}{4}}$ graphs, providing an explanation for why graph recovery improves with larger graph sizes.}
    \label{fig:MAP-and-smoother-analyt-dist}
\end{figure}

\noindent\textbf{Scaling in the unrolled setting.}
See Figure~\ref{fig:MAP-and-smoother-analyt-dist} (left) for the empirical growth trends of MAP estimates of the Unrolled DPG model ($D=200$) parameters on ER$_{\frac{1}{4}}$ graphs. Parameter $\theta$ seems to follow the trend predicted by our analysis, but $\delta$ does not. Indeed it seems to follow a strongly linear trend, and $b$ seems to follow a power law. We can extrapolate from these trends to parameterize Unrolled DPG networks on significantly larger graphs with excellent performance; see Figure~\ref{fig:MAP-and-smoother-analyt-dist} (left). % we use a linear fit to \delta(N) = 7.76*N -68.54 , and use constant \theta = 0.36, b = 11.
\begin{figure}[t]
    \centering
    \includegraphics[width=\textwidth, vshift=0cm]{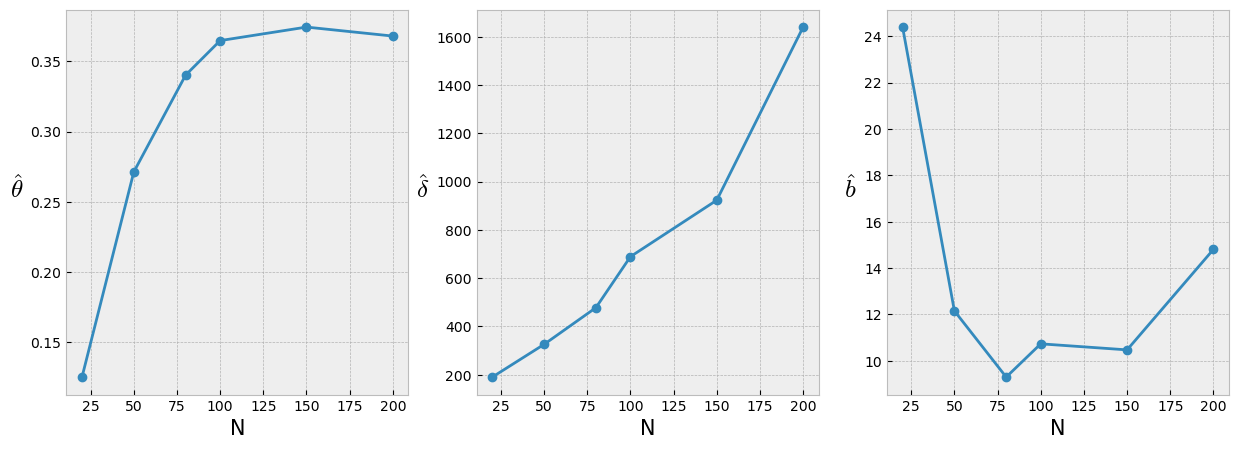}
    \caption{Learning MAP estimates $\hat{\boldsymbol{\Theta}}$ of the DPG parameters on ER$_{\frac{1}{4}}$ graphs of increasing size $N$ shows that $\hat{\theta}(N)$ resemble logarithmic growth, $\hat{\delta}(N)$ are linear, while $\hat{b}(N)$ show a sort of power law decay.}
    \label{fig:size_gen_map_growth_trend}
\end{figure}

\noindent\textbf{Larger ER graphs are smoother.} We also notice an interesting trend when constructing smooth graph as in Section~\ref{sec:graph-learning-smooth-signals}. We find that the analytical distance matrices produced by ER$_{\frac{1}{4}}$ graphs get smoother as their size increases. For instance, see Figure~\ref{fig:MAP-and-smoother-analyt-dist} (right) where the mean and maximum entries of the analytic distance matrix get smaller as $N$ grows larger. This explains why the graph recovery problem gets easier and thus our performance gets better, as shown in Figure~\ref{fig:MAP-and-smoother-analyt-dist} (left).
\subsection{Further Experiments}
\label{app:further_experiments}
\begin{figure}
\centering
\begin{minipage}[b]{0.325\textwidth}
    % RG Unsupervised
    \includegraphics[width=\textwidth, vshift=0cm]{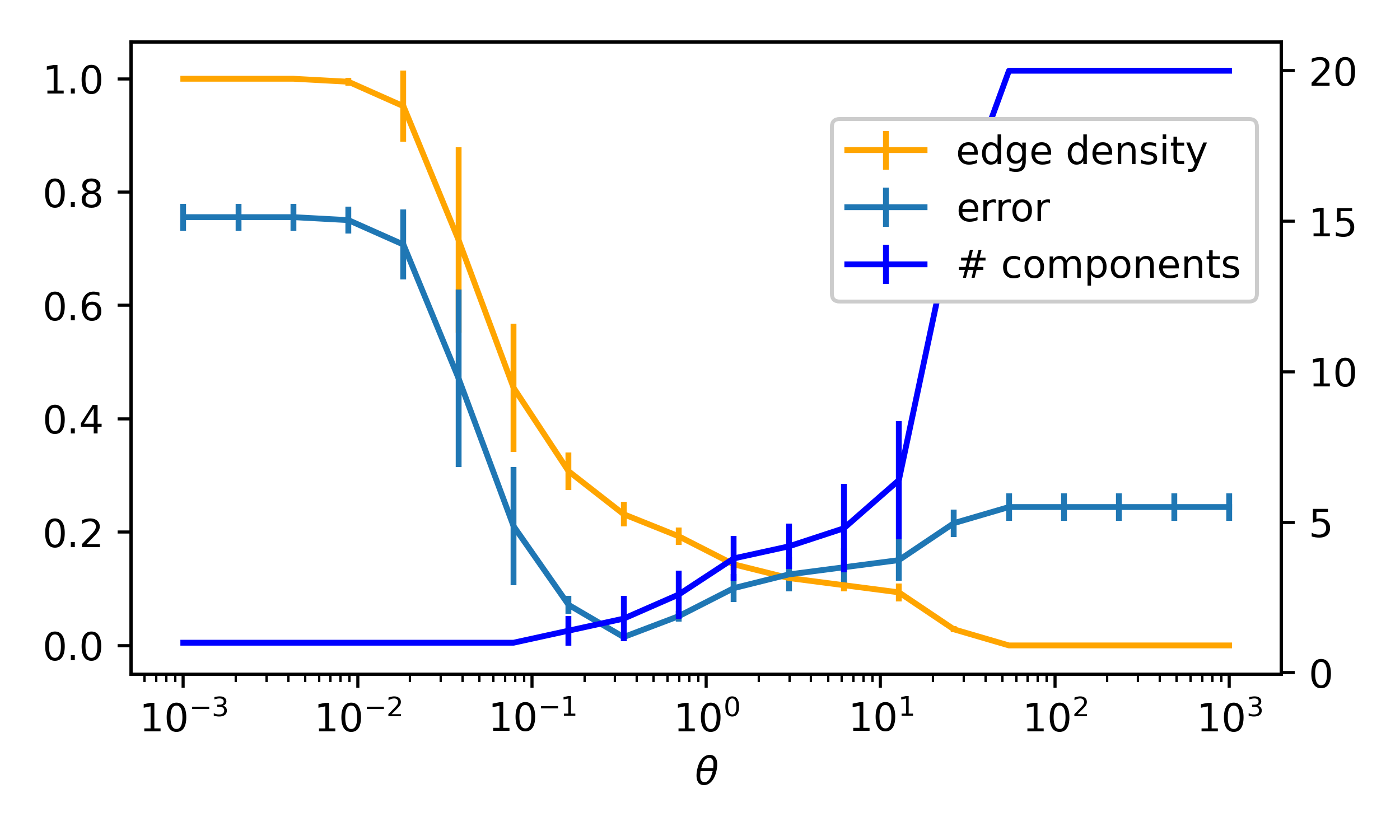}
\end{minipage}
\hfill
\begin{minipage}[b]{0.325\textwidth}
    % ER Unsupervised
    \includegraphics[width=\textwidth, vshift=0cm]{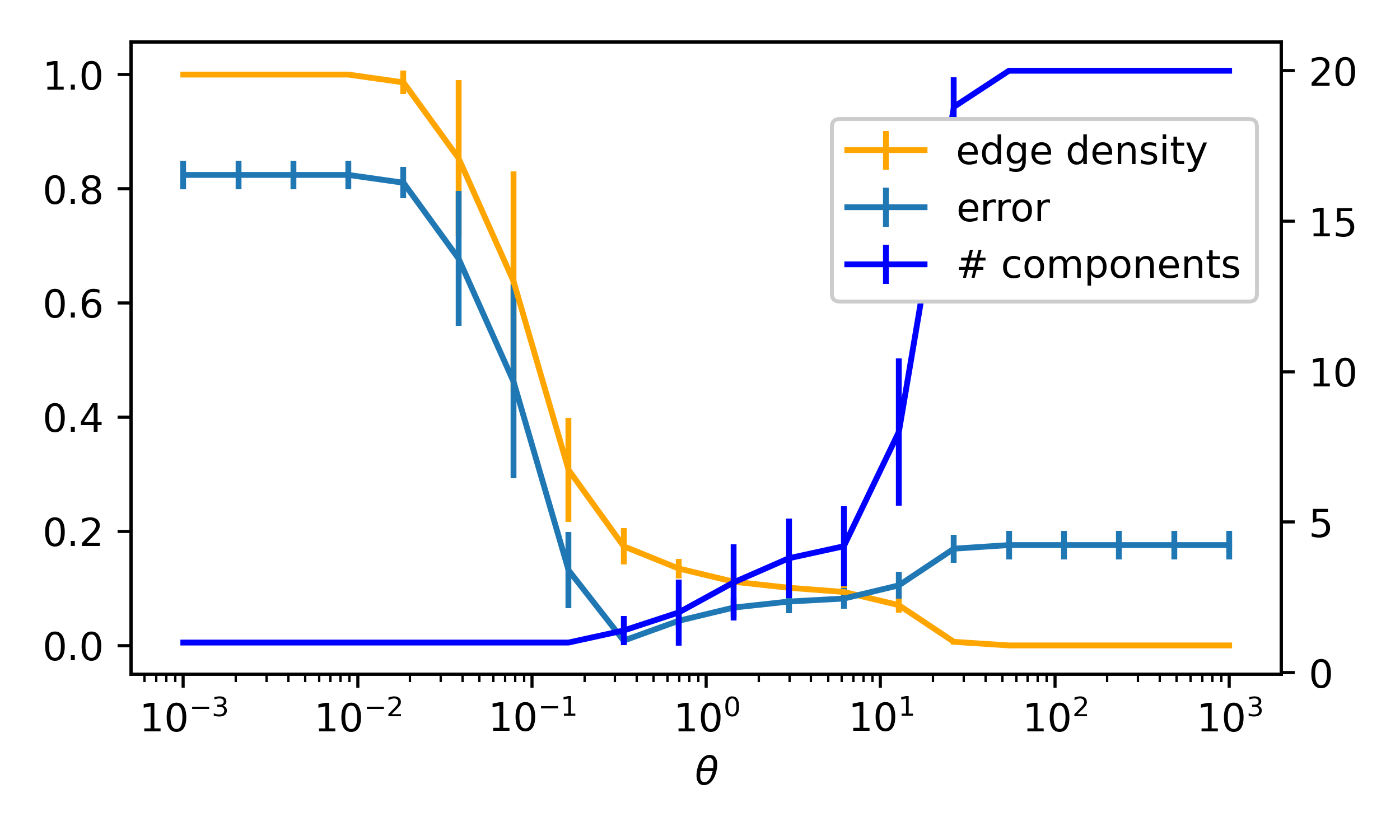}
\end{minipage}
\hfill
\begin{minipage}[b]{0.325\textwidth}
    % BA Unsupervised
    \includegraphics[width=\textwidth, vshift=0cm]{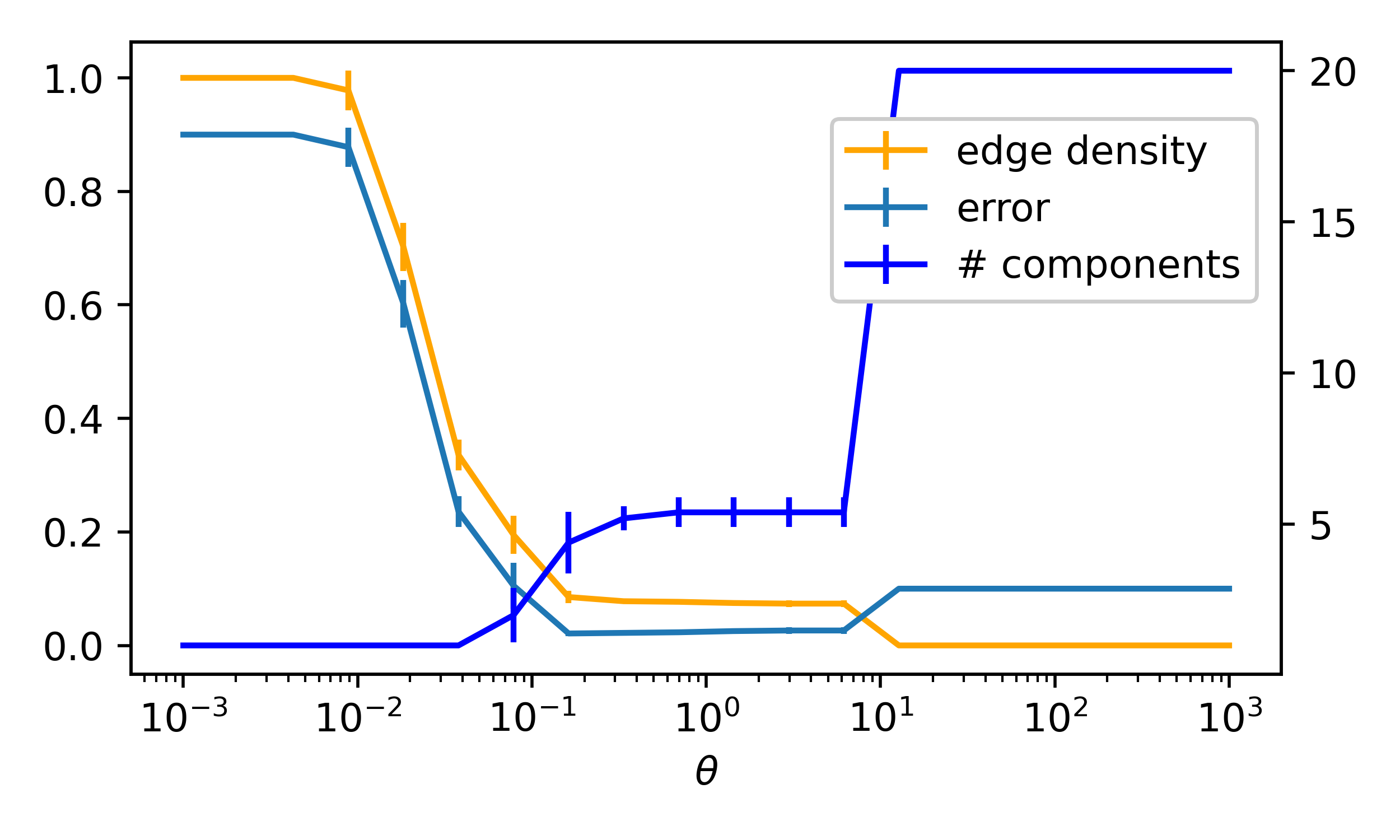}
\end{minipage}
\caption{Prior modeling of $\theta$ for synthetic ensembles RG$_{\frac{1}{3}}$, ER$_{\frac{1}{4}}$, and BA$_1$ from left to right. The edge density and number of connected components do not require the presence of labels, while the error does. A threshold of $10^{-5}$ is used to decide the existence of an edge in order to remove the influence of small numerical effects.}
\label{fig:appendix_theta_prior_modeling}
\end{figure}

\begin{figure}
\begin{minipage}[b]{0.325\textwidth}
    % RG Unsupervised
    \includegraphics[width=\textwidth, vshift=0cm]{figures/prior_modeling_RG_edge_weight_boxplot.png}
\end{minipage}
\hfill
\begin{minipage}[b]{0.325\textwidth}
    % ER Unsupervised
    \includegraphics[width=\textwidth, vshift=0cm]{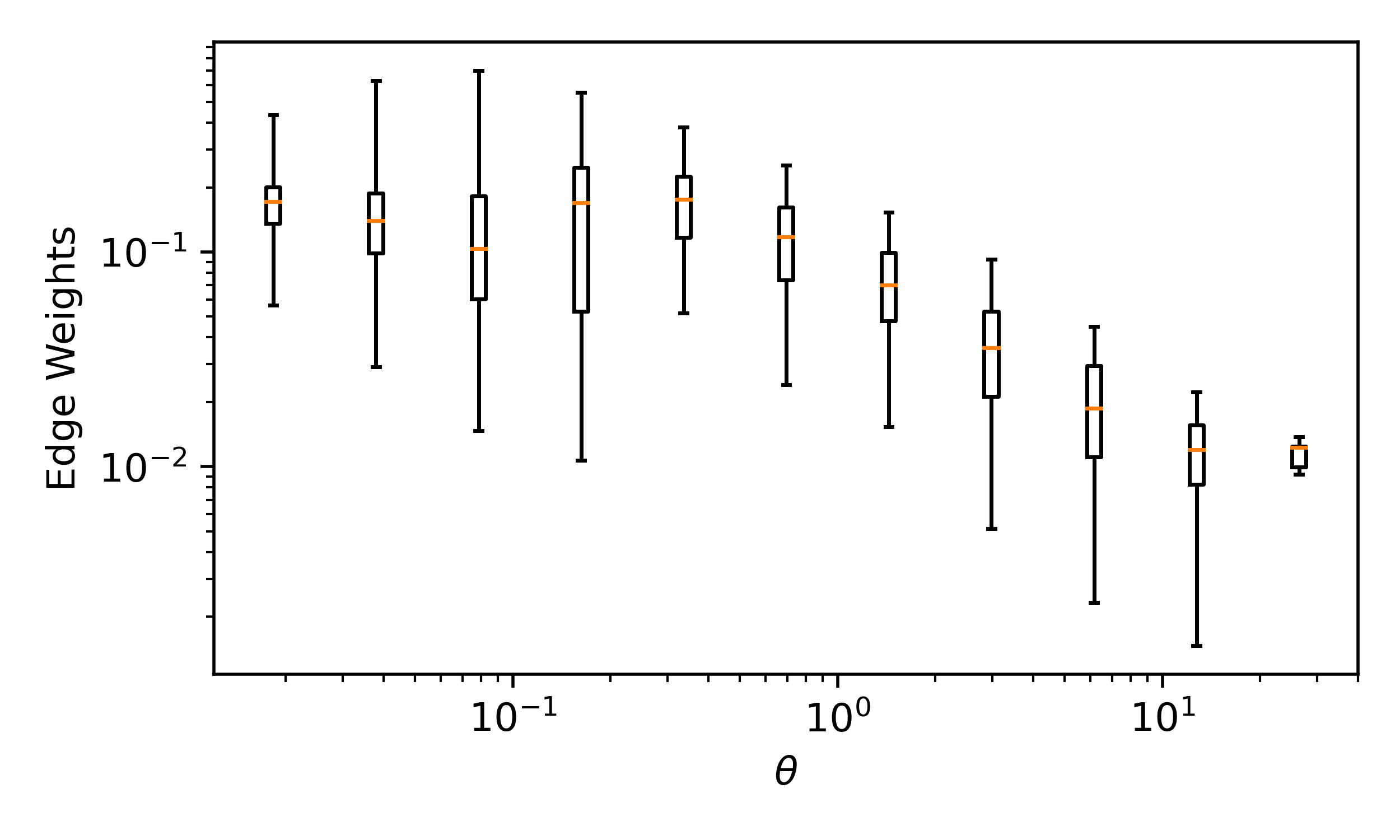}
\end{minipage}
\hfill
\begin{minipage}[b]{0.325\textwidth}
    % BA Unsupervised
    \includegraphics[width=\textwidth, vshift=0cm]{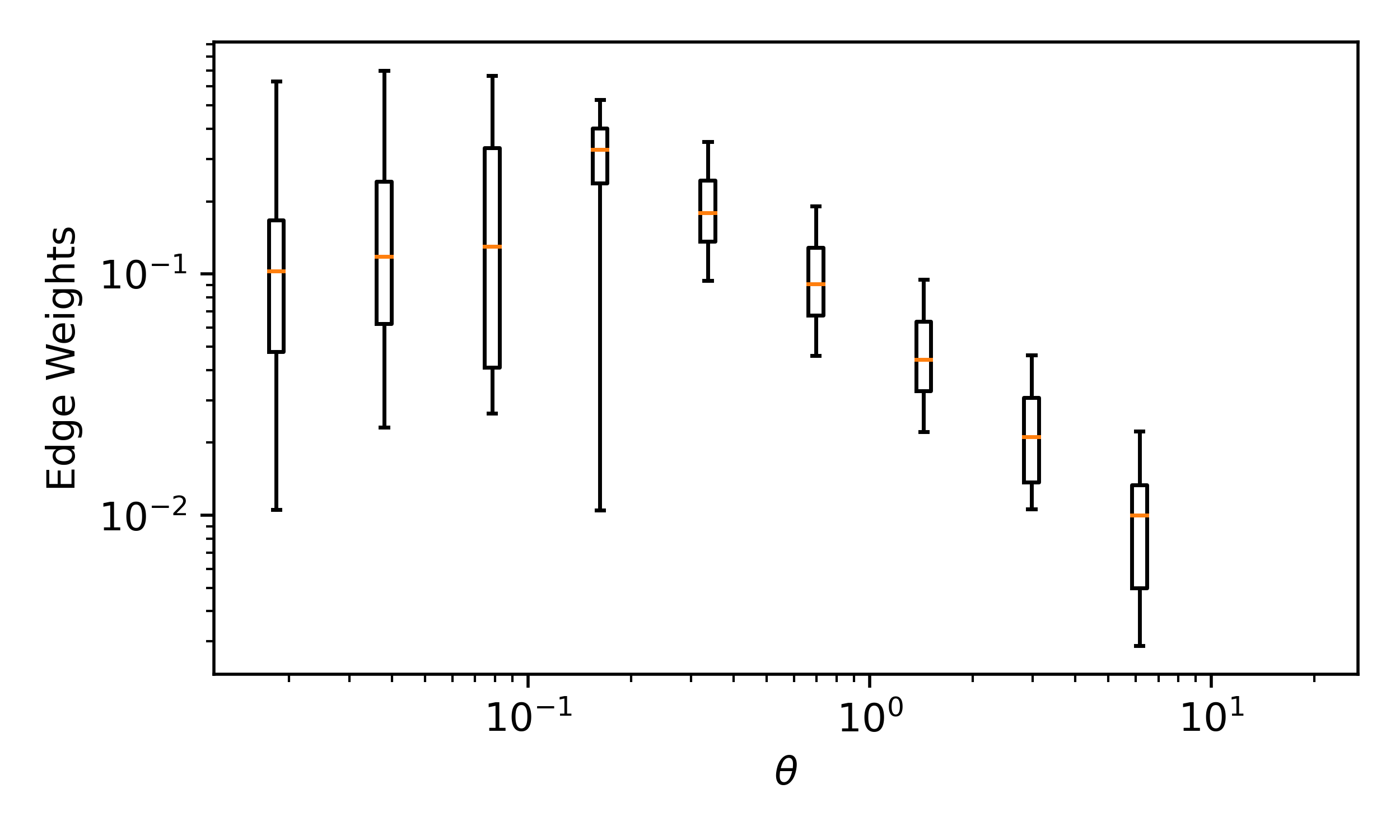}
\end{minipage}
\caption{Prior modeling of $b$ and $\delta$ for synthetic ensembles RG$_{\frac{1}{3}}$, ER$_{\frac{1}{4}}$, and BA$_1$ from left to right. We inspect the edge weight distributions recovered running the held out data through Algorithm~\ref{alg:dpg-iterates} for discretized values of $\theta$. The recovered edge weights are non-negative, and so the relevant scaling of $\delta$ and $b$ is a function of the median and maximum values.}
\label{fig:appendix_edge_weight_boxplot}
\end{figure}

\noindent\textbf{Prior modeling.}
The prior modeling techniques introduced in Section~\ref{sec:bayesian_modeling} work across multiple random graph ensembles. Here, we replicate the prior modeling experiments done on RG$_{\frac{1}{3}}$ in Section~\ref{sec:bayesian_modeling} for ER$_{\frac{1}{4}}$, and BA$_{1}$. See Figure~\ref{fig:appendix_theta_prior_modeling} and Figure~\ref{fig:appendix_edge_weight_boxplot} and compare to Figure~\ref{fig:prior_modeling_theta_delta} (left).

\noindent\textbf{I.i.d. generalization across multiple random graph distributions.}
Our method is performant across the random graph distributions we've tested. To supplement the results shown in Section~\ref{sec:experiments}, we show i.i.d. generalization results across more random graph distributions, as shown in Table~\ref{tab:iid_extra_experiments}.
\begin{table}[ht]
\centering
% Depth 30, 500 warm up, 1000 samples
\caption{I.I.D. Generalization of DPG across Random Graph Distributions}
\label{tab:iid_extra_experiments}
\begin{tabular}{
  l
  S[table-format=2.3(3)]
  S[table-format=1.3(3)]
  S[table-format=1.3(3)]
  S[table-format=2.2]
}
\toprule
{Graph Distribution} & {NLL} & {BS (\si{\times 10^{-2}})} & {Error (\%)} & {ECE (\si{\times 10^{-3}})} \\
\midrule 
%RG$_{\frac{1}{3}}$   & 11.183\pm5.652   & 1.392 \pm 0.743 & 1.721 \pm 1.034  & 3.32  \\% 46.789 s
RG$_{\frac{1}{2}}$   & 8.993\pm9.516    & 0.988 \pm 0.945 & 1.111 \pm 1.267  & 4.37  \\% 228.95 s
BA$_1$               & 14.724\pm3.704   & 2.614 \pm 0.751 & 2.916 \pm 1.003  & 30.63  \\% 345.43 s
ER$_{.15}$           & 7.677\pm5.021    & 0.997 \pm 0.569 & 1.237 \pm 0.879  & 3.04  \\% time 100.627 s
ER$_{\frac{1}{2}}$           & 0.962\pm3.029    & 0.090 \pm 0.236 & 0.084 \pm 0.265  & 0.94  \\% 239.637 s
\bottomrule
\end{tabular}
\end{table}

\begin{figure}
    \centering
    \includegraphics[width=.8\textwidth, vshift=0cm]{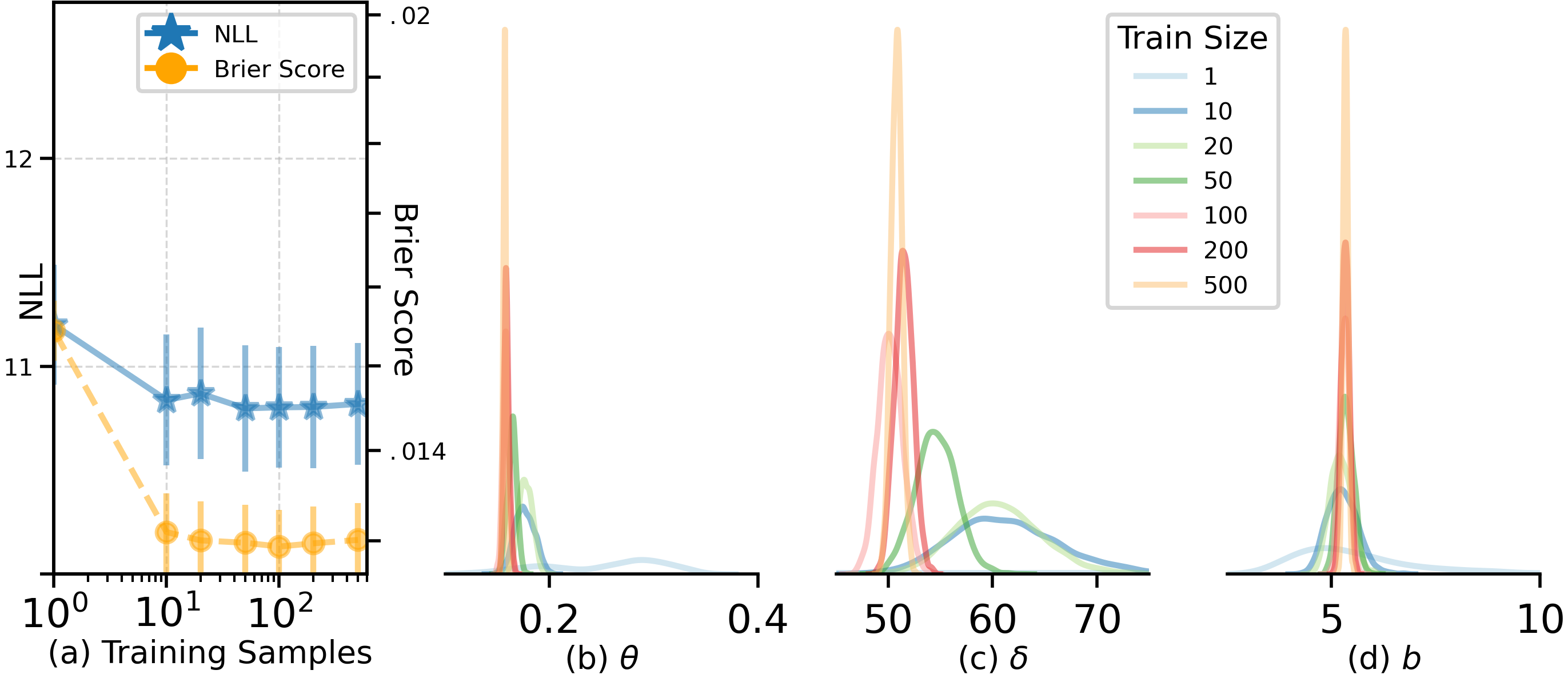}
    \caption{Impact of training set size on i.i.d. generalization performance. We use DPG with 
    %$D_i=10^2, D_p=500$, %to ensure computation not limiting us
    $D=200$ on RG$_{\frac{1}{3}}$ graphs with analytic distance matrices to estimate generalization. 
    The leftmost plot shows the generalization performance converge after $\approx 50$ samples on these same test graphs (error bars indicate $.05*$stdv for compact visualization). The rightmost plots depict posterior marginals for the $3$ parameters - $\theta$, $\delta$, and $b$ - for increasing sizes of training data. The observed reduction in variance of the posterior is a reflection of lower epistemic uncertainty. The posterior shows little change after $\approx 10^2$ samples.}
    \label{fig:data_efficieny}
\end{figure}

\noindent\textbf{Data efficiency.} To further investigate how robust DPG is to sparse data settings, in Figure~\ref{fig:data_efficieny} we display i.i.d. generalization performance as a function of training set size $T$ on RG$_{\frac{1}{3}}$ graphs and their corresponding analytic distance matrices. We additionally visualize the parameter posterior marginals. We find that past $T \approx 50$, i.i.d. generalization did not significantly improve, while past $T \approx 10^2$, the parameter posterior marginals converge.

\noindent\textbf{Prior ablation study}. In sparse data regimes DPG is more performant and has more efficient inference than DPG with un-informative priors (DPG-U). Indeed when the lack of data weakens the likelihood relative to the prior, a useful prior should maintain efficient Bayesian inference~\citep{gelman2017prior}. Table~\ref{tab:ablation-prior} illustrates this using RG$_{\frac{1}{3}}$ graphs and their corresponding analytic distance matrices on data sets of varying sizes $T$; see Section~\ref{tab:ablation-prior} for further discussion.

\begin{table}
\caption{Ablation Study: Removing Informative Priors with Limited Training Data. 
%In sparse data regimes DPG is more performant and has more efficient inference than DPG with un-informative priors (DPG-U). The following shows i.i.d. evaluation on data sets of varying sizes $T$ using RG$_{\frac{1}{3}}$ graphs. % and their analytic distance matrices.
%Informative Prior Benefits in Low Data Setting. Decreasing amount of RG$_{\frac{1}{3}}^*$ training data $T$ shows the benefit of informative prior modeling (DPG) over uninformative prior modeling (DPG-U): more efficient inference and more performant models. clear the more efficient posterior inference and more performant models.
%The informative prior allows for efficient inference and ensures graceful degradation of generalization performance as the dataset size $T$ decreases. Both models use inference and prediction depths of $30$. `DPG-U` refers to the uninformative DPG model.
}
\label{tab:ablation-prior}
\centering
\begin{tabular}{
    l
    l
    S[table-format=3.2]
    S[table-format=2.3]
    S[table-format=2.5]
    S[table-format=1.5]
    S[table-format=1.2]
    S[table-format=2.1]
}
\toprule
{} & {Model} & {Time (s)} & {NLL} & {BS (\si{\times 10^{-2}})} & {Error (\%)} & {ECE (\si{\times 10^{-3}})} & {ESS\(_\theta\)} \\
\midrule
\multirow{2}{*}{\(T = 50\)} & {DPG-U} & 343.97 & 11.185 & 1.394 & 1.721 & 3.41 & 3.9 \\
                            & {DPG}   & 44.15  & 11.183 & 1.392 & 1.721 & 3.32 & 19.7 \\
\addlinespace
\multirow{2}{*}{\(T = 5\)}  & {DPG-U} & 15.15  & 11.476 & 1.508 & 1.847 & 5.62 & 15.5 \\
                            & {DPG}   & 10.63  & 11.398 & 1.506 & 1.842 & 4.87 & 15.7 \\
\addlinespace
\multirow{2}{*}{\(T = 2\)}  & {DPG-U} & 17.06  & 12.125 & 1.540 & 1.821 & 8.75 & 10.5 \\
                            & {DPG}   & 10.66  & 11.661 & 1.522 & 1.832 & 7.55 & 11.3 \\
\addlinespace
\multirow{2}{*}{\(T = 1\)}  & {DPG-U} & 11.11  & 11.942 & 1.673 & 1.789 & 7.98 & 7.1 \\
                            & {DPG}   & 6.59   & 10.976 & 1.498 & 1.642 & 4.23 & 7.8 \\
\bottomrule
\end{tabular}
\end{table}

\end{document}